\documentclass[11pt]{article}

\usepackage[final]{acl}

\usepackage{times}
\usepackage{latexsym}

\usepackage[T1]{fontenc}

\usepackage[utf8]{inputenc}

\usepackage{microtype}

\usepackage{inconsolata}

\usepackage{graphicx}
\usepackage{algorithm}
\usepackage{algpseudocode}
\usepackage{textcomp}
\usepackage{newfloat}
\usepackage{listings}
\usepackage{booktabs}
\usepackage{amsmath}
\usepackage{amsfonts}
\usepackage{xcolor}
\usepackage[table]{xcolor}
\usepackage{multirow}
\usepackage{makecell}
\usepackage{subcaption}
\usepackage{booktabs}
\usepackage{pifont}
\usepackage{tikz}
\usepackage{soul} 

\definecolor{BenchGreen}{RGB}{3,151,92}
\definecolor{BenchRed}{RGB}{201,52,52}
\definecolor{BenchOrange}{RGB}{255,129,3}

\newcommand{\cmark}{\textcolor{BenchGreen}{\ding{51}}}
\newcommand{\xmark}{\textcolor{BenchRed}{\ding{55}}}
\newcommand{\pmark}{\tikz[baseline=-0.6ex]{
  \draw[BenchOrange, line width=0.5pt] (0,0) circle (0.35em);
  \draw[BenchOrange, line width=0.5pt] (-0.18em,0) -- (0.18em,0);
}}

\title{Beyond Factual Knowledge: Benchmarking and Learning Step-Level Procedural Rule Reasoning in Large Language Models}

\author{Bohan Yu$^{1,2}$, Pengfei Cao$^{2,3}$\thanks{\,\,\,Corresponding authors.}, Chen Han$^{1,6}$, Chenxi Zhou$^{1,2}$, \\
\textbf{Zhiheng Zhang$^{2,3}$, Zhiyang Xie$^{2,3}$, Wenhao Teng$^4$, Xiangwen Liao$^5$, Jun Zhao$^{2,3}$, Kang Liu$^{2,3}$\footnotemark[1]} \\
         $^1$School of Advanced Interdisciplinary Sciences, University of Chinese Academy of Sciences \\
  $^2$The Key Laboratory of Cognition and Decision Intelligence for Complex Systems,\\ Institute of Automation, Chinese Academy of Sciences\\
  $^3$School of Artificial Intelligence, University of Chinese Academy of Sciences \\
  $^4$Department of Gastrointestinal Surgery, Fujian Provincial Cancer Hospital\\
  $^5$College of Computer and Data Science,  Fuzhou University \\
  $^6$Academy of Mathematics and Systems Science, Chinese Academy of Sciences \\
  \texttt{yubohan2025@ia.ac.cn \quad \{pengfei.cao,jzhao,kliu\}@nlpr.ia.ac.cn}}

\begin{document}
\maketitle

\begin{abstract}
Large language models (LLMs) excel at text understanding and generation, yet still struggle to reliably understand and apply externally provided procedural rules at scale. To evaluate this capability, we introduce \textbf{RuleWorld}, a large-scale benchmark that reformulates rules as globally reusable abstract units rather than instance-specific facts. In RuleWorld, several scenarios, including single-rule, parallel multi-rule, and multi-hop reasoning, are settled for comprehensive evaluation. 
We further propose \textbf{DynaRule}, an end-to-end framework that injects the given rules into the KV cache and turns retrieval into an internal, learnable, step-wise process.
Specifically, DynaRule employs Stacked Step-Level Attention Training with a special \texttt{<search>} token to enable dynamic rule re-attention and updating during inference. In this way, the model can re-attend to the most relevant rules at each step, dynamically replacing outdated ones to support more stable multi-step reasoning.
Experiments on RuleWorld show that existing LLMs face challenges under large rule pools, while DynaRule improves average QA accuracy by up to 19 points and achieves over 85\% Recall@1 at 10K rules, outperforming strong baselines by large margins. We make our code and dataset available here: \href{https://github.com/SharkSpicy-NLP/Beyond-Factual-Knowledge}{Beyond-Factual-Knowledge}.
\end{abstract}

\section{Introduction}
Large language models (LLMs) excel at text understanding, analysis, and generation~\cite{openai2024gpt4technicalreport,zhao2023survey}, motivating a substantial body of work that evaluates their knowledge capabilities~\cite{petroni2019language-models-as-knowledge-bases,lin2022truthfulqa,huang2023towards-reasoning-llms}.
However, these evaluations primarily emphasize the capabilities of understanding and applying factual knowledge, while often ignoring procedural knowledge, i.e., knowledge of how to act or reason.
In realistic settings, procedural knowledge is maintained as a large repository of reusable procedural entries (e.g., policies, guidelines, or rules), requiring models to localize and compose the relevant items across reasoning steps.
In this paper, we focus on rules as a concrete and widely used form of procedural knowledge.
To evaluate related capabilities of LLMs on rules, existing benchmarks~\cite{sun2024instructionfollowingevaluatinginferential,han2024folionaturallanguagereasoning,proverqa} typically provide the relevant rules directly as problem-specific premises in each reasoning instance.
This design reduces reusable procedural rules to per-question hints and thus ignores the shared nature of rule knowledge across QA instances.
As a result, they obscure whether a model can (i) identify the correct reusable rules from a large, shared rule repository for a query and (ii) apply them reliably in multi-step reasoning. 

\begin{figure}[t]
    \centerline{\includegraphics[scale=0.294]{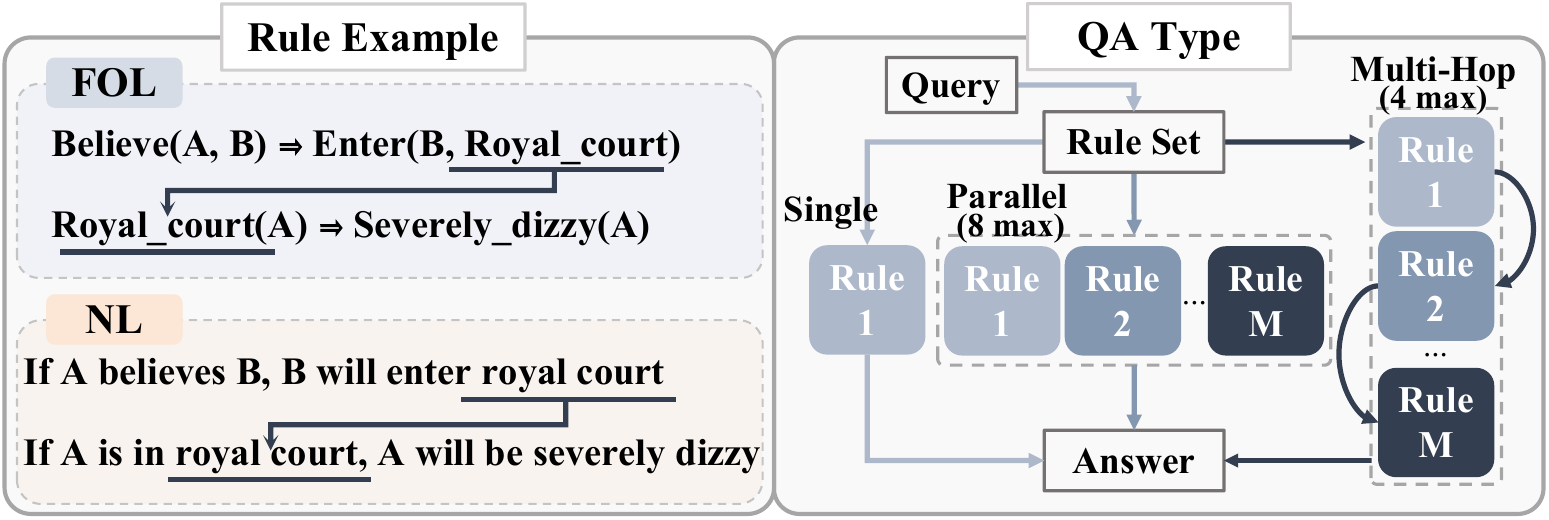}}%
    \caption{Illustration of rule examples and RuleWorld QA types. Rules from different types may interact, and the benchmark includes three QA types: single-rule, parallel multi-rule, and multi-hop reasoning.}
    \label{fig:rulebench}
\end{figure}

To specifically evaluate whether LLMs could correctly localize and apply rules from large-scale injected rule repository according to the given query, this paper constructs \textbf{RuleWorld}, a benchmark designed to measure the rule utilization capability of LLMs. RuleWorld comprises 4.94 million externally provided procedural rules that are abstract, non-commonsense, and entity-independent, which prevents models from relying on internalized knowledge.
For instance, a rule like ``if $X$ is happy, then $X$ will go to the forest'' leaves $X$ unspecified, requiring the model to ground the placeholder to the query before applying the rule.
Consequently, strong performance requires selecting a small subset of relevant rules from a large rule set and executing them correctly during reasoning. 
In RuleWorld, the rules are organized into four major types (\textit{Attributes}, \textit{Environments}, \textit{Actions}, and \textit{States}) with seven subtypes, available in both first-order logic (FOL) and natural-language (NL) forms. 
Crucially, RuleWorld is compositional by design: a rule’s conclusion can become the premise of other rules, yielding cross-type interactions (e.g., an \textit{Action} can change the \textit{Environment} and cause downstream changes; Figure~\ref{fig:rulebench}, left) that pose realistic challenges for rule localization and utilization during reasoning.
To systematically assess these capabilities, we define three QA types that capture distinct modes of rule utilization (Figure~\ref{fig:rulebench}, right).
\textit{Single-Rule QA} requires applying exactly one rule to answer a question, providing a basic test of rule grounding and application.
\textit{Parallel Multi-Rule QA} involves several sub-questions in one instance. Each sub-question is an independent, non-sequential step that may require multiple rules. The model solves these steps separately, then aggregates the sub-answers into a combined final answer (up to eight gold rules per instance).
\textit{Multi-Hop Rule QA} requires chaining reasoning steps, where each hop may involve applying multiple rules, and earlier conclusions become later premises, requiring the model to track intermediate states across the chain (up to four hops).
Together, they form eleven sub-tasks by varying the number of gold rules and the number of hops, progressively increasing the difficulty of rule localization and application.

Through detailed evaluations on RuleWorld, we find that there are two major challenges for current LLMs to effectively leverage large injected rule sets.
(i) Injected rules are highly abstract and thus weakly match question surface forms, making both rule retrieval and utilization brittle at scale.
(ii) The ability of step-wise rule utilization is required under both parallel aggregation and multi-hop state tracking, where the target rules change across steps, making retrieval and reasoning unstable.
Existing methods such as retrieval-augmented generation (RAG)~\cite{lewis2020retrieval-augmented-generation} are sensitive to retrieval quality and prone to semantic mismatch, while internal injection~\cite{wangkblam,yu2025srkiscalablerealtimeknowledge} often exhibits unstable rule selection at scale and degrades substantially in multi-step reasoning.

To this end, we introduce \textbf{DynaRule}, which conducts dynamic rule selection and updating entirely within the model’s internal space. Inspired by recent work that treats attention as key-value association~\cite{wangkblam,yu2025srkiscalablerealtimeknowledge}, external rules are encoded with a pretrained sentence encoder, aligned via single-layer adapters, and directly injected into the LLMs' KV cache.
In specific, DynaRule is trained in two stages. 
First, we identify the model’s \textit{confidence layer} as the layer at which attention entropy over injected rules is minimized, indicating the most decisive focus on a small set of relevant rules while suppressing irrelevant ones.
Second, we perform \textit{Stacked Step-Level Attention Training} on this layer by introducing a special \texttt{<search>} token, which guides attention to the rules required at each reasoning step and turns retrieval into an internal, learnable process.
During inference, emitting \texttt{<search>} triggers rule re-attention and re-retrieval, replacing outdated rule entries in the KV cache with newly selected ones. 
This design internally aligns abstract rule selection with step-wise decoding via \texttt{<search>}-triggered re-attention, stabilizing retrieval and reasoning as the target rules change across steps.

Comprehensive experiments demonstrate that RuleWorld poses substantial challenges to existing LLMs (e.g., DeepSeek V3.2~\cite{deepseekai2025deepseekv32pushingfrontieropen}) and GPT-5.5~\cite{openai2026gpt55}. DynaRule consistently outperforms all baselines across diverse rule injection settings, improving average QA accuracy by up to 19 points and achieving gains of up to 48 points on specific sub-tasks. It also achieves over 90\% Recall@10 and over 85\% Recall@1 at 10K rules, surpassing the strongest baseline by more than 60 points. 
These results show that step-level dynamic integration can identify the relevant procedural rules at each step and reliably leverage them during reasoning.

Our contributions are two-fold:
\begin{itemize}
\item We introduce \textbf{RuleWorld}, a large-scale benchmark containing millions of abstract procedural rules in FOL and NL, covering four types, seven sub-types, and eleven QA sub-tasks for evaluating rule-application ability.
\item We propose \textbf{DynaRule}, an end-to-end rule integration framework with learnable retrieval that injects external rules into the KV cache and uses \texttt{<search>}-driven step-wise attention to select and update rules, yielding large gains in QA and retrieval accuracy and enabling more reliable procedural reasoning.
\end{itemize}

\section{Related Work}

\begin{table*}[t]
\centering

\begin{subfigure}[t]{0.7\textwidth}
\vspace{0pt}
\centering
\small
\setlength{\tabcolsep}{8pt}
\resizebox{\textwidth}{!}{%
\begin{tabular}{lccccc}
\toprule
Benchmark 
& \makecell[c]{FOL\\Rule}
& \makecell[c]{NL\\Rule}
& \makecell[c]{Faithful Natural\\Language Reasoning\\Chain}
& \makecell[c]{Controlled\\Composition\\Taxonomy}
& \makecell[c]{Large Shared\\Rule Pool\\with Localization} \\
\midrule
RuleTaker~\cite{clark2020transformerssoftreasonerslanguage}      & \xmark & \cmark & \xmark & \pmark & \xmark \\
ProofWriter~\cite{tafjord2021proofwritergeneratingimplicationsproofs}    & \xmark & \cmark & \cmark & \pmark & \xmark \\
ProntoQA~\cite{saparov2023languagemodelsgreedyreasoners}       & \cmark & \cmark & \cmark & \pmark & \xmark \\
FOLIO~\cite{han2024folionaturallanguagereasoning}          & \cmark & \cmark & \xmark & \pmark & \xmark \\
LogicBench~\cite{parmar2024logicbenchsystematicevaluationlogical}     & \pmark & \cmark & \xmark & \cmark & \xmark \\
Multi-LogiEval~\cite{patel2024multilogievalevaluatingmultisteplogical} & \pmark & \cmark & \xmark & \cmark & \xmark \\
RuleArena~\cite{zhou2025rulearenabenchmarkruleguidedreasoning}      & \xmark & \cmark & \pmark & \pmark & \cmark \\
ProverQA~\cite{proverqa}       & \cmark & \cmark & \cmark & \cmark & \xmark \\
\midrule
\textbf{RuleWorld (Ours)}      & \cmark & \cmark & \cmark & \cmark & \cmark \\
\bottomrule
\end{tabular}%
}
\label{tab:rule_benchmark_comparison}
\end{subfigure}
\hspace{0.01\textwidth}
\begin{subfigure}[t]{0.25\textwidth}
\vspace{-8pt}
\centering
\includegraphics[width=\textwidth]{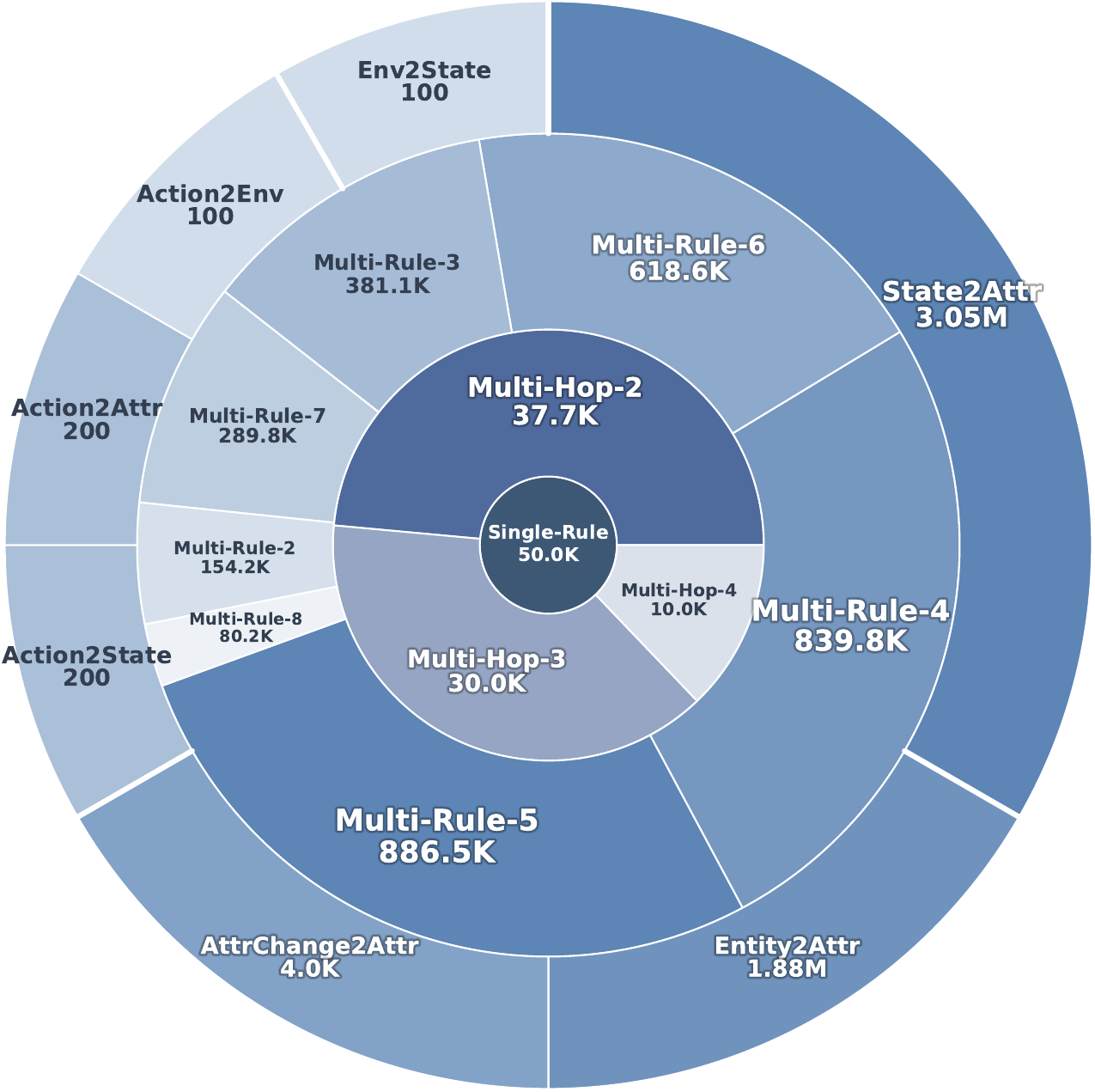}
\label{fig:qa_type_statistics}
\end{subfigure}
\vspace{-10pt}
\caption{\textbf{Left}: Comparison of rule-guided reasoning benchmarks across five dimensions. 
\protect\cmark{} = Yes, \protect\xmark{} = No, \protect\pmark{} = Partial. \textbf{Right}: Distribution of 3.37M QA instances over 11 sub-tasks; the outer ring shows seven rule sub-types, with thick separators marking major types. Suffixes denote gold rule counts or hops.}
\label{fig:table_and_qa_stats}
\vspace{-0.5em}
\end{table*}

\paragraph{Benchmarks for Logical Reasoning}
Existing logical reasoning benchmarks usually model rules as per-instance premises rather than as a reusable and globally consistent rule system~\cite{clark2020transformerssoftreasonerslanguage,tafjord2021proofwritergeneratingimplicationsproofs,nakamura-etal-2023-logicattack,han2024folionaturallanguagereasoning,parmar2024logicbenchsystematicevaluationlogical,patel2024multilogievalevaluatingmultisteplogical,zhou2025rulearenabenchmarkruleguidedreasoning,proverqa}. As shown in Table~\ref{fig:table_and_qa_stats} (left), prior benchmarks each cover only part of the space, such as FOL or NL rules, faithful reasoning chains, or controlled composition, but they generally do not combine large rule pool localization with shared rule knowledge across instances. This limits diagnosis of whether models truly retrieve and apply reusable rules, especially beyond aggregate accuracy in true/false or multiple choice settings~\cite{tian-etal-2021-diagnosing,logicQA,parmar2024logicbenchsystematicevaluationlogical,patel2024multilogievalevaluatingmultisteplogical,zhou2025rulearenabenchmarkruleguidedreasoning,proverqa}. RuleWorld addresses this gap by introducing a contradiction free shared rule set and controlled evaluation of single-rule, parallel multi-rule, and multi-hop reasoning.

\paragraph{Methods for Rule Reasoning}
Existing methods for rule reasoning with LLMs mainly fall into three categories. (1) Full fine-tuning internalizes reasoning patterns through large synthetic corpora~\cite{xu2024symbol-llm,morishita2024enhancing-synthetic-logic-corpus,jiang2025logicpro}, but is costly and does not explicitly model globally reusable rules. (2) Prompting-based approaches decompose reasoning into multiple stages such as planning, rewriting, or solver based verification~\cite{pan2023logic-lm-solvers,sun-etal-2024-determlr,servantez2024chain-of-logic}, yet often rely on strong closed-source models or external tools and remain brittle under difficult rule localization. (3) Retrieval based approaches inject rules at inference time, but both standard RAG~\cite{lewis2020retrieval-augmented-generation} and recent KV cache based methods~\cite{wangkblam,yu2025srkiscalablerealtimeknowledge} still struggle with reliable step-level rule selection. DynaRule addresses this gap by jointly learning step-level retrieval and reasoning in an end-to-end manner.

\section{Preliminaries}
\paragraph{Rule Formulation}
Rules follow a causal structure in which the \textit{conclusion} holds only when the \textit{premise} is met~\cite{brachman2004knowledge}. 
Each rule is represented as a pair $(\mathbf{p}, \mathbf{c})$, with $\mathbf{p}$ defined as the premise and $\mathbf{c}$ as the conclusion. A rule set with $M$ rules is written as $\mathcal{R} = \left\{ \left(\mathbf{p}_{m}, \mathbf{c}_{m} \right) \right\}_{m=1}^{M}$.

\paragraph{Rule Knowledge Injection via KV Cache}
Inspired by~\cite{wangkblam,yu2025srkiscalablerealtimeknowledge}, we treat attention as a normalized key–value matching mechanism. Under this view, each rule $(\mathbf{p}_{m}, \mathbf{c}_{m})$ is represented as a key–value pair, where the full rule $(\mathbf{p}_{m}, \mathbf{c}_{m})$ acts as the key and the conclusion $\mathbf{c}_{m}$ as the value. Specifically, we encode each rule using a pretrained sentence encoder to obtain $(\mathbf{k}_m, \mathbf{v}_m)=\textproc{Encode}((\mathbf{p}_m,\mathbf{c}_m),\mathbf{c}_m)$, and project them from the encoder dimension $P$ into the model embedding space $D$ via single-linear adapters at layer $l$: $(\tilde{\mathbf{k}}^{l}_m,\tilde{\mathbf{v}}^{l}_m)=(\mathbf{k}_m\tilde{\mathbf{W}}^{l}_{K},\mathbf{v}_m\tilde{\mathbf{W}}^{l}_{V})$, where $\tilde{\mathbf{W}}^{l}_{K},\tilde{\mathbf{W}}^{l}_{V}\in\mathbb{R}^{P\times D}$. This design allows each rule instance to be uniquely identified, even when multiple rules share identical premises. Trained over large-scale rule corpora, layer-specific adapters learn robust mappings that enable rule embeddings to be smoothly integrated into the attention mechanism via rectangular attention. 
At each layer, the projected rule representations with $M$ entries are directly inserted into the corresponding KV cache, which contains $N$ contextual key-value pairs $\mathbf{K}^l,\mathbf{V}^l\in\mathbb{R}^{N\times D}$, resulting in expanded representations $\hat{\mathbf{K}}^{l},\hat{\mathbf{V}}^{l}\in\mathbb{R}^{(M+N)\times D}$. Formally:
\begin{equation} 
    \scalebox{1.1}{$ 
        \hat{\mathbf{K}}^{l} = \big[\, \tilde{\mathbf{K}}^{l} \;\; \mathbf{K}^{l} \,\big], \quad \hat{\mathbf{V}}^{l} = \big[\, \tilde{\mathbf{V}}^{l} \;\; \mathbf{V}^{l} \,\big],
    $} 
\end{equation}
where $\tilde{\mathbf{K}}^{l} = [\tilde{\mathbf{k}}^{l}_1,\ldots,\tilde{\mathbf{k}}^{l}_M],
\tilde{\mathbf{V}}^{l} = [\tilde{\mathbf{v}}^{l}_1,\ldots,\tilde{\mathbf{v}}^{l}_M]$.
To obtain auxiliary queries, we use a dedicated projection adapter $\tilde{\mathbf{W}}^{l}_{Q}$ that transforms the hidden states $\mathbf{X}^l$ into a new query matrix $\tilde{\mathbf{Q}}^{l} = \mathbf{X}^l \tilde{\mathbf{W}}^{l}_{Q} \in \mathbb{R}^{N \times D}$. In contrast, the original query matrix $\mathbf{Q}^l$ is retained for standard contextual self-attention. Since only the key–value sequence length is modified, the output dimensionality remains identical to standard self-attention, allowing rule information to be seamlessly integrated into the hidden states. The resulting attention computation is:
\begin{equation}
\scalebox{0.75}{$
\text{Attention}
=
\text{Softmax}\!\left(
\left[
\dfrac{\tilde{\mathbf{Q}}^{l}\left(\tilde{\mathbf{K}}^{l}\right)^{\!\top}}{\sqrt{D}}
\;\middle|\;
\dfrac{\mathbf{Q}^{l}\left(\mathbf{K}^{l}\right)^{\!\top}}{\sqrt{D}}
\right]
\right)
\,\hat{\mathbf{V}}^{l}.
$}
\end{equation}

\section{RuleWorld Benchmark}
RuleWorld is designed to evaluate whether LLMs can localize and apply reusable procedural rules from a large shared repository, yielding a large-scale, rule-centric QA benchmark with 3.37M instances, as shown in Table~\ref{fig:table_and_qa_stats} (right).
RuleWorld comprises 4.94 million rules in both FOL and NL forms that are globally consistent, abstract, intentionally non-commonsense (e.g., ``\textit{entering a forest sets an entity on fire}''), and conflict-free, making the same rules reusable across all QA instances without relying on prior knowledge. 
The rules are organized into four major types and seven sub-types, where the conclusion of one rule can become the premise of another, enabling cross-rule interactions.
Detailed data statistics, the full rule and QA construction pipeline, and complete examples are provided in Appendices~\ref{appendix:RuleWorld-Data-Statistics},~\ref{appendix:RuleWorld-Construction-Process}, and~\ref{appendix:RuleWorld-Data-Samples}, respectively. Moreover, we categorize RuleWorld into three task types, each targeting a distinct dimension of rule application:

\textbf{(1) Single-Rule QA (50K instances)}.
Each instance requires applying exactly one rule to derive the correct answer. This setting serves as the simplest form of rule grounding and application.

\textbf{(2) Parallel Multi-Rule QA (3.25M instances)}.
Each instance contains several parallel sub-questions, which the model answers independently and then combines the resulting conclusions into a final answer. Difficulty scales with the number of gold rules, up to 8 per instance.

\textbf{(3) Multi-Hop Rule QA (77.7K instances)}. 
Each instance requires sequential reasoning, where the conclusion of one step serves as the premise for the next, necessitating tracking of intermediate states across the chain.
Difficulty scales with the number of hops, up to 4 per instance.

Together, these tasks systematically evaluate LLMs' ability to localize and apply rules across 11 sub-tasks by varying gold-rule and hop counts.

\begin{figure*}[htbp]
    \centerline{\includegraphics[scale=0.478]{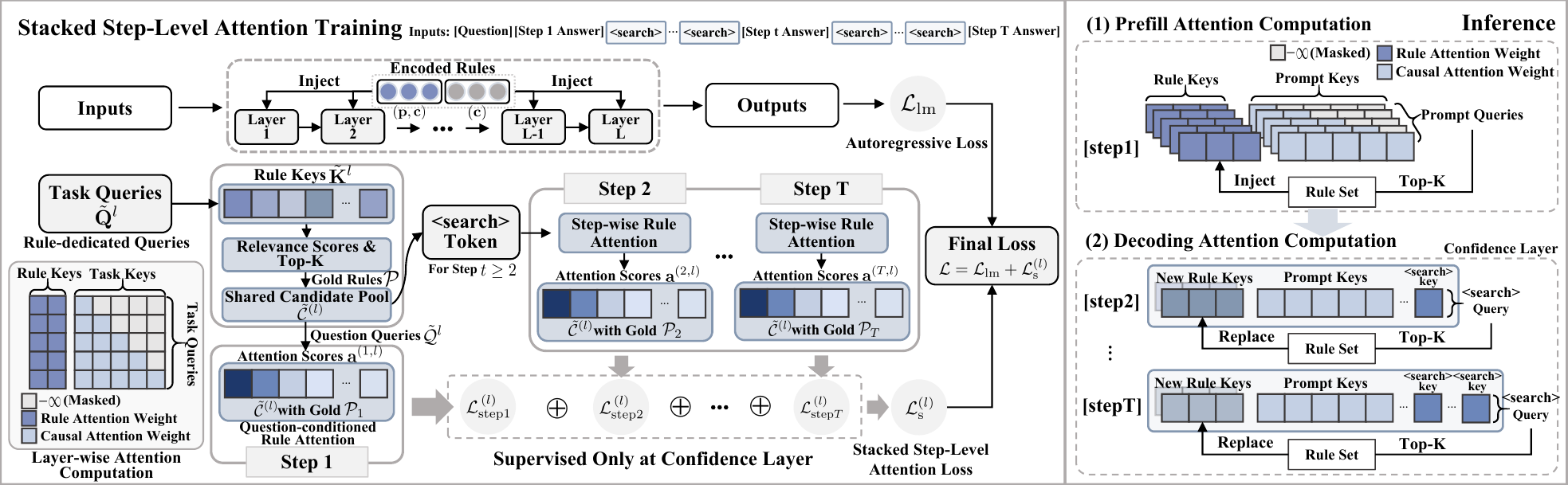}}%
    \caption{Overview of Stacked Step-Level Attention Training (\textbf{left}) and dynamic rule attention at inference (\textbf{right}). Training builds a shared candidate pool $\tilde{\mathcal{C}}^{(l)}$ from sequence-level relevance between $\tilde{\mathbf{Q}}^{l}$ and injected rule keys $\tilde{\mathbf{K}}^{l}$, and learns step-level retrieval with question and \texttt{<search>} tokens. At inference, top-$K$ rules are selected at prefill and updated through re-attention triggered by \texttt{<search>} tokens. Values are omitted for clarity.}
    \label{fig:main}
\vspace{-0.5em}
\end{figure*}

\section{DynaRule}
\subsection{Training Process}
We adopt a two-stage scheme: (1) identify the model’s \textit{confidence layer} via layer-wise attention entropy; (2) introduce a \texttt{<search>} token and train this layer with a Stacked Step-Level Attention Loss for step-wise rule retrieval (Figure~\ref{fig:main}, left). Rule attention uses queries projected by dedicated $\tilde{\mathbf{W}}^{l}_{Q}$. We freeze the base model in both stages and update $\tilde{\mathbf{W}}^{l}_{Q}$, $\tilde{\mathbf{W}}^{l}_{K}$, and $\tilde{\mathbf{W}}^{l}_{V}$, additionally unfreezing the embedding layer in Stage 2 to learn \texttt{<search>}.
\paragraph{Confidence Layer Identification}
Not all layers are equally informative for rule selection when external rule knowledge is injected into the KV cache. 
In particular, we observe that certain layers exhibit significantly sharper attention distributions over rule keys, indicating that rule relevance is most clearly differentiated at these depths. To quantify this effect, let the softmax-normalized attention weights over the $M$ injected rule keys at layer $l$, given a query $q$, be $\{\alpha^{l}_m(q)\}_{m=1}^{M}$, and define the corresponding attention entropy as
$H^{l}(q) = - \sum_{m=1}^{M} \alpha^{l}_{m}(q) \log \alpha^{l}_{m}(q)$.
We identify the \textit{confidence layer} as the layer with the lowest average attention entropy across the query set, i.e.,
$\arg\min_{l}\mathbb{E}_{q \sim \mathcal{D}_{q}}\!\left[H^{l}(q)\right]$,
as this layer concentrates attention on a small subset of relevant rules and provides the most decisive and stable rule relevance signals for guiding step-level rule retrieval.
\paragraph{Stacked Step-Level Attention Training Objective}
We supervise the \textit{confidence layer}’s rule attention to enable efficient step-wise rule retrieval at inference.
During training, we use the full teacher-forced sequence (question plus answer), with the answer divided into $T$ reasoning steps. We treat step $t=1$ as purely question-conditioned retrieval, since the question fully specifies the retrieval intent. For $t\ge2$, we insert a \texttt{<search>} token before the step-$t$ answer content, and use it as an explicit retrieval trigger to support step-wise retrieval conditioned on intermediate reasoning states.
To align training with inference-time per-layer pruning and to expose the \textit{confidence layer} to hard candidates, we first construct a stable Top-$K$ candidate set at each layer.
Specifically, at each layer $l$, we compute a \emph{sequence-level} relevance score for each rule by averaging full-sequence token queries
$\tilde{\mathbf{Q}}^{l}=\{\tilde{\mathbf{q}}^{l}_{n}\}_{n=1}^{N}$
against injected rule keys
$\tilde{\mathbf{K}}^{l}=\{\tilde{\mathbf{k}}^{l}_{m}\}_{m=1}^{M}$,
i.e.,
$\mathbf{s}^{(l)}_{m}=\frac{1}{N}\sum_{n=1}^{N}\big(\tilde{\mathbf{q}}^{l}_{n}\big)^{\!\top}\tilde{\mathbf{k}}^{\,l}_{m}$,
and retain the Top-$K$ index set as $\mathcal{C}^{\text{top},(l)}$.
This token-averaged scoring makes Top-$K$ selection less sensitive to step boundaries, yielding a stable candidate set under batched training, while explicit supervision is only applied at the \textit{confidence layer}.
\begin{equation}
\label{eq:topk_select}
\scalebox{0.695}{$
\tilde{\mathcal{C}}^{(l)}
=
\textproc{KeepDim}\!\left(
\underbrace{\textproc{TopK}\big( \{ \mathbf{s}^{(l)}_{m} \}_{m=1}^{M}, K \big)}_{\mathcal{C}^{\text{top},(l)}}
\;\cup\;
\Big( \bigcup_{t=1}^{T} \mathcal{P}_{t} \Big),
\; K
\right)
$}
\end{equation}

Let $\mathcal{P}_t$ denote the gold rule indices required at each step $t$. As shown in Eq.~(\ref{eq:topk_select}), at the \textit{confidence layer} we augment the Top-$K$ set with the gold indices from all steps, i.e., $\bigcup_{t=1}^{T}\mathcal{P}_{t}$, while removing an equal number of other indices to maintain dimensional consistency via a $\textproc{KeepDim}(\cdot)$ operation, yielding a shared candidate index pool $\tilde{\mathcal{C}}^{(l)}$ across steps.
This prevents pruning away future-step gold indices and enables cross-step discrimination: at step $t$, rules indexed by $\mathcal{P}_{t}$ are positives, while the remaining rules in $\tilde{\mathcal{C}}^{(l)}$, including gold rules from other steps, act as hard negatives.
\begin{equation}
\label{eq:attention_score}
\scalebox{0.9}{$
\mathbf{a}^{(t,l)}_{m}
=
\begin{cases}
\displaystyle
\frac{1}{\sqrt{D}}\frac{1}{|\tilde{\mathcal{Q}}^l|}
\sum_{p \in \tilde{\mathcal{Q}}^l}
\big( \tilde{\mathbf{q}}^{l}_{p} \big)^{\!\top} \tilde{\mathbf{k}}^{\,l}_{m},
& t = 1,
\\[12pt]
\displaystyle
\frac{1}{\sqrt{D}}
\big( \tilde{\mathbf{q}}^{l}_{\texttt{<search>},t} \big)^{\!\top} \tilde{\mathbf{k}}^{\,l}_{m},
& t \ge 2.
\end{cases}
$}
\end{equation}

We then apply Eq.~(\ref{eq:attention_score}) to each rule $m \in \tilde{\mathcal{C}}^{(l)}$ to compute an attention score $\mathbf{a}_m^{(t,l)}$, which serves as the retrieval signal for step-conditioned rule selection.
For step $t=1$, retrieval is question-conditioned; thus we compute $\mathbf{a}_m^{(1,l)}$ by averaging dot products between the rule key $\tilde{\mathbf{k}}^l_m$ and the question-token queries indexed by $\tilde{\mathcal{Q}}^l$.
For subsequent steps $t \ge 2$, retrieval intent depends on intermediate reasoning states rather than the static question; therefore we use the query state of the step-specific \texttt{<search>} token inserted before the answer at step $t$, denoted as $\tilde{\mathbf{q}}^{l}_{\texttt{<search>},t}$, to form a step-adaptive query for rule identification.
Since single-step QA requires only one step, no \texttt{<search>} token is involved.
\begin{equation}
\label{eq:training_loss}
\scalebox{0.8}{$
    \begin{aligned}
        \mathcal{L}
        &=
        \underbrace{
        \left(
        - \frac{1}{N}
        \sum_{n=1}^{N}
        \log p_\theta\big( x_{n} \mid x_{<n} \big)
        \right)
        }_{\mathcal{L}_\text{lm}\text{: Autoregressive Language Modeling Loss}}
        \\[4pt]
        &\qquad
        +\;
        \underbrace{
        \left(
        \scalebox{0.85}{$
        - \frac{1}{\displaystyle\sum_{t=1}^{T} |\mathcal{P}_{t}|}
        \displaystyle\sum_{t=1}^{T}
        \displaystyle\sum_{i \in \mathcal{P}_{t}}
        \log
        \frac{
            \exp\big( \mathbf{a}^{(t,l)}_{i} / \mathcal{T} \big)
        }{
            \displaystyle\sum_{j \in \tilde{\mathcal{C}}^{(l)} \setminus (\mathcal{P}_t \setminus \{i\})}
            \mkern-32mu \exp\!\big( \mathbf{a}^{(t,l)}_{j} / \mathcal{T} \big)
        }
        $}
        \right)
        }_{\mathcal{L}^{(l)}_\text{s}\text{: Stacked Step-Level Attention Loss at layer $l$}}
        \end{aligned}
    $}
\end{equation}

Next, for each step, we apply a cross-entropy loss per positive rule over the candidate pool $\tilde{\mathcal{C}}^{(l)}$, where other positives at the same step are masked out, encouraging higher attention on the corresponding rule against hard negatives. A temperature coefficient $\mathcal{T}$ is introduced to sharpen the contrast between scores, amplifying differences in rule relevance. Finally, the losses from all steps are stacked, forming a Stacked Step-Level Attention Loss $\mathcal{L}^{(l)}_\text{s}$ that jointly supervises correct rule usage across the entire reasoning chain. This formulation transforms the retrieval of external abstract rules into an internal, learnable, and fully differentiable process. The final loss, as shown in Eq.~(\ref{eq:training_loss}), is $\mathcal{L}=\mathcal{L}_\text{lm} + \mathcal{L}^{(l)}_\text{s}$, where $\mathcal{L}_\text{lm}$ is the standard autoregressive language modeling loss.

\subsection{Inference Process}
During inference, the model performs rule retrieval, updating, and reasoning in a fully end-to-end manner within its internal intent space (Figure~\ref{fig:main}, right). Inference consists of a prefill stage followed by decoding. In the prefill stage, attention over injected rule keys and Top-$K$ selection are computed at all layers, while explicit rule selection and KV cache updating are carried out only at the \textit{confidence layer}. The model has been trained to emit the special \texttt{<search>} token at the points where additional rule retrieval is required, since multi-step training answers include supervised \texttt{<search>} insertions. Accordingly, if the question requires only one step, decoding proceeds directly to the final answer without emitting \texttt{<search>}. For multi-step cases, whenever the decoder outputs \texttt{<search>}, the hidden state of this token is used as a new query to recompute attention over the rule set at the \textit{confidence layer}, and the newly selected rules replace the previous rules in the KV cache, enabling step accurate grounding throughout the reasoning chain.

\section{Experiments}
\subsection{Experiment Settings}
We provide the detailed training and evaluation implementation settings in Appendix~\ref{appendix:Training_and_Evaluation_Settings}.
\paragraph{Model Selection}
We adopt Qwen2.5-7B-Instruct~\cite{qwen2025qwen25technicalreport} as the base model and Qwen3-Embedding-8B~\cite{yang2025qwen3technicalreport} as the text encoder. To analyze the stability of the confidence layer across model architectures and scales, we further evaluate Llama-3-8B-Instruct~\cite{dubey2024llama}, smaller Qwen2.5 models (1.5B and 3B)~\cite{qwen2025qwen25technicalreport}, and bge-m3~\cite{chen2024bge}. 
For strong prompting baselines, we include larger Qwen2.5 models (14B, 32B, 72B)~\cite{qwen2025qwen25technicalreport}, Qwen3-32B~\cite{yang2025qwen3technicalreport}, Llama3 variants (Llama-3.1-70B-Instruct, Llama-3.3-70B-Instruct)~\cite{dubey2024llama}, deepseek-chat (DeepSeek V3.2)~\cite{deepseekai2025deepseekv32pushingfrontieropen}, gpt-5.1-2025-11-13~\cite{openai2025systemcard}, gpt-5.5-2026-04-23~\cite{openai2026gpt55}, and claude-sonnet-4-6~\cite{anthropic2026sonnet46}.

\begin{table*}[t]
  \centering
  \small
  \setlength{\tabcolsep}{8.8pt}
  \scalebox{0.7}{%
  \begin{tabular}{c *{6}{c} *{6}{c} c}
  \toprule
  \multirow{2}{*}{Method}
    & \multicolumn{6}{c}{\textbf{First-order Logic}}
    & \multicolumn{6}{c}{\textbf{Natural Language}}
    & \multirow{2}{*}{\textbf{Avg.}} \\
  \cmidrule(lr){2-7}
  \cmidrule(lr){8-13}
    & Single & PM (2--4) & PM (5--8) & MH-2 & MH-3 & MH-4
    & Single & PM (2--4) & PM (5--8) & MH-2 & MH-3 & MH-4
    & \\
  \midrule

  \multicolumn{14}{c}{\textit{Rule Num = 100}} \\ \midrule
  Prompting & 0.8200 & 0.7389 & 0.4917 & 0.5000 & 0.0600 & 0.0400 & \underline{0.8200} & 0.8194 & 0.5684 & 0.3400 & 0.1000 & 0.1200 & 0.4515 \\
  KBLaM & \underline{0.9800} & \textbf{0.9939} & \textbf{0.9871} & \textbf{1.0000} & 0.9000 & \underline{0.9400} & \textbf{1.0000} & \underline{0.9917} & \textbf{0.9863} & \textbf{0.9600} & 0.4800 & \textbf{0.7200} & \underline{0.9116} \\
  SR-KI & \textbf{1.0000} & 0.9883 & 0.9750 & \underline{0.9800} & \underline{0.9200} & 0.9000 & \textbf{1.0000} & 0.9750 & 0.9583 & \textbf{0.9600} & \underline{0.5400} & 0.5600 & 0.8964 \\
  \rowcolor{blue!7} DynaRule & \textbf{1.0000} & \underline{0.9917} & \underline{0.9863} & 0.9600 & \textbf{0.9300} & \textbf{0.9500} & \textbf{1.0000} & \textbf{0.9967} & \underline{0.9825} & \underline{0.9200} & \textbf{0.9600} & \underline{0.7000} & \textbf{0.9481} \\
  \midrule

  \multicolumn{14}{c}{\textit{Rule Num = 1000}} \\ \midrule
  Prompting & 0.8000 & 0.6211 & 0.2654 & 0.2800 & 0.0400 & 0.0000 & 0.6200 & 0.6472 & 0.2792 & 0.0600 & 0.0200 & 0.0000 & 0.3027 \\
  $\text{RAG}_{\text{dense}}$ & 0.6600 & 0.3589 & 0.2029 & 0.0400 & 0.0200 & 0.0200 & \underline{0.8800} & 0.5750 & 0.2383 & 0.0200 & 0.0000 & 0.0000 & 0.2513 \\
  $\text{RAG}_{\text{bm25}}$ & 0.5200 & 0.5483 & 0.2825 & 0.0000 & 0.0000 & 0.0000 & 0.5800 & 0.3772 & 0.2379 & 0.0000 & 0.0000 & 0.0000 & 0.2122 \\
  $\text{RAG}_{\text{hybrid}}$ & 0.6200 & 0.6044 & 0.3429 & 0.0000 & 0.0000 & 0.0200 & 0.8200 & 0.5300 & 0.3154 & 0.1600 & 0.0000 & 0.0200 & 0.2861 \\
  KBLaM & 0.9000 & 0.8639 & \underline{0.8888} & \underline{0.8400} & \underline{0.4200} & \underline{0.4400} & 0.7800 & 0.8272 & \underline{0.8729} & \underline{0.6800} & \underline{0.3400} & \underline{0.4200} & \underline{0.6894} \\
  SR-KI & \underline{0.9800} & \underline{0.9483} & 0.8779 & 0.7800 & 0.2200 & 0.2800 & \textbf{0.9800} & \underline{0.8767} & 0.8263 & 0.5800 & 0.1400 & 0.0400 & 0.6274 \\
  \rowcolor{blue!7} DynaRule & \textbf{1.0000} & \textbf{0.9931} & \textbf{0.9729} & \textbf{0.8500} & \textbf{0.6250} & \textbf{0.6500} & \textbf{0.9800} & \textbf{0.9833} & \textbf{0.9642} & \textbf{0.8400} & \textbf{0.6400} & \textbf{0.6800} & \textbf{0.8482} \\
  \midrule
  \multicolumn{14}{c}{\textit{Rule Num = 10000}} \\ \midrule
  $\text{RAG}_{\text{dense}}$ & 0.4000 & 0.2411 & 0.1088 & 0.0000 & 0.0400 & 0.0000 & 0.6600 & 0.2744 & 0.0842 & 0.0000 & 0.0000 & \underline{0.1000} & 0.1590 \\
  $\text{RAG}_{\text{bm25}}$ & 0.3400 & 0.1983 & 0.0950 & 0.0200 & 0.0000 & 0.0000 & 0.6400 & 0.2817 & 0.1638 & 0.0000 & 0.0000 & 0.0400 & 0.1482 \\
  $\text{RAG}_{\text{hybrid}}$ & 0.4400 & 0.4022 & 0.1800 & 0.0000 & 0.0200 & 0.0000 & 0.6000 & 0.4072 & 0.2158 & 0.0200 & 0.0000 & 0.0400 & 0.1938 \\
  KBLaM & 0.0000 & 0.0000 & 0.0000 & 0.0000 & 0.0000 & 0.0000 & 0.0000 & 0.0000 & 0.0000 & 0.0000 & 0.0000 & 0.0000 & 0.0000 \\
  SR-KI & \underline{0.7400} & \underline{0.6194} & \underline{0.4946} & \underline{0.3800} & \textbf{0.2000} & \underline{0.1200} & \underline{0.7400} & \underline{0.5133} & \underline{0.4071} & \underline{0.1200} & \underline{0.0800} & 0.0400 & \underline{0.3712} \\
  \rowcolor{blue!7} DynaRule & \textbf{0.9200} & \textbf{0.9211} & \textbf{0.7904} & \textbf{0.5200} & \underline{0.1600} & \textbf{0.1400} & \textbf{0.8000} & \textbf{0.8600} & \textbf{0.7288} & \textbf{0.6000} & \textbf{0.1600} & \textbf{0.2400} & \textbf{0.5700} \\
  \bottomrule
  \end{tabular}%
  }
  \caption{Exact match accuracy on RuleWorld with Qwen2.5-7B-Instruct. \textbf{Single}: Single-Rule QA. \textbf{PM}: Parallel Multi-Rule QA, where columns 2--4 and 5--8 report mean exact match over the corresponding rule counts. \textbf{MH}: Multi-Hop Rule QA, where the numeric suffix denotes hop count. $\text{RAG}_{\text{dense}}$ and $\text{RAG}_{\text{bm25}}$ use Qwen3-Embedding-8B and BM25 retrievers, respectively; $\text{RAG}_{\text{hybrid}}$ uses RRF~\cite{Reciprocal_rank_fusion_outperforms_condorcet_and_individual_rank_learning_methods} with a fusion constant of 60. Detailed results for additional rule sizes are reported in Tables~\ref{tab:FOl_main_qa_results} and~\ref{tab:NL_main_qa_results}.}
  \label{tab:fol_nl_main_qa_results}
\end{table*}

\paragraph{Baselines} We compare our approach against the following representative methods:

\textbf{(1) Prompting (full-context rule injection)}. All rules are prepended to the prompt, with GPU memory limiting injection to 2000 rules for the 7B model and 1000 for larger models.

\textbf{(2) RAG}~\cite{lewis2020retrieval-augmented-generation}. An external retriever selects top-$K$ rules, which are prepended to the prompt. Retrieval is performed once before generation, with no updates during reasoning.

\textbf{(3) KBLaM}~\cite{wangkblam}. A state-of-the-art end-to-end method that integrates external knowledge by projecting key–value pairs into the model’s internal KV cache space.

\textbf{(4) SR-KI}~\cite{yu2025srkiscalablerealtimeknowledge}. A novel end-to-end method that supervises retrieval through an attention-based retrieval loss in the KV cache, but does not explicitly decompose multi-step reasoning, leading to a single-step retrieval signal.

\begin{figure}[htbp]
    \centerline{\includegraphics[scale=0.155]{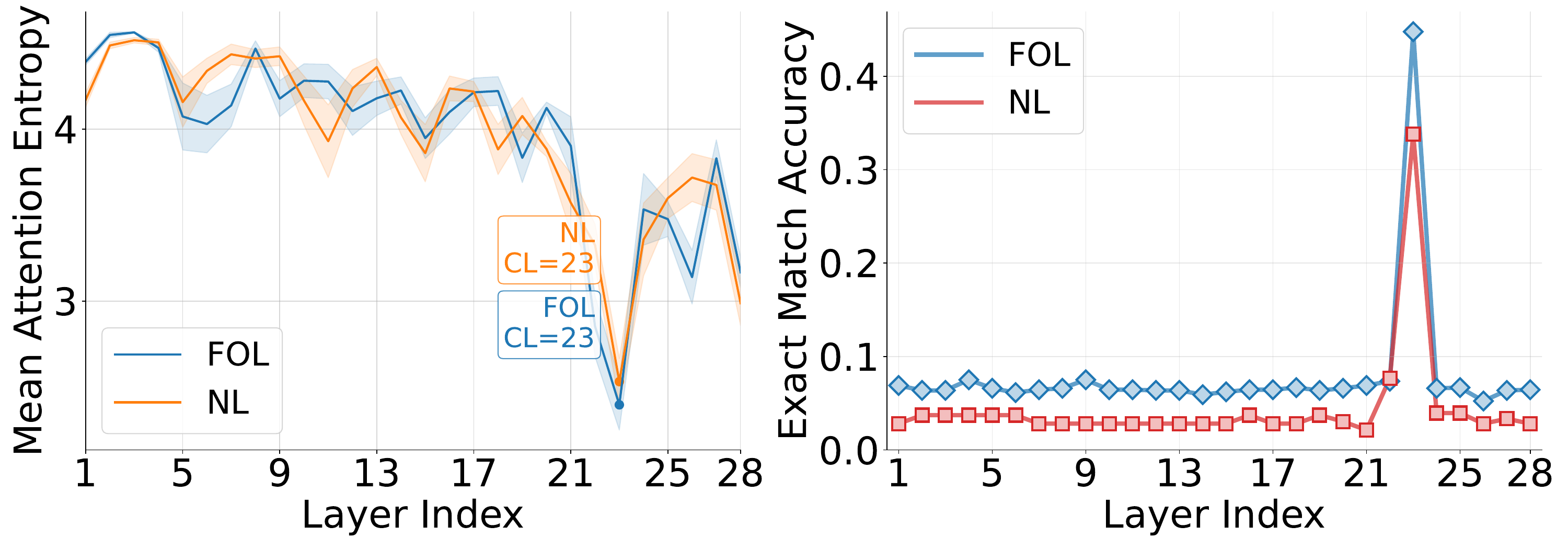}}%
    \caption{Confidence layer (CL) identification (FOL/NL), with shading showing attention entropy standard deviation (\textbf{left}). Per-layer retention analysis with correct rules injected into a single target layer (\textbf{right}).}
    \label{fig:qwen_7_combined_attention_entropy_and_layer_performance}
\end{figure}

\subsection{Confidence Layer Identification}
We identify Qwen2.5-7B-Instruct’s confidence layer by injecting 100 rules into the KV cache under both FOL and NL representations (Figure~\ref{fig:qwen_7_combined_attention_entropy_and_layer_performance}, left). In both cases, attention entropy is minimized at layer 23, showing that the same confidence layer emerges regardless of representation. Additional identification results across settings are reported in Appendix~\ref{appendix:extended_cl_identification}.
To further assess the functional importance of this layer, we run a per-layer retention experiment with both FOL and NL rules: the 100 injected rules are correct only at one target layer, while all other layers receive randomly sampled incorrect rules. As shown in Figure~\ref{fig:qwen_7_combined_attention_entropy_and_layer_performance} (right), accuracy peaks when correct rules are retained at the identified confidence layer for both representations. This result highlights the critical role of the confidence layer and validates it as the base layer for applying Stacked Step-Level Attention Training. 

\subsection{Main Results}

\subsubsection{Experiments on Rule Reasoning}

\begin{figure*}[t]
  \centering
  \begin{subfigure}{0.32\textwidth}
    \centering
    \includegraphics[width=\linewidth]{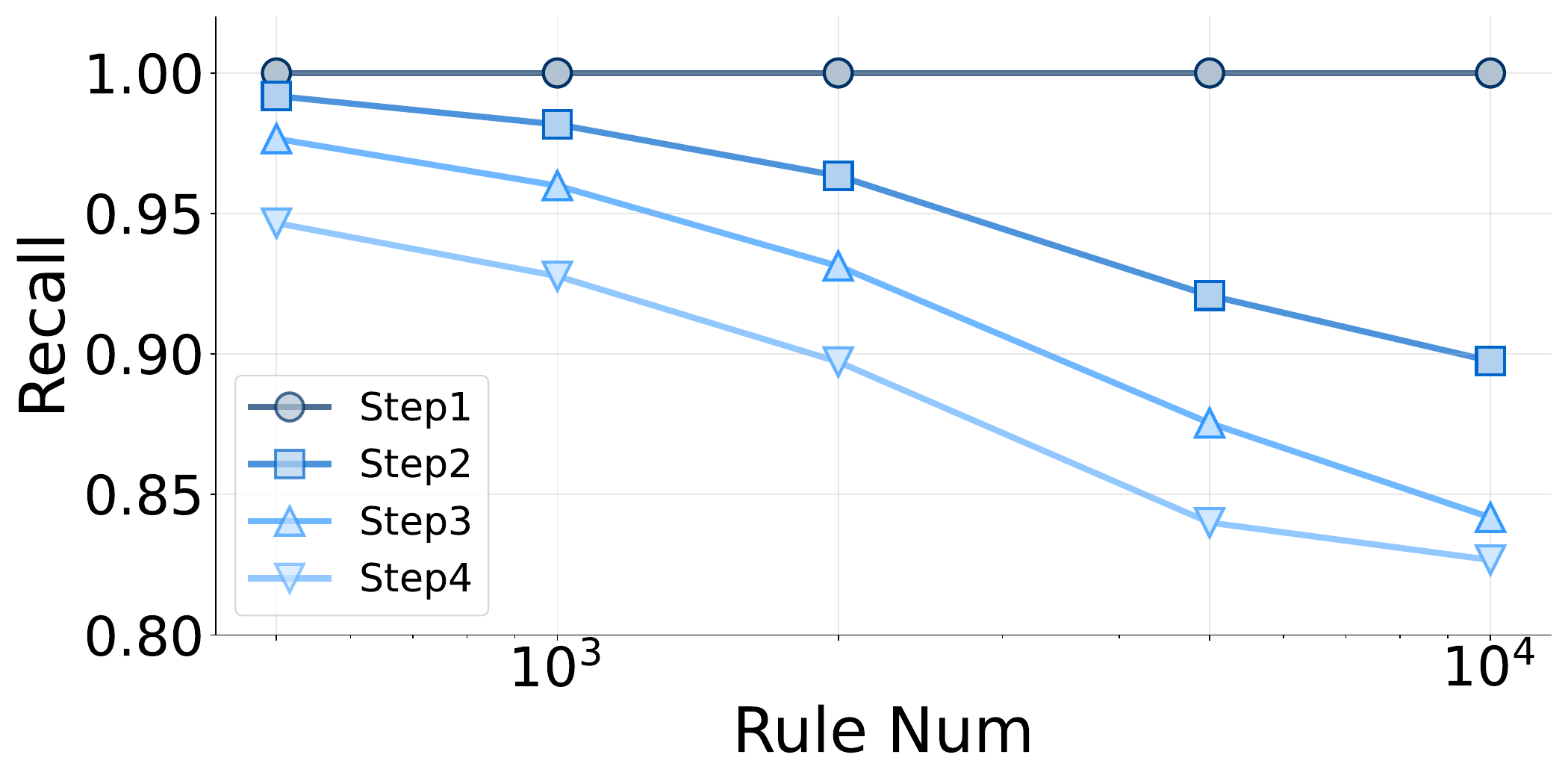}
    \caption{DynaRule Recall@100 across reasoning steps under FOL.}
    \label{fig:fol_recall_step_at_100}
  \end{subfigure}
  \hspace{0.005\textwidth}
  \begin{subfigure}{0.32\textwidth}
    \centering
    \includegraphics[width=\linewidth]{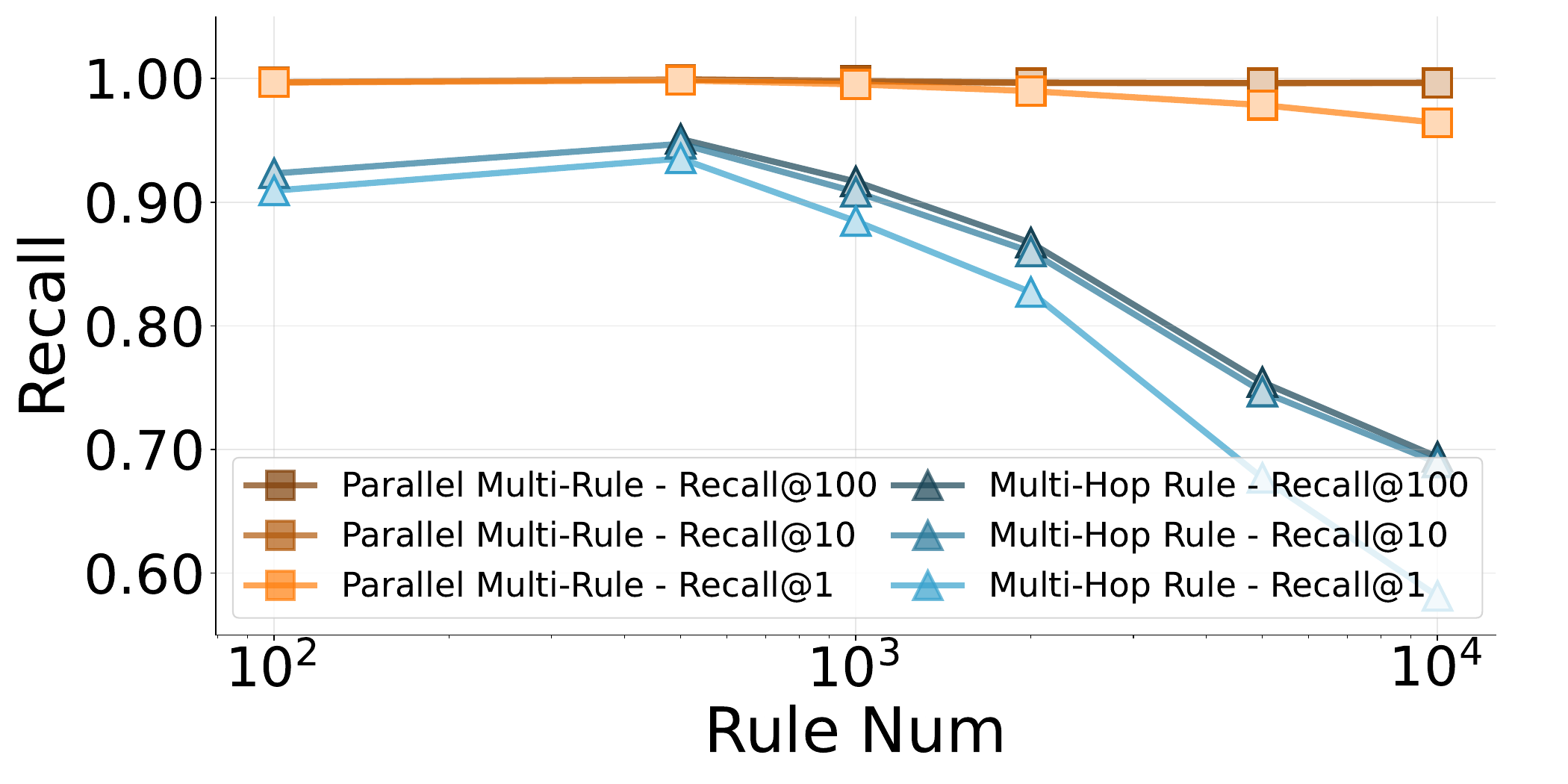}
    \caption{DynaRule recall under FOL in parallel multi-rule and multi-hop settings.}
    \label{fig:fol_recall1_comparison}
  \end{subfigure}
  \hspace{0.005\textwidth}
  \begin{subfigure}{0.32\textwidth}
    \centering
    \includegraphics[width=\linewidth]{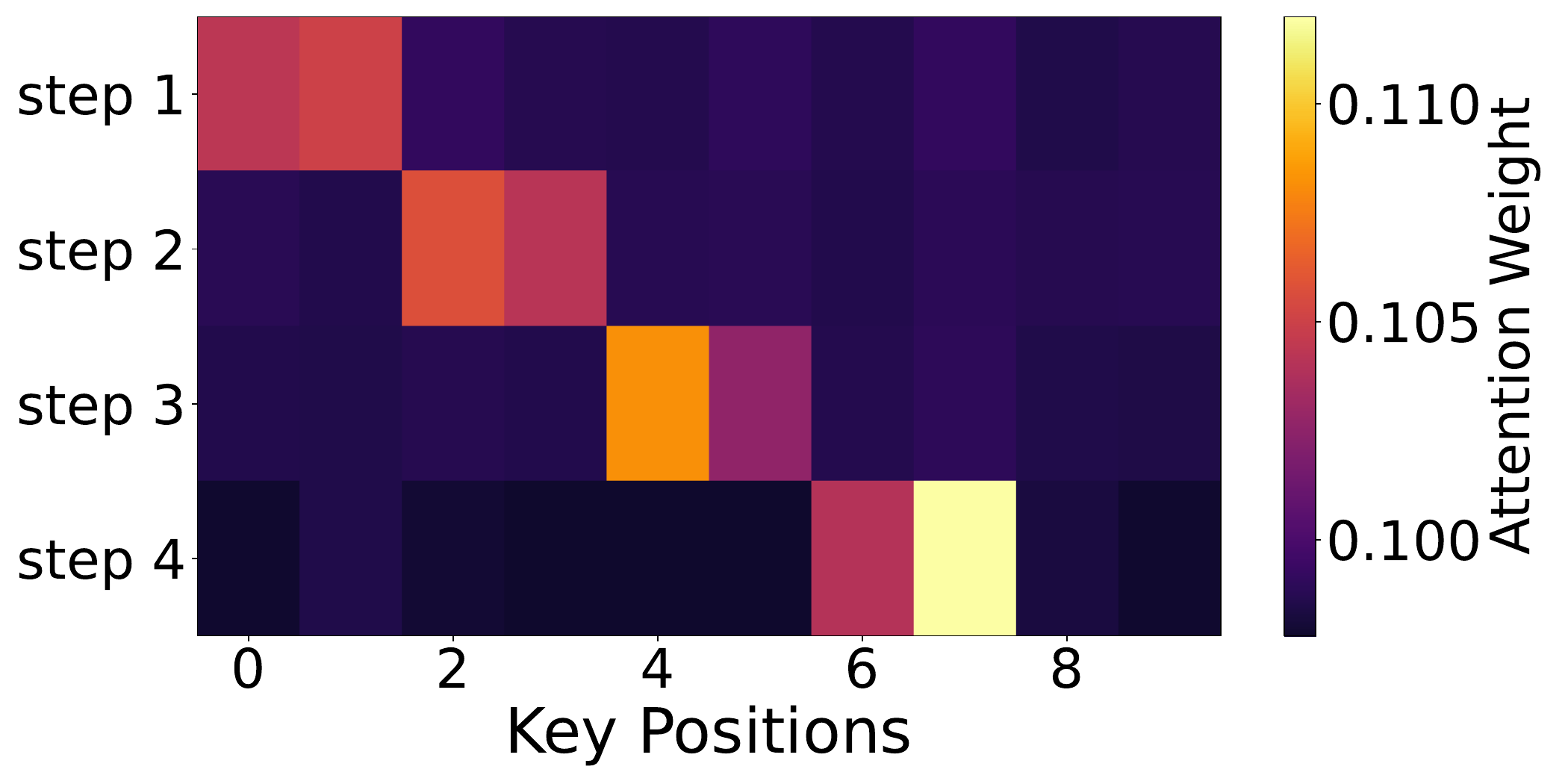}
    \caption{Rule attention weights (4-step). Required rules are ordered for clarity.}
    \label{fig:step_4_new}
  \end{subfigure}
  \caption{Retrieval performance (FOL) across QA types and reasoning steps, with step-wise attention visualization.}
  \label{fig:fol_vs_nl_comparison_and_qwen_node_recall_by_layer_save_}
\end{figure*}
We use exact match accuracy as the evaluation metric. Table~\ref{tab:fol_nl_main_qa_results} presents the main results on both FOL and NL rules, from which we draw two key observations. Detailed results under additional rule size settings are provided in Tables~\ref{tab:FOl_main_qa_results} and~\ref{tab:NL_main_qa_results}.
(1) Prompting only works at very small scales and fails on parallel and multi-hop settings, and RAG performs worse due to semantic mismatch. Existing end-to-end knowledge injection methods do not scale: KBLaM and SR-KI are competitive at small scales but degrade sharply as the number of rules increases, especially on parallel and multi-hop QA.
(2) Our method consistently performs best across rule scales, exceeding the strongest baseline by up to 19.88 points on average and staying strong under 10000 rules on parallel multi-rule QA (e.g., 0.72+ on multi-rule (5--8), >1.5× relative gain). Multi-hop QA poses a greater challenge than parallel settings, as it relies on rule chaining.
We further report prompting results of strong LLMs under a 1000-rule setting in Table~\ref{tab:icl_resutls_part}, while detailed results and case studies are shown in Appendices~\ref{appendix:in-context-results} and~\ref{appendix:case-study}. Even these strong models remain inferior to DynaRule, which achieves better overall performance.

\begin{table}
  \centering
  \setlength{\tabcolsep}{6pt}
  \scalebox{0.48}{%
    \begin{tabular}{c *{3}{c} *{3}{c} c}
    \toprule
    \multirow{2}{*}{Method}
      & \multicolumn{3}{c}{\textbf{First-order Logic}}
      & \multicolumn{3}{c}{\textbf{Natural Language}}
      & \multirow{2}{*}{\textbf{Avg.}} \\
    \cmidrule(lr){2-4}
    \cmidrule(lr){5-7}
      & \multicolumn{1}{c}{S-Rule}
      & \multicolumn{1}{c}{PM-Rule}
      & \multicolumn{1}{c}{MH-Rule}
      & \multicolumn{1}{c}{S-Rule}
      & \multicolumn{1}{c}{PM-Rule}
      & \multicolumn{1}{c}{MH-Rule} \\  
    \midrule
      Qwen2.5-14B-Instruct & 0.7200  & 0.4486  & 0.1933  & 0.6200  & 0.4031  & 0.1933  & 0.4297  \\ 
        Qwen2.5-32B-Instruct & 0.8400  & 0.5038  & 0.2067  & 0.7400  & 0.5357  & 0.2200  & 0.5077  \\ 
        Qwen2.5-72B-Instruct & 0.7200  & 0.5945  & 0.2533  & 0.7600  & 0.6698  & 0.1867  & 0.5307  \\ 
        Llama-3.1-70B-Instruct & 0.8600  & 0.5571  & 0.2067  & 0.9600  & 0.5633  & 0.1467  & 0.5490  \\ 
        Llama-3.3-70B-Instruct & 0.8600  & 0.4555  & 0.1667  & 0.9600  & 0.4686  & 0.1867  & 0.5162  \\ 
        Qwen3-32B & 0.5800  & 0.5326  & 0.2600  & 0.8800  & 0.4933  & 0.3333  & 0.5132  \\ 
        deepseek-chat & 0.8800  & 0.7821  & 0.3467  & 0.9000  & 0.7236  & 0.3800  & 0.6687  \\ 
        gpt-5.1-2025-11-13 & \underline{0.9600}  & 0.7576  & 0.3200  & \textbf{1.0000}  & 0.7602  & 0.2800  & 0.6796  \\ 
        gpt-5.5-2026-04-23 & 0.9800 & 0.8593 & 0.3533 & 0.8000 & \underline{0.9107} & 0.4667 & 0.7283 \\
        claude-sonnet-4-6 & 0.9000 & \underline{0.9117} & \underline{0.5133} & 0.8300 & 0.8595 & \underline{0.6433} & \underline{0.7763} \\
        \midrule
        \rowcolor{blue!7} DynaRule & \textbf{1.0000}  & \textbf{0.9816}  & \textbf{0.7083}  & 
        \underline{0.9800}  & \textbf{0.9724}  & \textbf{0.7200}  & \textbf{0.8937} \\ 
    \bottomrule
    \end{tabular}%
  }
  \caption{Partial prompting comparison results on RuleWorld under 1000 rules. \textbf{S-Rule}: Single-Rule QA; \textbf{PM-Rule}: Parallel Multi-Rule QA; \textbf{MH-Rule}: Multi-Hop Rule QA. Detailed results are provided in Appendix~\ref{appendix:in-context-results}.}
  \label{tab:icl_resutls_part}
\end{table}

\begin{figure*}[h]
  \centering
  \begin{subfigure}[t]{0.32\textwidth}
    \centering
    \includegraphics[width=\linewidth]{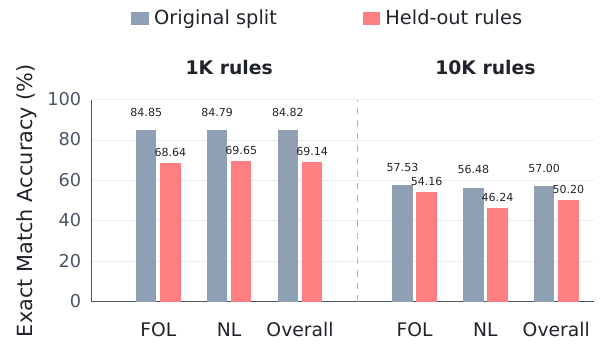}
    \caption{Exact match accuracy on original and held-out rules.}
    \label{fig:heldout_rule_generalization}
  \end{subfigure}\hfill
  \begin{subfigure}[t]{0.32\textwidth}
    \centering
    \includegraphics[width=\linewidth]{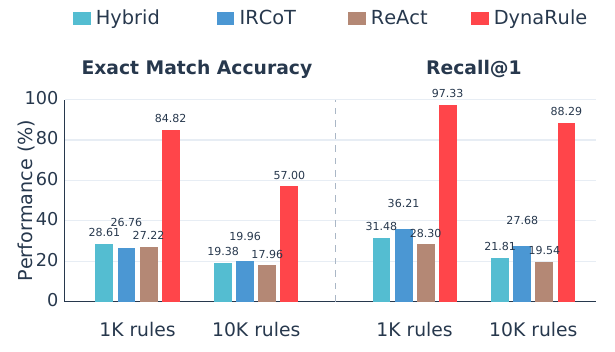}
    \caption{Exact match accuracy and Recall@1 results compared with Iterative RAG.}
    \label{fig:iterative_rag_analysis}
  \end{subfigure}\hfill
  \begin{subfigure}[t]{0.32\textwidth}
    \centering
    \includegraphics[width=\linewidth]{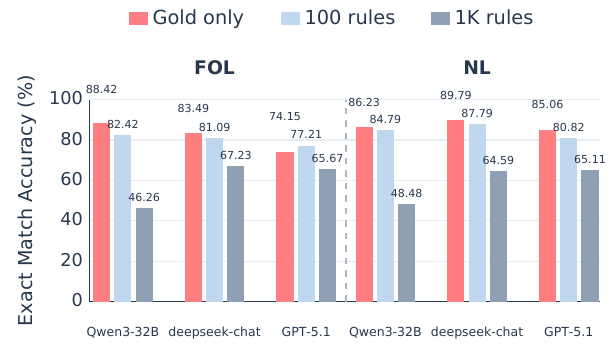}
    \caption{Exact match accuracy under only the gold rules, 100 rules, and 1K rules.}
    \label{fig:gold_rule_only_analysis}
  \end{subfigure}
  \caption{Rule generalization, iterative retrieval, and gold-rule application analyses.}
  \label{fig:additional_analysis}
  \vspace{-0.4em}
\end{figure*}

\subsubsection{Experiments on Rule Retrieval}
Table~\ref{tab:small_retrieval_results} reports main retrieval performance on FOL and NL rules, where Recall@1 sets $K$ to the number of gold rules per instance or step. Table~\ref{tab:retrieval_results} further provides results for additional rule size settings. For DynaRule, we report step-level correctness averaged across steps. We observe two findings.
(1) Baselines such as dense retrieval, BM25 and hybrid search are limited by semantic mismatch; KBLaM fails due to the lack of retrieval supervision, and SR-KI’s single-step retrieval paradigm cannot capture interacting rules. In contrast, DynaRule performs best across all settings, beating the strongest baseline by up to 61.98 (FOL) and 53.42 (NL) points.
(2) Multi-step retrieval compounds errors. As shown in Figures~\ref{fig:fol_recall_step_at_100} and~\ref{fig:fol_recall1_comparison}, DynaRule's recall drops at deeper steps, and multi-hop underperforms parallel multi-rule at the same scale, consistent with early-step mistakes propagating and leading to weaker performance in multi-hop settings.

The rule reasoning and retrieval results are stable. Detailed seed-level standard deviations and $t$-based 95\% confidence intervals are reported in Appendix~\ref{appendix:statistical_robustness}.
Detailed step-wise and QA-type analysis are provided in Appendix~\ref{appendix:Retrieval-Performance-across-different-Reasoning-Steps}.

\begin{table}
    \centering 
    \setlength{\tabcolsep}{8pt} 
    \scalebox{0.48}{%
      \begin{tabular}{%
        c
        *{3}{c}
        *{3}{c}
      }
      \toprule
      \multirow{2}{*}{Method}
        & \multicolumn{3}{c}{\textbf{First-order Logic}}
        & \multicolumn{3}{c}{\textbf{Natural Language}} \\
      \cmidrule(lr){2-4}
      \cmidrule(lr){5-7}
        & \multicolumn{1}{c}{Recall@100}
        & \multicolumn{1}{c}{Recall@10}
        & \multicolumn{1}{c}{Recall@1}
        & \multicolumn{1}{c}{Recall@100}
        & \multicolumn{1}{c}{Recall@10}
        & \multicolumn{1}{c}{Recall@1} \\  
      \midrule
      \multicolumn{7}{c}{\textit{Rule Num = 100}} \\ \midrule
          Dense & - & 0.5325 & 0.4448 & - & 0.6646 & 0.5414 \\
          BM25 & - & 0.6901 & 0.4375 & - & 0.5869 & 0.5292 \\
          Hybrid & - & 0.6364  & 0.4793 & - & 0.5972  & 0.5128  \\
          KBLaM & - & 0.5151 & 0.4122 & - & 0.5278 & 0.4025 \\
          SR-KI & - & \underline{0.9100} & \underline{0.8000} & - & \underline{0.8963} & \underline{0.7768} \\
          \rowcolor{blue!7} DynaRule & - & \textbf{0.9773} & \textbf{0.9733} & - & \textbf{0.9909} & \textbf{0.9857} \\
      \midrule
      \multicolumn{7}{c}{\textit{Rule Num = 1000}} \\ \midrule
          Dense & 0.5529 & 0.3635 & 0.3028 & 0.6961 & 0.4490 & 0.3587 \\
          BM25 & 0.7559 & 0.2962 & 0.1887 & 0.5946 & 0.4892 & 0.4424 \\
          Hybrid & 0.7935  & 0.3561  & 0.2456 & 0.7618  & 0.4464  & 0.3839  \\
          KBLaM & 0.5237 & 0.3532 & 0.1739 & 0.5389 & 0.3351 & 0.1686 \\
          SR-KI & \underline{0.9331} & \underline{0.7369} & \underline{0.5318} & \underline{0.9288} & \underline{0.7213} & \underline{0.5429} \\
          \rowcolor{blue!7} DynaRule & \textbf{0.9762} & \textbf{0.9737} & \textbf{0.9656} & \textbf{0.9932} & \textbf{0.9892} & \textbf{0.9810} \\
      \midrule
      \multicolumn{7}{c}{\textit{Rule Num = 10000}} \\ \midrule
          Dense & 0.3710 & 0.2573 & 0.2227 & 0.4619 & 0.2940 & 0.2241 \\
          BM25 & 0.3017 & 0.1102 & 0.0418 & 0.4970 & 0.4058 & \underline{0.3703} \\
          Hybrid & 0.5023  & 0.2344  & 0.1338 & 0.5876  & 0.3559  & 0.3023 \\
          KBLaM & 0.3884 & 0.0642 & 0.0239 & 0.3554 & 0.0776 & 0.0282 \\
          SR-KI & \underline{0.7758} & \underline{0.4180} & \underline{0.2415} & \underline{0.7539} & \underline{0.4465} & 0.2898 \\
          \rowcolor{blue!7} DynaRule & \textbf{0.9145} & \textbf{0.9130} & \textbf{0.8613} & \textbf{0.9614} & \textbf{0.9526} & \textbf{0.9045} \\
      \bottomrule
      \end{tabular}%
    }
    \caption{Retrieval performance on FOL and NL rules, with extended results in Table~\ref{tab:retrieval_results}.}
    \label{tab:small_retrieval_results}
  \end{table}

\subsection{Analysis}
\paragraph{Step-Aware Retrieval Analysis}
Figure~\ref{fig:step_4_new} shows 4-step rule retrieval, where each step contains two gold rules. In Step 1, attention focuses on Rules 0--1 (the gold rules for Step 1). In Step 2, it shifts to Rules 2--3 (the gold rules for Step 2) while attention to Rules 0--1 drops. Later steps follow the same pattern, indicating targeted, step-aware retrieval triggered by \texttt{<search>}.
Detailed attention visualizations across steps are provided in Appendix~\ref{appendix:attention_weight_across_steps}.

\paragraph{Generalization to Unseen Rules}
We additionally evaluate the same trained DynaRule models on QA instances constructed from a rule repository disjoint from the training rule subset, without any further adaptation. The held-out setting follows the same setting as the main experiments. In Figure~\ref{fig:heldout_rule_generalization}, Overall denotes the aggregate across FOL and NL results. Performance decreases on unseen rules, but remains substantial: at 10K rules, the overall exact match accuracy is 50.20\%, only 6.80 percentage points below the original split. This result indicates that the adapters learn a transferable alignment from rule embeddings to the model's internal knowledge state rather than simply memorizing the training repository.

\paragraph{Comparison with Iterative RAG}
To test whether stronger inference-time retrieval closes the gap, we compare DynaRule with IRCoT~\cite{trivedi-etal-2023-interleaving} and a ReAct-style~\cite{yao-etal-2023-react} retrieval agent implemented with LangGraph~\cite{langgraph}, using Qwen2.5-7B-Instruct and the same Hybrid RAG retriever as in the main experiments (BM25 and Qwen3-Embedding-8B combined with RRF, $k=60$). Each retrieval step returns 100 rules. The final answer is generated from the final top-100 aggregated rules, which are also used to compute Recall@1. IRCoT and ReAct use a four-round/loop budget. All comparisons follow the same evaluation setting as the main experiments. Figure~\ref{fig:iterative_rag_analysis} reports exact match accuracy (left) and Recall@1 (right), each averaged across the FOL and NL task results. The results show that merely adding multiple inference-time retrieval steps does not necessarily yield better rule localization and application. At 10K rules, IRCoT and ReAct achieve 19.96\%/17.96\% exact match accuracy and 27.68\%/19.54\% Recall@1, respectively, both far lower than DynaRule's 57.00\% exact match accuracy and 88.29\% Recall@1. DynaRule is more reliable because its internal step-level retrieval is jointly trained with reasoning, enabling rule localization and rule utilization to be aligned with the decoding process.

\paragraph{Gold-Rule-Only Analysis}
Figure~\ref{fig:gold_rule_only_analysis} compares average exact match accuracy over 11 QA subtasks under 100-rule, 1K-rule, and gold-rule-only prompting for Qwen3-32B, deepseek-chat, and gpt-5.1-2025-11-13 (GPT-5.1). Increasing the pool from 100 to 1K lowers accuracy across models and representations, highlighting irrelevant-rule noise. Providing only gold rules removes this noise and isolates rule application: the models attain 74.15--89.79\% exact match accuracy across FOL and NL, indicating that the main challenge is accurate rule localization and application.

\begin{figure}[t]
  \centering
  \begin{subfigure}{0.48\textwidth}
    \centering
    \includegraphics[width=\linewidth]{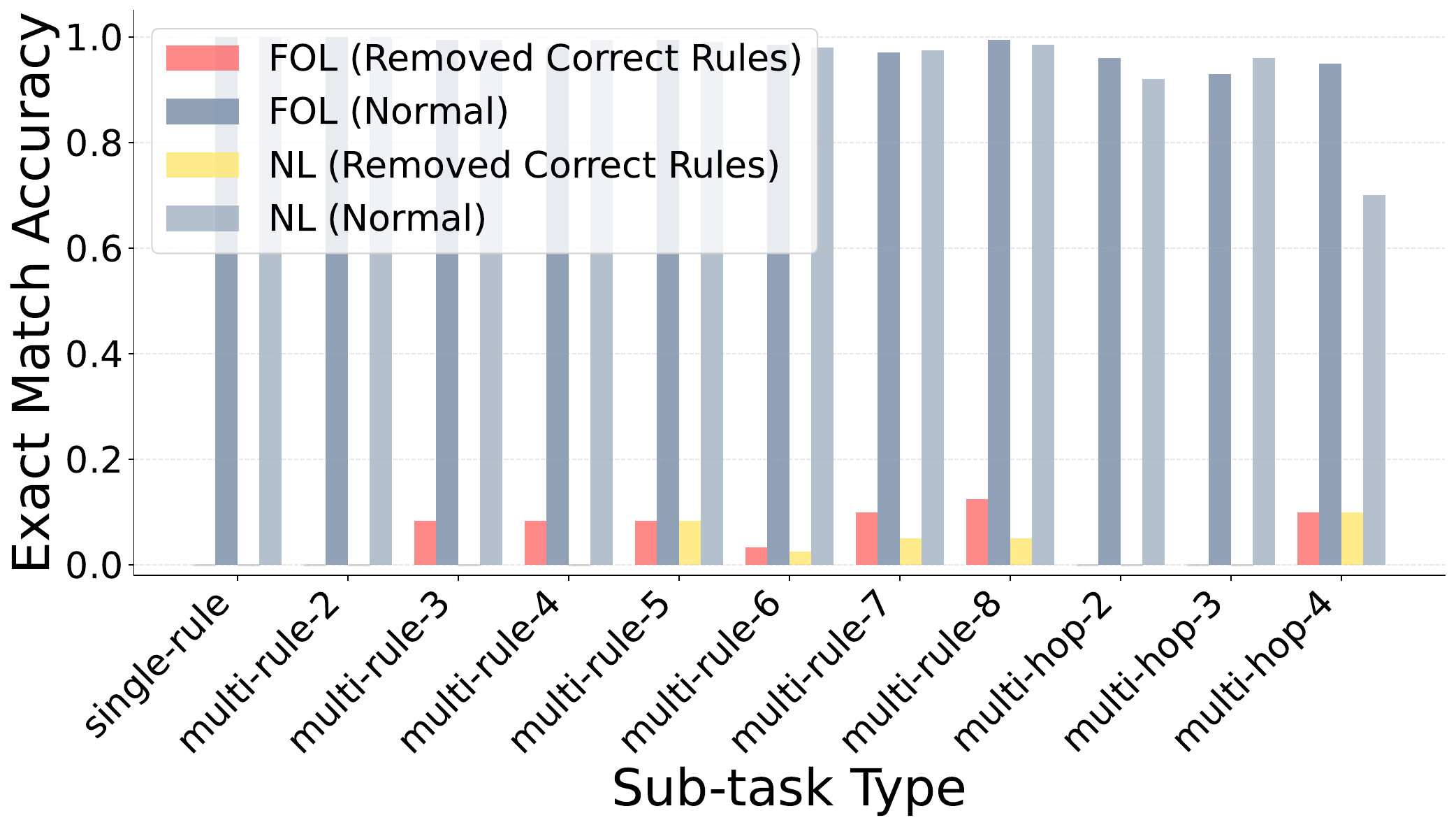}
    \caption{Comparison between removing correct rules and the normal setting.}
    \label{fig:fol_vs_nl_comparison}
  \end{subfigure}
  \hspace{0.005\textwidth}
  \begin{subfigure}{0.48\textwidth}
    \centering
    \includegraphics[width=\linewidth]{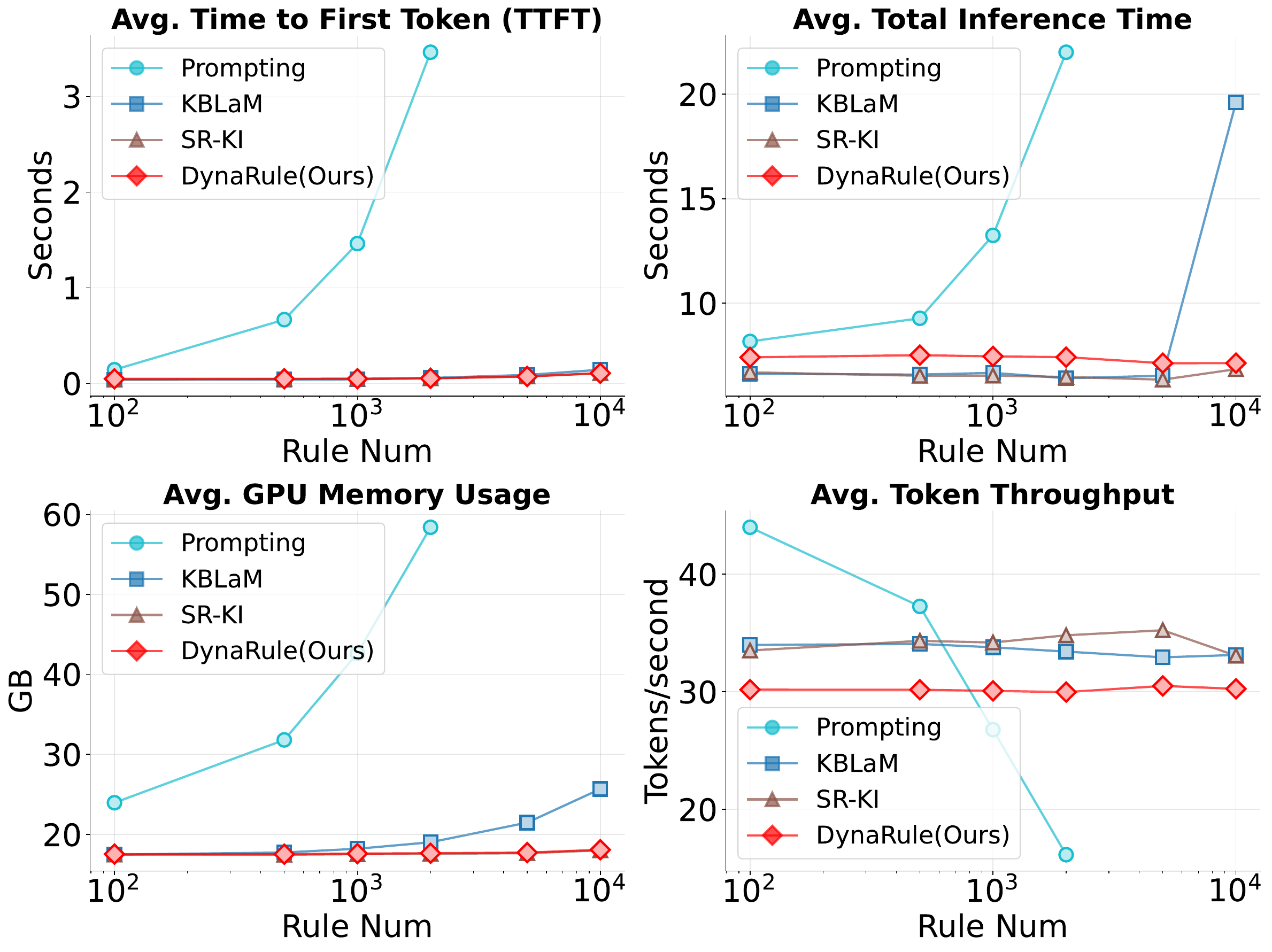}
    \caption{Efficiency analysis of TTFT (top left), inference time (top right), GPU memory usage (bottom left), and token throughput (bottom right) on a single A800 GPU, using Qwen-2.5-7B-Instruct as the base model.}
    \label{fig:efficiency_performance_comparison_2x2}
  \end{subfigure}
  \caption{Effectiveness and efficiency evaluation.}
  \label{fig:fol_vs_nl_comparison_and_efficiency_performance_comparison_2x2}
\end{figure}

\paragraph{Rule Dependence Analysis} 
To verify genuine reliance on injected rules, we conduct the ablation study in Figure~\ref{fig:fol_vs_nl_comparison}. With 100 injected rules, removing all correct rules from every layer collapses accuracy under both FOL and NL, often to zero. This shows the model relies on injected rules, and suggests that the linear adapters primarily learn rule alignment rather than memorizing specific rules.

\paragraph{Efficiency Analysis}
\label{section:Efficiency_Analysis}
Efficiency is measured by TTFT, total inference time, GPU memory, and token throughput, as shown in Figure~\ref{fig:efficiency_performance_comparison_2x2}.
KV cache injection scales as $O((M+N)N)$, far cheaper than quadratic prompt injection with long rule prompts. This reduces TTFT, inference time, and memory, and improves throughput under large rule sets. Moreover, \texttt{<search>} is a single token, so each step adds only $O(M)$ retrieval cost, incurring minimal latency and only a slight drop in throughput.

\section{Conclusion}
We introduce \textbf{RuleWorld}, a large-scale benchmark built from a globally consistent rule set for evaluating LLMs' ability to use external procedural rules in single-rule, parallel multi-rule and multi-hop settings. We further propose \textbf{DynaRule}, a step-level rule integration framework that unifies rule retrieval, updating, and reasoning entirely within the model’s internal space. Extensive experiments show that DynaRule substantially improves the reliability and accuracy of procedural rule application. 
This work exposes the gap between parametric knowledge and explicit rule grounding, advancing more dynamic reasoning.

\section*{Limitations}
Although RuleWorld provides globally consistent non-commonsense rules and supports diverse reasoning settings, its rules are currently limited to FOL and natural language forms. Future work may explore richer representations, such as executable code or more complex rule-triggering patterns.
DynaRule enables end-to-end rule retrieval, updating, and reasoning, but embedding-only rule representations inevitably lose information, which may cause retrieval or application errors as rule pools scale. While we evaluate up to 10,000 injected rules, the full rule set contains millions, and LLMs already degrade at the 1,000-rule scale. Future work may improve scalability through jointly trained rule encoders, hierarchical or clustered indexing, and more structured rule representations.

\bibliography{custom}

\appendix
\label{sec:appendix}

\section{Detailed Description of the RuleWorld Benchmark}
\subsection{Data Statistics}
\label{appendix:RuleWorld-Data-Statistics}

\begin{table}[htbp]
\centering
\scalebox{0.7}{
\setlength{\tabcolsep}{6pt}
\renewcommand{\arraystretch}{1.15}
\begin{tabular}{l l r r}
\toprule
\textbf{Major Type} & \textbf{Relation Type} & \textbf{All Rules} & \textbf{Train/Eval Rules} \\
\midrule
\multirow{2}{*}{Attribute}
    & Entity2Attr      & 1,877,380 & 74,888 \\
    & AttrChange2Attr  & 4,021     & 3,772 \\
\midrule
\multirow{3}{*}{Action}
    & Action2Env       & 100       & 100 \\
    & Action2Attr      & 200       & 200 \\
    & Action2State     & 200       & 200 \\
\midrule
Environment
    & Env2State        & 100       & 100 \\
\midrule
State
    & State2Attr       & 3,054,112 & 20,975 \\
\midrule
\multicolumn{2}{c}{\textbf{Total}} 
    & \textbf{4,936,113} & \textbf{100,235} \\
\bottomrule
\end{tabular}
}
\caption{Count distribution of full rule set and the subset used for training and evaluation.}
\label{tab:rules_full_and_training_and_evaluation}
\end{table}

RuleWorld contains 4.94 million abstract, conflict-free procedural rules spanning four major types and seven relation sub-types (Table~\ref{tab:rules_full_and_training_and_evaluation}). All rules are fully grounded and self-consistent, allowing direct injection into any QA instance without additional post-processing, which supports systematic evaluation under large-scale rule injection. Here, ``A2B'' denotes that A influences or determines B, such as \textit{Entity2Attr} and \textit{State2Attr}. Most rules belong to \textit{Entity2Attr} and \textit{State2Attr}, with 1.87M and 3.05M instances, respectively. To build large-scale coverage, we generate many non-commonsense \textit{Entity2Attr} rules as the atomic basis of RuleWorld, while keeping other relation types at moderate scale to maintain balanced reasoning patterns. Together, these relations interact compositionally and support the construction of a large and diverse QA space.

Based on this rule corpus, we instantiate 3.37 million rule-based QA pairs spanning three reasoning settings: Single-Rule QA, Parallel Multi-Rule QA, and Multi-Hop Rule QA (Table~\ref{tab:qa_type_whole_training}). Parallel Multi-Rule QA covers up to eight gold rules per instance, while Multi-Hop Rule QA spans up to four reasoning hops, yielding eleven sub-tasks of increasing complexity. Most QA instances come from Parallel Multi-Rule QA, which contributes 3.25M examples and reflects the combinatorial difficulty of jointly applying multiple procedural rules. In contrast, Multi-Hop QA contains 77,737 instances and focuses on sequential inference across reasoning steps. This distribution supports fine-grained analysis of model behavior under varying reasoning depth and rule composition complexity.

For training and evaluation, we further sample 100,235 rules from the full corpus while preserving its structural distribution (Table~\ref{tab:rules_full_and_training_and_evaluation}). Based on these sampled rules, we construct a training subset of 111,200 QA instances (Table~\ref{tab:qa_type_whole_training}), including 25,000 Single-Rule, 49,200 Parallel Multi-Rule, and 37,000 Multi-Hop examples. The test split uses the same sampled rules but entirely disjoint QA instances, and is uniformly sampled across difficulty levels with manual verification, ensuring balanced and reliable evaluation.

\begin{table}[t]
\centering
\scalebox{0.6}{
\setlength{\tabcolsep}{5.4pt}
\renewcommand{\arraystretch}{1.15}
\begin{tabular}{l c r r}
\toprule
\textbf{QA Type} & \textbf{Sub-task Type} & \textbf{All QA Count} & \textbf{Train QA Count} \\
\midrule
\multirow{1}{*}{Single-Rule QA} 
    & single-rule & 50,000 & 25,000 \\
\midrule
\multirow{7}{*}{Parallel Multi-Rule QA}
    & multi-rule-2 & 154,167 & 6,702 \\
    & multi-rule-3 & 381,116 & 8,696 \\
    & multi-rule-4 & 839,755 & 12,384 \\
    & multi-rule-5 & 886,455 & 10,924 \\
    & multi-rule-6 & 618,577 & 6,824 \\
    & multi-rule-7 & 289,761 & 2,863 \\
    & multi-rule-8 & 80,169  & 807 \\
\midrule
\multirow{3}{*}{Multi-Hop Rule QA}
    & multi-hop-2 & 37,737 & 19,500 \\
    & multi-hop-3 & 30,000 & 12,500 \\
    & multi-hop-4 & 10,000 & 5,000 \\
\midrule
\multicolumn{2}{c}{\textbf{Total}} 
    & \textbf{3,377,737} & \textbf{111,200} \\
\bottomrule
\end{tabular}
}
\caption{Count distributions for full and training QA instances across RuleWorld QA types.}
\label{tab:qa_type_whole_training}
\end{table}

\begin{figure*}[t]
    \centerline{\includegraphics[scale=0.555]{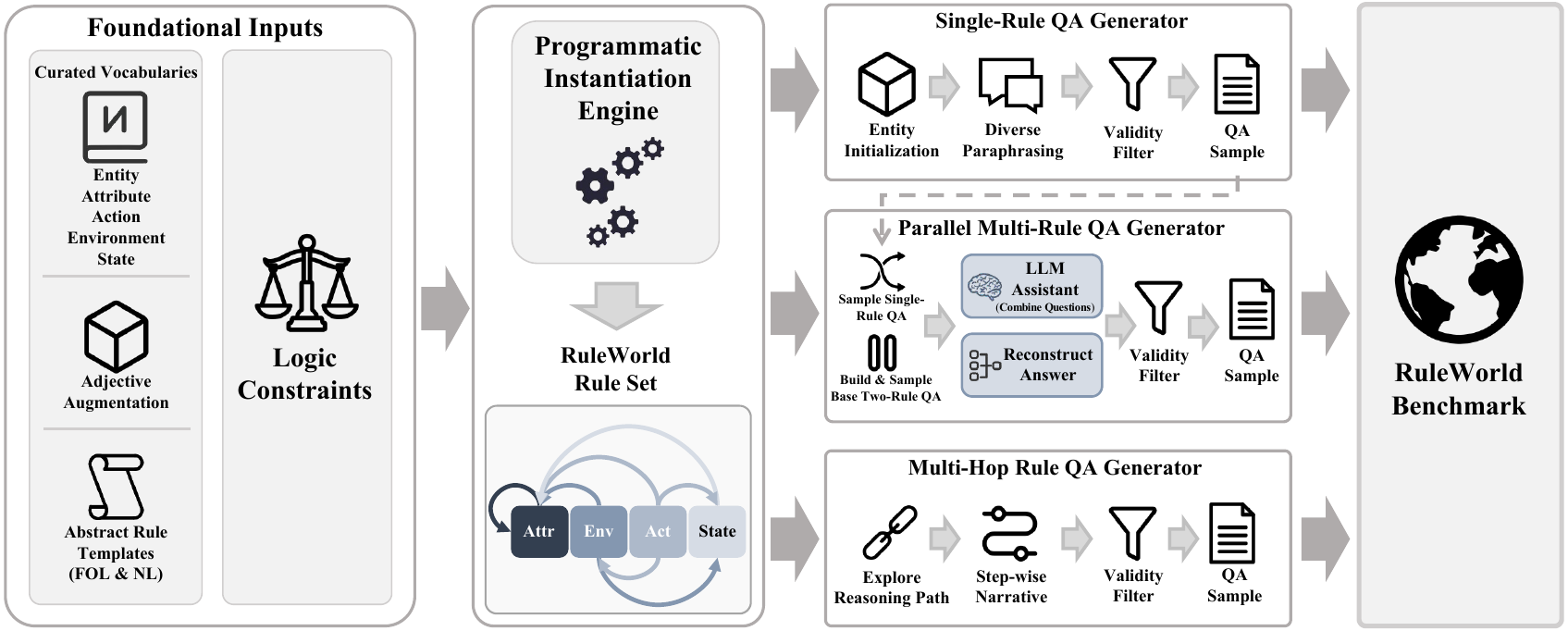}}
    \caption{Illustration of the construction pipelines for the RuleWorld rule set (\textbf{left}) and the RuleWorld QA benchmark (\textbf{right}), including single-rule, parallel multi-rule, and multi-hop QA generation. The modular process integrates curated vocabularies, logic constraints, programmatic instantiation, and task-specific generation workflows. In the rule set, Attr, Env, Act, and State denote the four major types \textit{Attribute}, \textit{Environment}, \textit{Action}, and \textit{State}, and the directed arrows between them indicate their interaction relations.}
    \label{fig:construction}
\end{figure*}

\subsection{Construction Process}
\label{appendix:RuleWorld-Construction-Process}

As shown in Figure~\ref{fig:construction}, RuleWorld is primarily built through a programmatic generation framework that synthesizes both the rule set and the QA instances at scale. For \textit{Parallel Multi-Rule QA}, LLMs are further used to merge multiple sub-questions into coherent composite questions. To ensure reliable evaluation, we manually annotate the \textit{Parallel Multi-Rule QA} portion of the test sets.

\paragraph{Rule Set Construction}
As illustrated in Figure~\ref{fig:construction}, RuleWorld begins with predefined vocabularies of \textit{entities}, \textit{attributes}, \textit{actions}, \textit{environments}, and \textit{states}, which define the synthetic world. We further expand lexical diversity by applying adjective-based augmentation, especially to \textit{entities}, \textit{attributes}, and \textit{states}. Based on these vocabularies, we specify abstract interaction patterns, such as action to attribute, state to attribute, or environment to state, and instantiate them into rule templates in both NL and FOL formats. A programmatic engine then enumerates and samples valid condition effect combinations to generate millions of rules.
To ensure logical soundness, all generated rules are filtered through consistency checks. In particular, we remove cyclic or alternative causal paths for single-step ``A2B'' transformations, so that each atomic rule corresponds to a unique and unambiguous mapping. Although individual rules are strictly pruned, the resulting rule set still supports rich multi-hop compositions. This pipeline yields a grounded, conflict-free, and semantically diverse rule set for controlled reasoning evaluation.

\begin{figure}[htbp]
    \centerline{\includegraphics[scale=0.25]{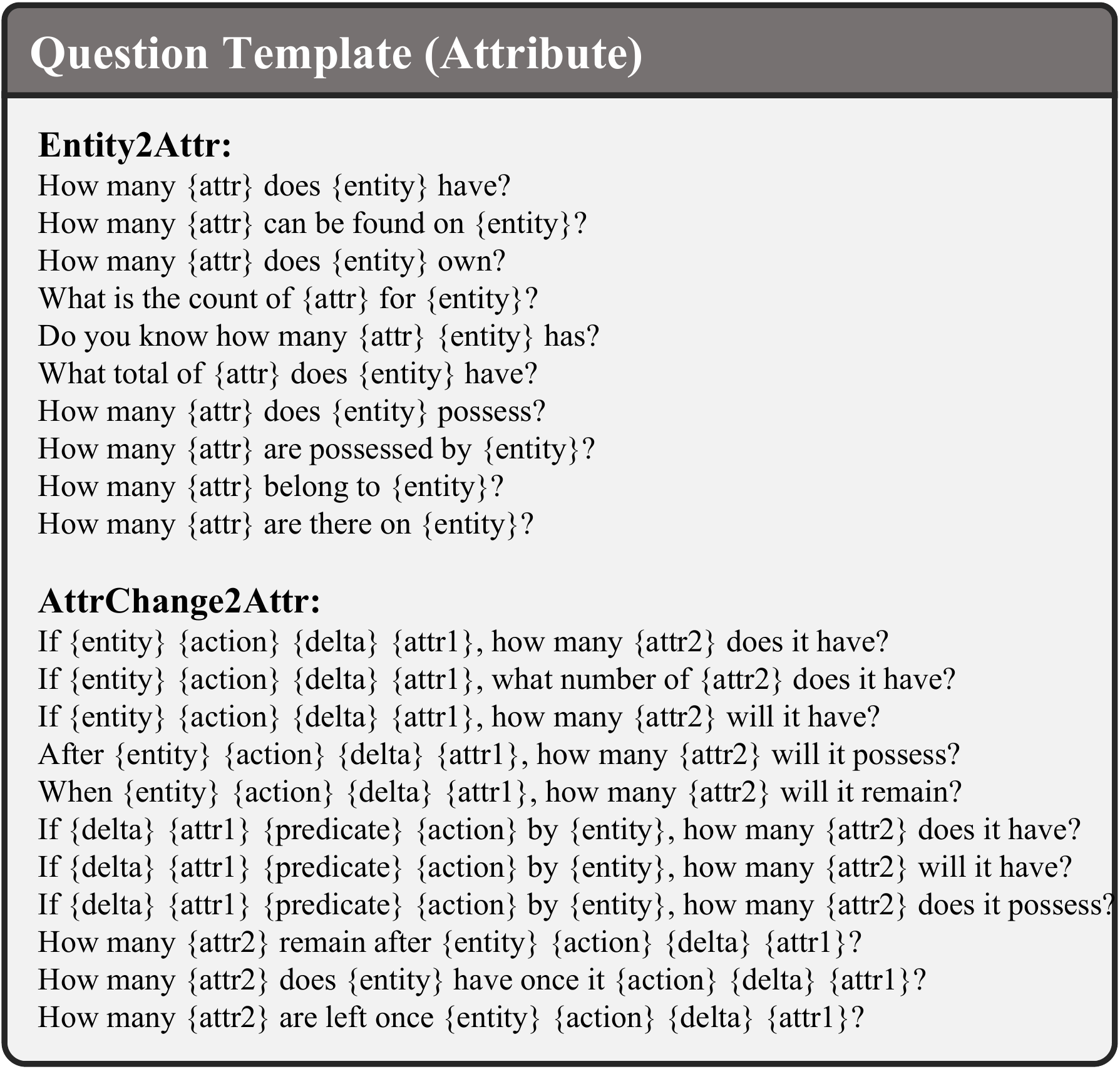}}%
    \caption{Basic question templates for Attribute-type rules.}
    \label{fig:QA_template_attr}
\end{figure}

\begin{figure}[htbp]
    \centerline{\includegraphics[scale=0.25]{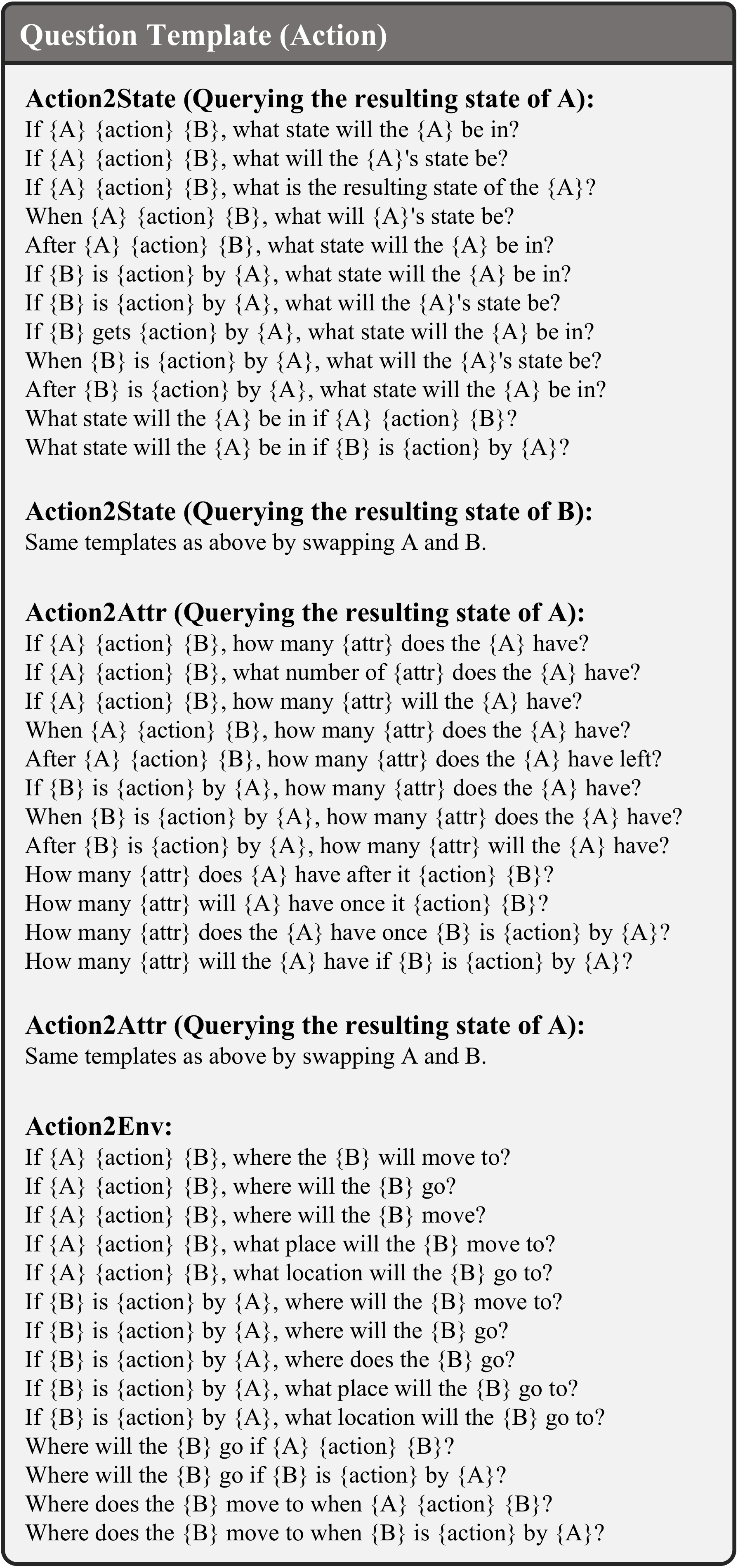}}%
    \caption{Basic question templates for Action-type rules.}
    \label{fig:QA_template_act}
\end{figure}

\begin{figure}[htbp]
    \centerline{\includegraphics[scale=0.25]{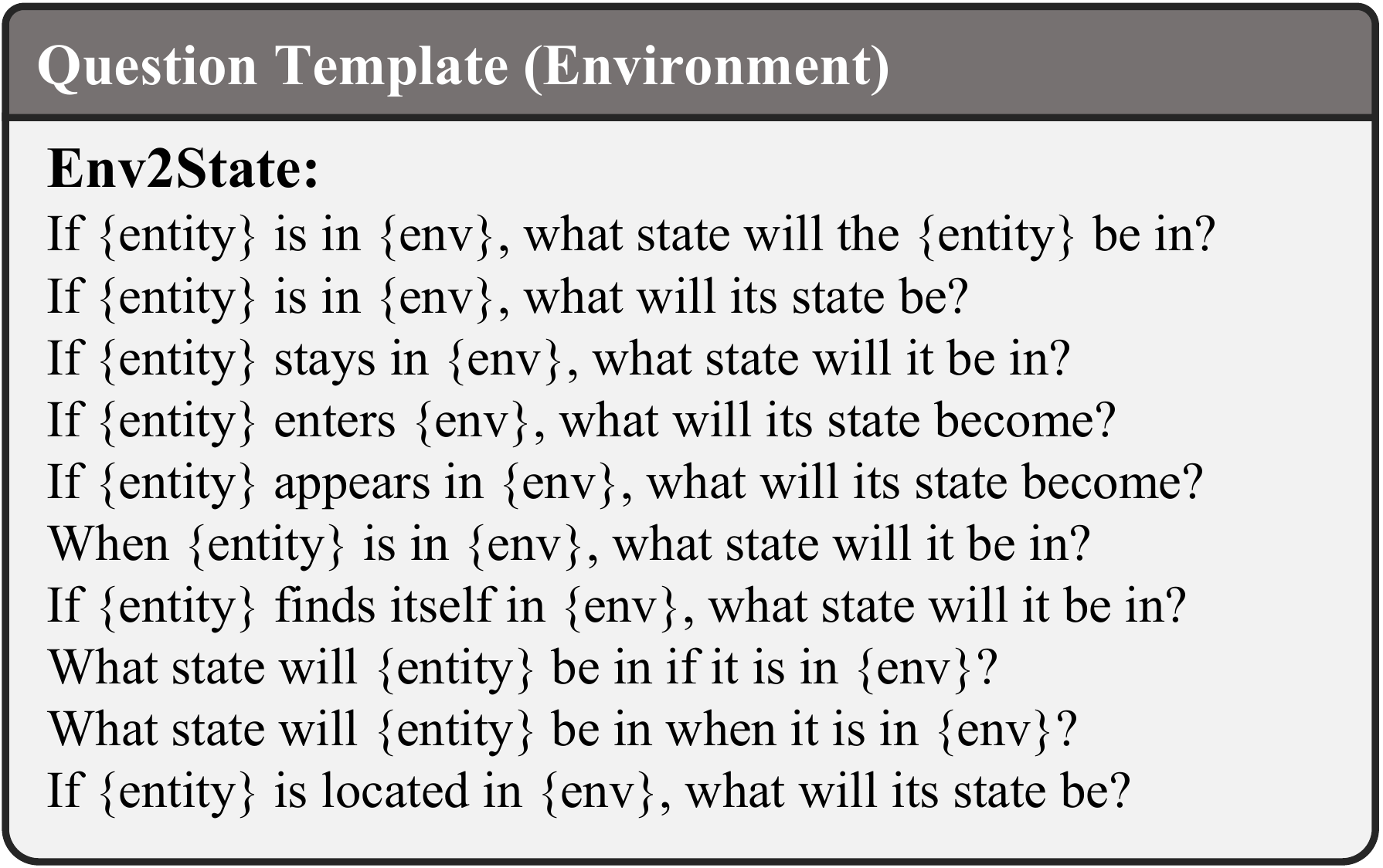}}%
    \caption{Basic question templates for Environment-type rules.}
    \label{fig:QA_template_env}
\end{figure}

\paragraph{Single-Rule QA Construction}
Given the rule set, we construct Single-Rule QA instances to evaluate whether a model can apply one atomic rule under a controlled initial state. For each rule type, we instantiate diverse question templates to reduce stylistic overfitting. Figures~\ref{fig:QA_template_attr}--\ref{fig:QA_template_env} show a superset of templates used across the dataset, covering multiple linguistic forms such as conditional, temporal, interrogative, and voice-variant expressions.

For each instance, the system assigns an initial entity configuration, applies one selected rule, and generates a question whose answer requires exactly one rule application. Each sample is produced automatically together with structured rule provenance, proof steps, and natural language explanations for supervision and analysis, while evaluation only uses the final answer. Invalid or semantically impossible samples, such as those producing negative attribute values, are removed automatically.

\begin{figure}[htbp]
    \centerline{\includegraphics[scale=0.25]{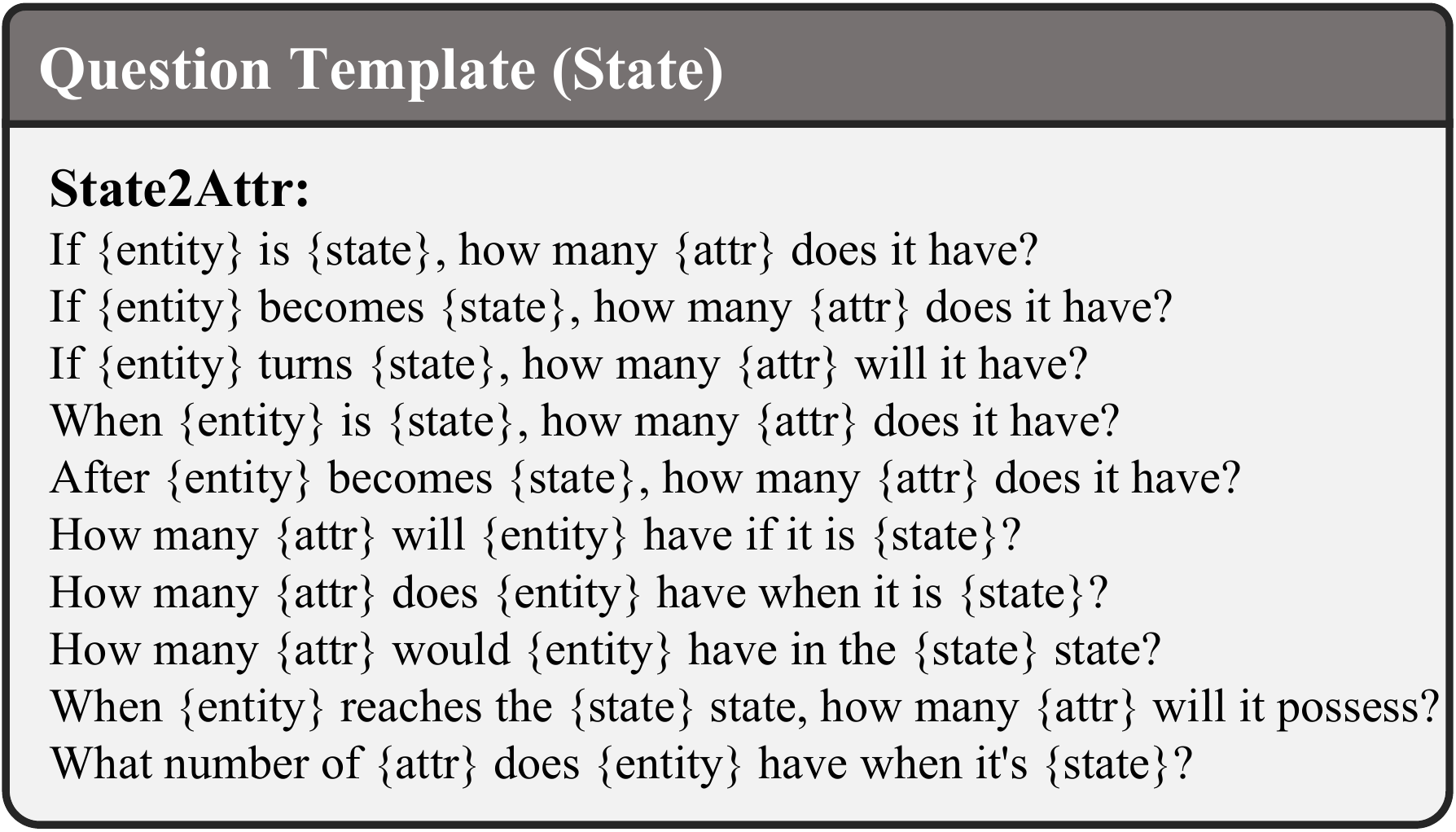}}%
    \caption{Basic question templates for State-type rules.}
    \label{fig:QA_template_state}
\end{figure}

\paragraph{Parallel Multi-Rule QA Construction}
Building on single-rule QA, we construct Parallel Multi-Rule QA instances by combining multiple independent sub-questions. Here, \emph{parallel} means that the reasoning steps are independent and order-invariant: no step depends on the intermediate result of another, even when multiple steps involve the same entity. As preparation, we first build two-rule QA instances for types such as \textit{AttrChange2Attr}, \textit{Action2Attr}, and \textit{State2Attr}, where an auxiliary \textit{Entity2Attr} rule provides the required initial attribute value. Examples of these templates are shown in Figures~\ref{fig:QA_template_attr}, \ref{fig:QA_template_act}, and \ref{fig:QA_template_state}.

\begin{figure}[htbp]
    \centerline{\includegraphics[scale=0.25]{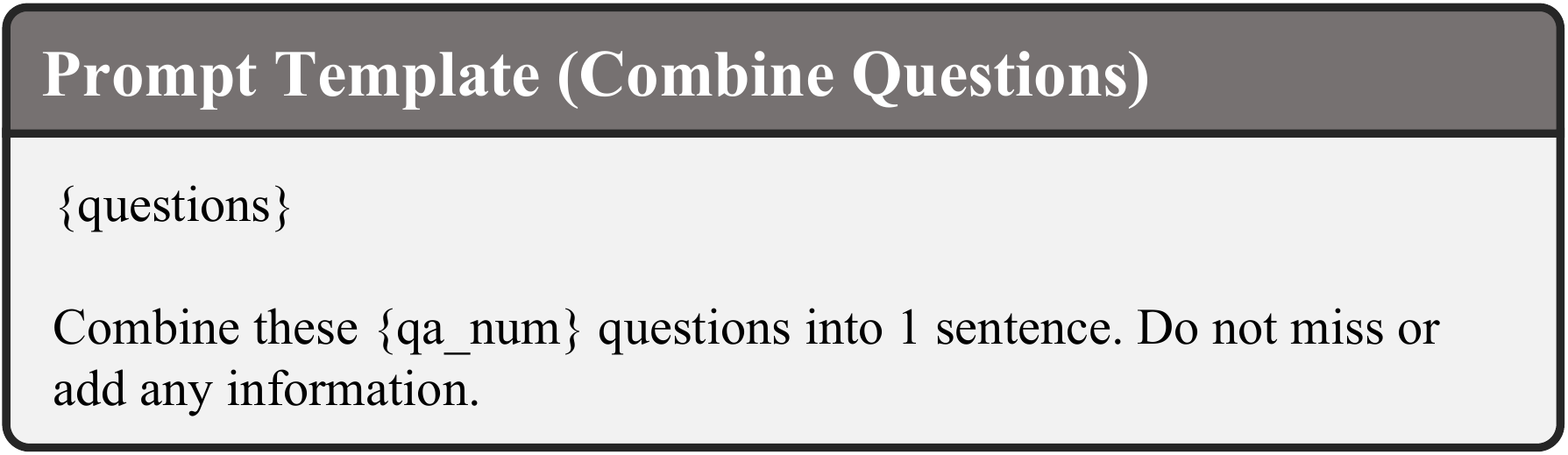}}%
    \caption{Template for combining questions to form Parallel Multi-Rule QA.}
    \label{fig:QA_template_combine_q}
\end{figure}

We then sample from the single-rule and two-rule pools and merge up to four questions into one combined query, covering at most eight atomic rules. The final answer is obtained by executing all rule effects in parallel and aggregating the original sub-question answers. To improve naturalness, we use Qwen2.5-72B-Instruct to merge separate questions into a coherent composite prompt without changing their semantics. The prompt template is shown in Figure~\ref{fig:QA_template_combine_q}. Each merged instance inherits the rule metadata of its constituent samples, enabling precise tracking of rule width and structure. Invalid merged samples are filtered using the same criteria as in single-rule QA construction.

\paragraph{Multi-Hop Rule QA Construction}
For sequential reasoning, we construct Multi-Hop Rule QA instances based on predefined reasoning skeletons, such as \textit{Action2Env} $\rightarrow$ \textit{Env2state} $\rightarrow$ \textit{State2Attr} $\rightarrow$ \textit{AttrChange2Attr}. The generator explores these skeletons by sampling valid rule combinations, simulating stepwise state transitions, and checking that every intermediate step is logically consistent. It then forms a question whose answer depends on the final result of the chain.

The generated metadata stores the reasoning depth, intermediate proofs, and full derivation trace. During construction, each rule application is also converted into natural language descriptions, and the same validity checks used in single-rule QA construction are applied to remove invalid samples. This process produces multi-hop QA instances with explicit causal dependencies and rich supervision signals.

Together, these procedures define RuleWorld as a scalable rule-centric benchmark that transforms procedural rules into controlled QA tasks. Through deterministic generation and LLM-assisted composition, RuleWorld supports single-rule, parallel, and multi-hop reasoning, enabling fine-grained evaluation of rule retrieval and rule execution in LLMs.

\begin{table*}[tbp]
\scalebox{0.70}{
\centering
\setlength{\tabcolsep}{4pt}
\renewcommand{\arraystretch}{1.25}

\begin{tabular}{c c l l}
\toprule
\textbf{Major Type} & \textbf{Relation Type} & \textbf{FOL} & \textbf{NL} \\
\midrule

\multirow{2}{*}{Attribute}
  & Entity2Attr
  & Tiny\_cat(A) $\Rightarrow$ Has(Strong\_horn, 4)
  & If A is a tiny cat, it has 4 strong horns. \\
  & AttrChange2Attr
  & Lose\_Strong\_fin(A, 1) $\Rightarrow$ Drop\_Mineral\_fur(A, 2)
  & If A loses 1 strong fin, A will drop 2 mineral furs. \\
\midrule

\multirow{5}{*}{Action}
  & Action2Env
  & Chase(A, B) $\Rightarrow$ Enter(B, Bridge)
  & If A chases B, B will enter bridge. \\
  & Action2Attr
  & Chase(A, B) $\Rightarrow$ Drop\_Floral\_paw(B, 1)
  & If A chases B, B will drop 1 floral paw. \\
  & Action2Attr
  & Chase(A, B) $\Rightarrow$ Get\_Stormy\_ember(A, 2)
  & If A chases B, A will get 2 stormy embers. \\
  & Action2State
  & Scratch(A, B) $\Rightarrow$ Completely\_numb(A)
  & If A scratches B, A will be completely numb. \\
  & Action2State
  & Scratch(A, B) $\Rightarrow$ Deeply\_glowing(B)
  & If A scratches B, B will be deeply glowing. \\
\midrule

Environment
  & Env2State
  & Desert(A) $\Rightarrow$ Slightly\_disappointed(A)
  & If A is in desert, A will be slightly disappointed. \\
\midrule

State
  & State2Attr
  & Deeply\_hungry(A) $\Rightarrow$ Lose\_Crystalline\_tongue(A, 1)
  & If A is deeply hungry, A will lose 1 crystalline tongue. \\
\bottomrule

\end{tabular}
}
\caption{Examples of abstract procedural rules represented in FOL and NL formats.}
\label{tab:rule_examples}
\vspace{-0.5em}
\end{table*}

\subsection{Data Examples}
\label{appendix:RuleWorld-Data-Samples}
Table~\ref{tab:rule_examples} shows representative examples covering all major rule types, where each rule is expressed in both FOL and NL forms. The FOL version provides a precise symbolic specification of entity attributes, state changes, environmental effects, and action-induced transitions, while the NL version preserves the same semantics in human-readable language. For action-triggered effects, RuleWorld further distinguishes changes applied to the acting entity from those applied to the target entity, enabling asymmetric interactions such as ``A scratches B'' to produce different outcomes for A and B. These paired forms constitute the atomic rule units underlying all RuleWorld QA tasks.

\begin{figure}[htbp]
    \centerline{\includegraphics[scale=0.25]{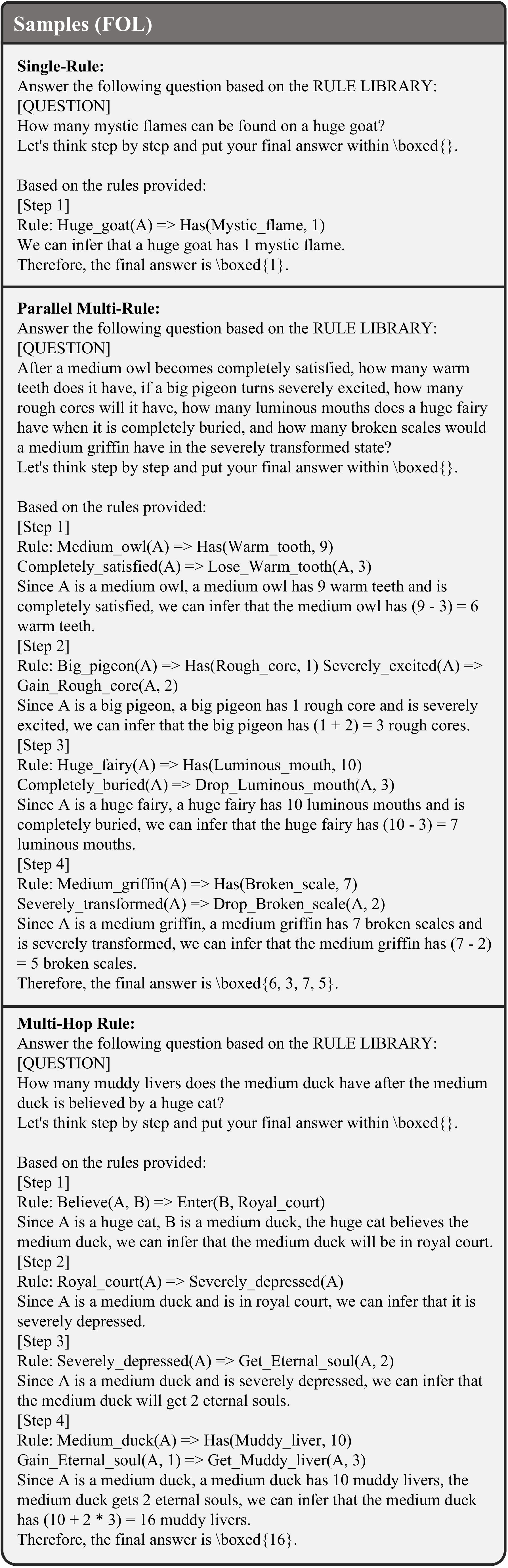}}%
    \caption{Examples of Single-Rule, Parallel Multi-Rule, and Multi-Hop Rule QA under FOL representation.}
    \label{fig:QA_template_samples_fol}
\end{figure}

\begin{figure}[htbp]
    \centerline{\includegraphics[scale=0.25]{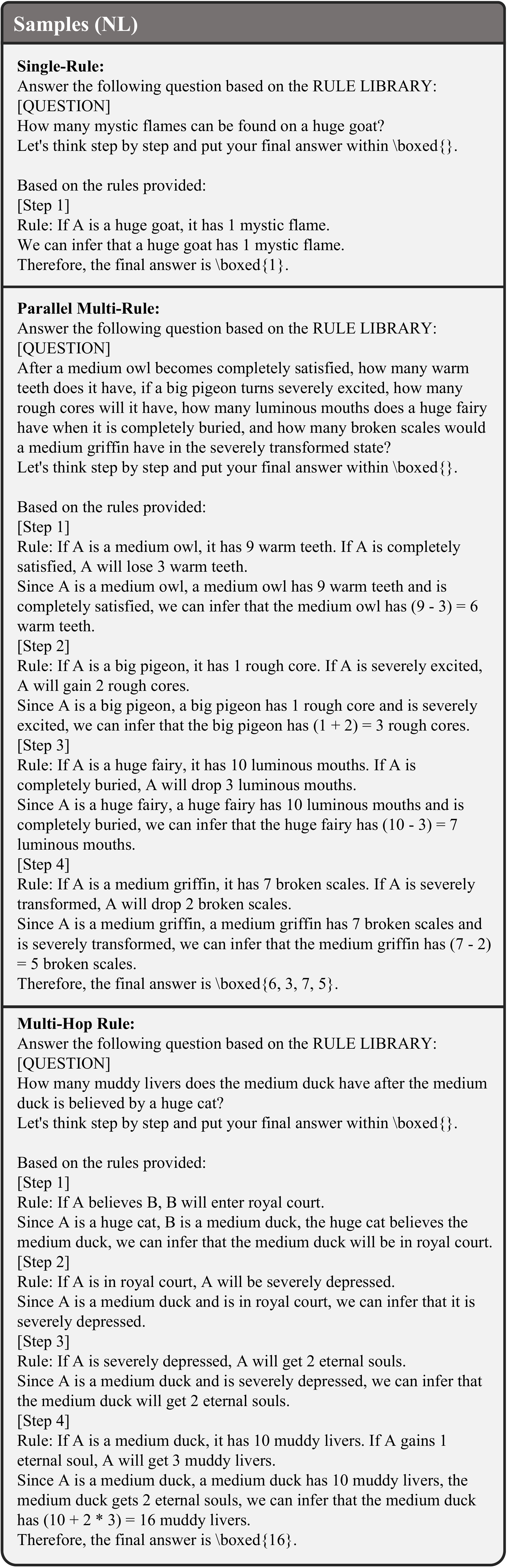}}%
    \caption{Examples of Single-Rule, Parallel Multi-Rule, and Multi-Hop Rule QA under NL representation.}
    \label{fig:QA_template_samples_nl}
\end{figure}

Figures~\ref{fig:QA_template_samples_fol} and~\ref{fig:QA_template_samples_nl} present representative examples of the three QA types: Single-Rule, Parallel Multi-Rule, and Multi-Hop Rule QA. Single-Rule questions require one atomic inference. Parallel Multi-Rule QA combines multiple independent reasoning threads in a single prompt, where each sub-question corresponds to one rule application step. Multi-Hop QA requires sequential reasoning over causally connected rules, where one inferred result triggers the next. Together, these examples illustrate how atomic rules can be composed into diverse reasoning trajectories, creating a large and structured space of QA instances for evaluating rule retrieval and application.

These examples also reflect the training format used in our Stacked Step-Level Attention Training. After each \texttt{[Step]} explanation, we insert a special token \texttt{<search>} to guide the model toward retrieving the rule needed for the next step. This encourages clear step-level attention separation and supports learning step-level rule application during autoregressive generation.

\section{Training and Evaluation Settings}
\label{appendix:Training_and_Evaluation_Settings}
\begin{figure}[htbp]
    \centerline{\includegraphics[scale=0.25]{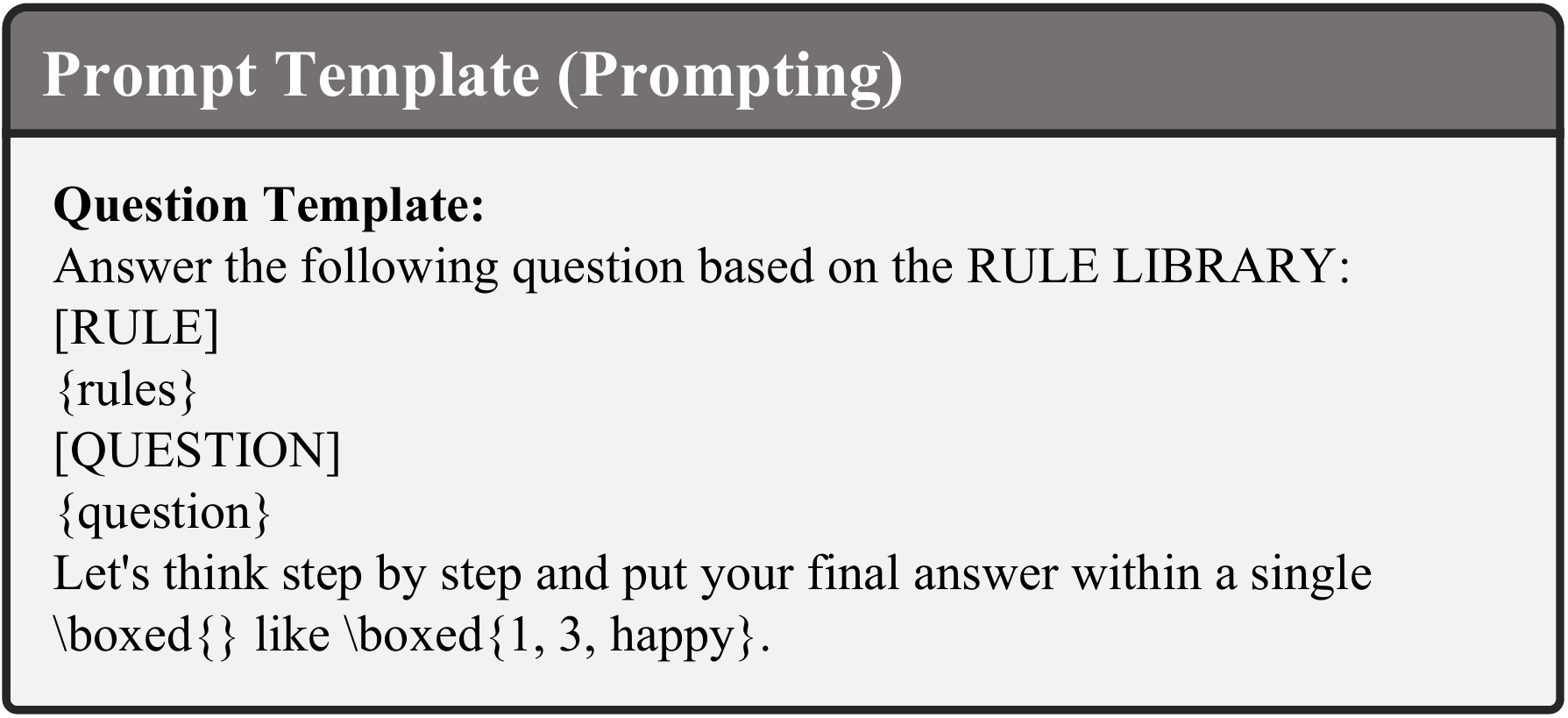}}%
    \caption{Question-answering template for prompting evaluation.}
    \label{fig:QA_template_icl}
\end{figure}

\paragraph{Training Settings}
We freeze the base model and randomly initialize $\mathbf{\tilde{W}}^l_{K}$ and $\mathbf{\tilde{W}}^l_{V}$, while copying $\mathbf{W}^l_{Q}$ to $\mathbf{\tilde{W}}^l_{Q}$. Training is performed on 4 A800 80GB GPUs using DeepSpeed ZeRO-2~\cite{rajbhandari2020zeromemoryoptimizationstraining} with CPU offloading and bf16 precision. We inject up to 100 rules for confidence layer identification in the first stage and 1000 rules for stacked step-level attention training, using top-$100$ selection. In the latter stage, we set $\mathcal{T}=0.05$, add \texttt{<search>} to the tokenizer and the model’s special-token list, resize the token embedding matrix accordingly, and unfreeze the embedding layer so that the \texttt{<search>} embedding can be learned. 
For end-to-end baselines, we use the model architectures provided in the released official implementation repository and train them with the same training configuration.
The per-device batch size is 10 with gradient accumulation 5 (effective batch size 50 per device; global batch size 200 across 4 GPUs), optimized via a cosine scheduler (learning rate $1\times10^{-4}$, warm-up ratio $1\times10^{-2}$, weight decay $1\times10^{-4}$). For training, we construct a RuleWorld subset containing 100K rules and 110K QA instances, as shown in Tables~\ref{tab:rules_full_and_training_and_evaluation} and~\ref{tab:qa_type_whole_training} in Appendix~\ref{appendix:RuleWorld-Data-Statistics}. Each batch consists of 30\% Single-Rule, 40\% Parallel Multi-Rule, and 30\% Multi-Hop instances to ensure balanced reasoning coverage.
\paragraph{Evaluation Settings}
We use the following evaluation settings:

\begin{itemize}
    \item \textbf{Evaluation data.}
    All evaluations use five random seeds, with 10 samples per task type across difficulty levels. This yields 110 samples per seed and 550 questions in total. The sampled parallel multi-rule instances are manually checked to ensure correct question combination. We use the same underlying rule pool as in training, but all QA instances come from a held-out set that is not used during training.

    \item \textbf{Rule injection and retrieval.}
    When the rule set size is 100, we inject all rules. For larger rule set sizes, we apply top-$K{=}100$ selection. KBLaM and the prompting baseline inject all rules without retrieval.

    \item \textbf{Baselines.}
    For RAG baselines, we use BM25~\cite{robertson1995okapi} and Qwen3-Embedding-8B as retrievers. End-to-end approaches perform confidence-layer retrieval. For hybrid search, we use Reciprocal Rank Fusion (RRF)~\cite{Reciprocal_rank_fusion_outperforms_condorcet_and_individual_rank_learning_methods} with $k = 60$ to fuse BM25 and dense retrieval results.

    \item \textbf{Answer evaluation.}
    Performance is measured by exact match accuracy on the comma-separated answer list in \texttt{\string\boxed\{\}}. After normalization, including lowercasing, removing LaTeX or string artifacts, and stripping extra whitespace, an instance scores 0 if the predicted and gold lists differ in length. Otherwise, its score is the fraction of position-wise elements that exactly match the gold answers.

    \item \textbf{Step-wise retrieval evaluation.}
    For DynaRule's multi-step retrieval, we compute step-wise recall by matching each retrieval step to the corresponding gold step in the reference solution. For example, the first retrieval is evaluated against the Step 1 gold rules. Missing steps are assigned a recall of 0, and extra steps are truncated.

    \item \textbf{Inference setup.}
    For inference, we set temperature to 0.6 for Qwen3-32B and 0 for other models. The QA template for prompting, which is also used for RAG, is provided in Figure~\ref{fig:QA_template_icl}.
\end{itemize}

\section{Extended Experiments and Analysis}
\label{appendix:extended_experiments}
\subsection{Confidence Layer Identification across Different Model Sizes, Series and Encoders}
\label{appendix:extended_cl_identification}

\begin{figure}[tbp]
  \centering
  \begin{subfigure}[t]{0.485\textwidth}
    \centering
    \includegraphics[width=\linewidth]{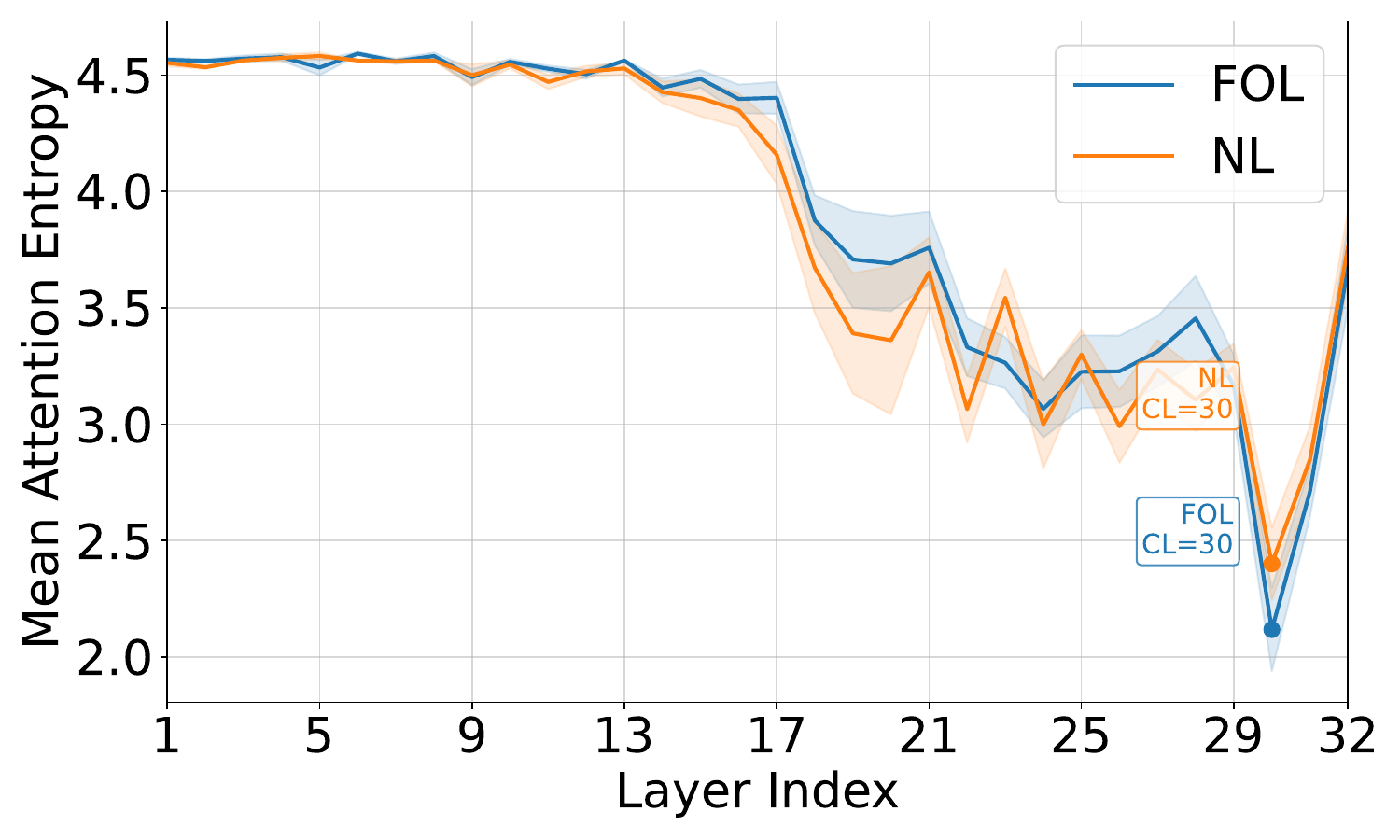}
    \caption{CL identification in Llama-3-8B-Instruct, with layer 30 identified as the CL.}
    \label{fig:llama_3_8b_instruct_entropy_plot}
  \end{subfigure}
  \begin{subfigure}[t]{0.485\textwidth}
    \centering
    \includegraphics[width=\linewidth]{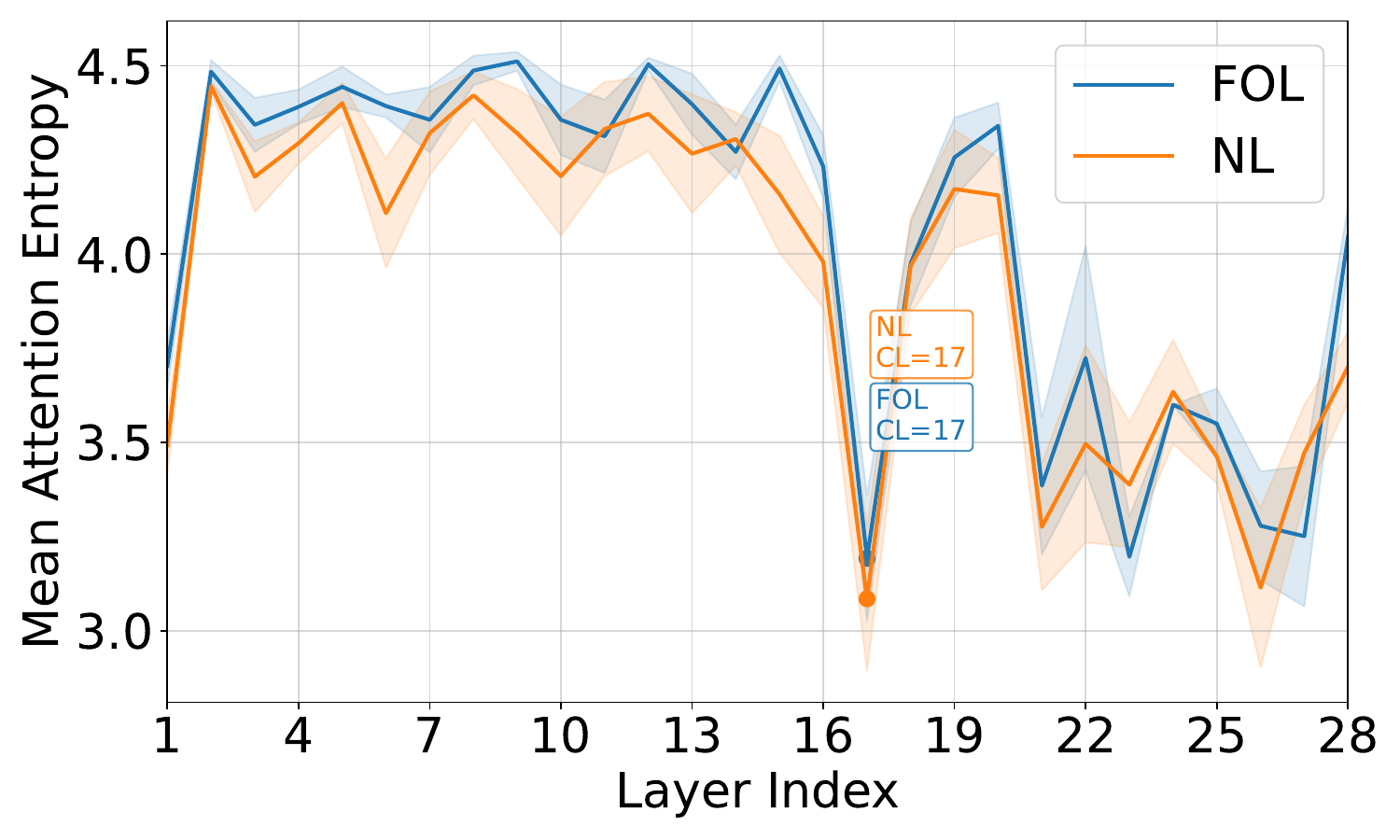}
    \caption{CL identification in Qwen2.5-1.5B-Instruct, with layer 17 identified as the CL.}
    \label{fig:qwen2.5_1.5b_instruct_entropy_plot}
  \end{subfigure}
  \begin{subfigure}[t]{0.485\textwidth}
    \centering
    \includegraphics[width=\linewidth]{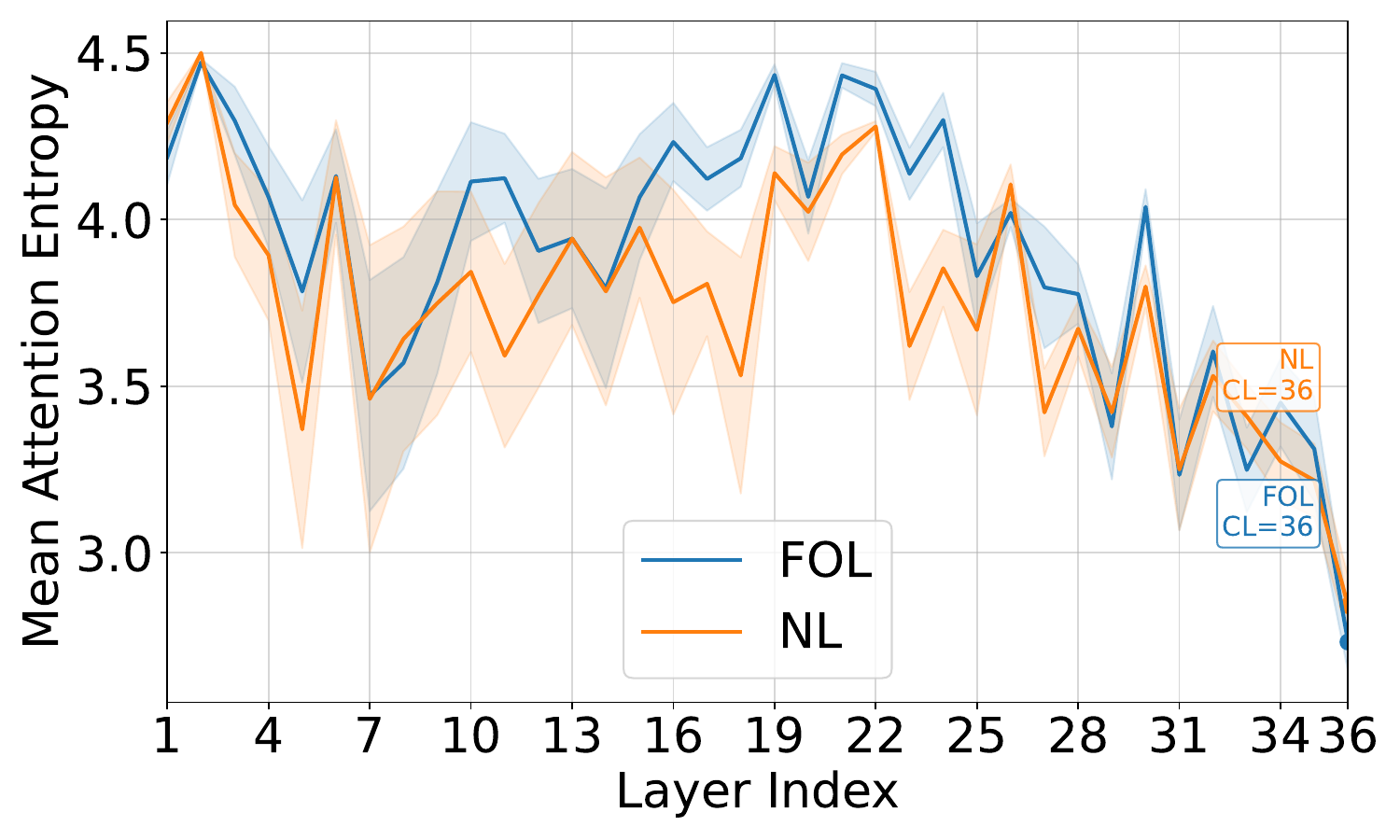}
    \caption{CL identification in Qwen2.5-3B-Instruct, with layer 36 identified as the CL.}
    \label{fig:qwen2.5_3b_instruct_entropy_plot}
  \end{subfigure}
  \begin{subfigure}[t]{0.485\textwidth}
    \centering
    \includegraphics[width=\linewidth]{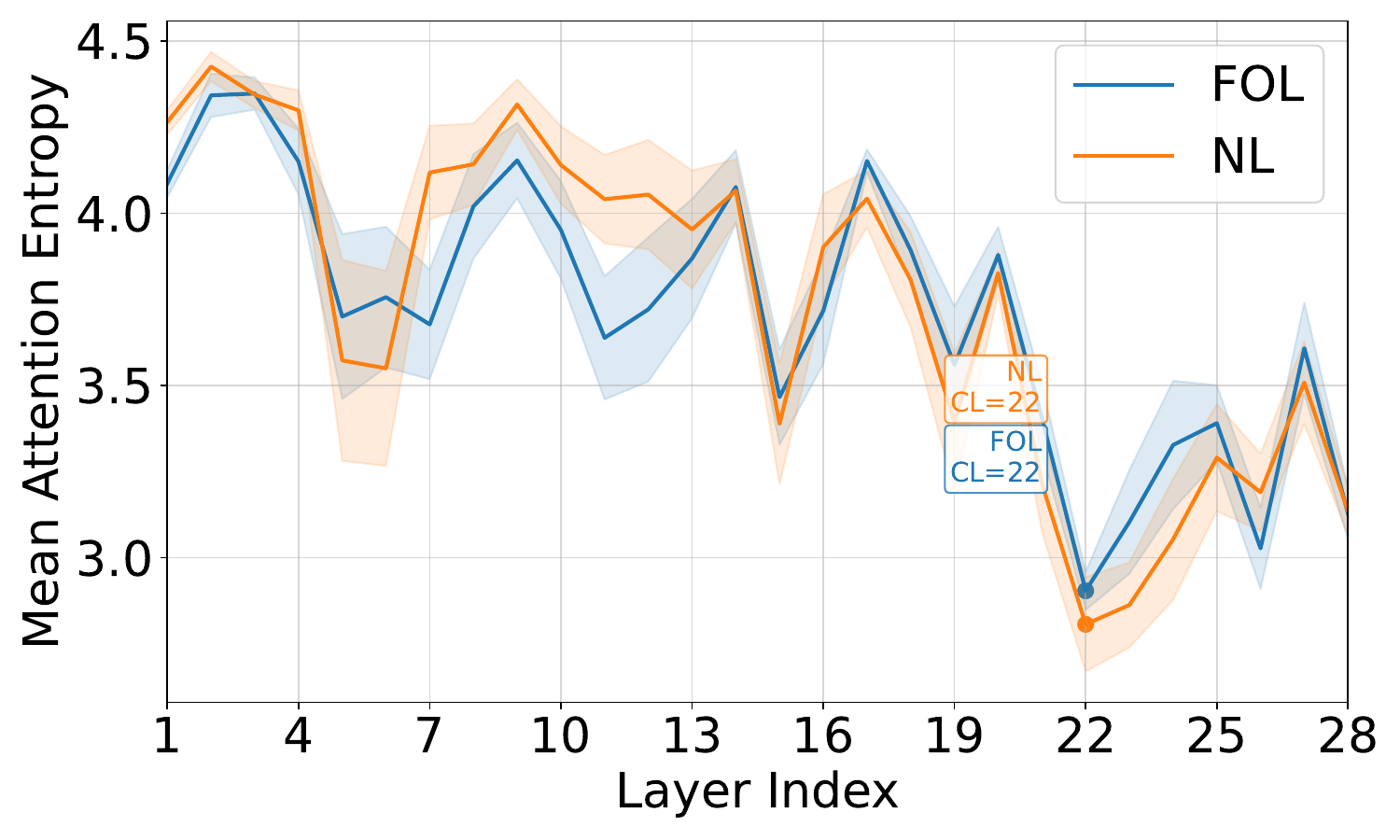}
    \caption{CL identification with bge-m3 as encoder, with layer 22 identified as the CL.}
    \label{fig:qwen2.5_7b_bgem3_instruct_entropy_plot}
  \end{subfigure}
  \caption{Extended confidence layer (CL) identification under FOL and NL rule representations. Shaded regions denote entropy standard deviation.}
  \label{fig:extended_entropy_plot}
\end{figure}

In this experiment, we identify the confidence layer by injecting 100 rules into different base models, including Llama-3-8B-Instruct, Qwen2.5-1.5B-Instruct, and Qwen2.5-3B-Instruct, using Qwen3-Embedding-8B as the encoder. Across both FOL and NL rule formats, as shown in Figures ~\ref{fig:llama_3_8b_instruct_entropy_plot}, 
\ref{fig:qwen2.5_1.5b_instruct_entropy_plot}, 
\ref{fig:qwen2.5_3b_instruct_entropy_plot}, all models consistently exhibit a clear confidence layer, and notably, each model’s confidence layer emerges at the same depth for both formats, specifically at layers 30, 17, and 36 respectively. We further evaluate Qwen2.5-7B-Instruct using bge-m3 as the text encoder and observe that the confidence layer again appears at an identical position under FOL and NL rule injections, consistently located at layer 22, as demonstrated in Figure~\ref{fig:qwen2.5_7b_bgem3_instruct_entropy_plot}. These results demonstrate that KV-cache-based rule injection reliably yields a stable and format-invariant confidence layer across different model sizes, model series, and encoders, highlighting the robustness of our method.

\subsection{Prompting Results with Strong LLMs}
\label{appendix:in-context-results}
Across both the FOL and NL settings, the prompting results in Table~\ref{tab:extended_icl_results} reveal four main findings.

\textbf{(1)} Existing LLMs struggle to maintain stable reasoning performance as the number of injected rules increases. While several models perform reasonably under 100 rules, their accuracy drops substantially at 500 and 1000 rules, especially on parallel multi-rule reasoning and 3--4 hop multi-hop QA. 
\textbf{(2)} Under most settings, claude-sonnet-4-6 is the strongest prompting baseline overall, particularly on multi-hop reasoning, while Qwen3-32B also demonstrates highly competitive performance under 100-rule setting. As discussed in Appendix~\ref{appendix:case-study}, this advantage appears to come from more faithful operator-level rule following and fewer missed rules on long reasoning chains.
\textbf{(3)} As the rule scale grows to 500 and 1000, most models show pronounced degradation, and Qwen3-32B no longer maintains its small-scale advantage. In these higher-load settings, GPT5.5 and claude-sonnet-4-6 become the strongest baselines, suggesting better robustness to large candidate rule pools.
\textbf{(4)} Despite the strength of these large prompting baselines, DynaRule consistently achieves the best overall results across all rule scales and both representations. It surpasses the strongest baseline by up to 13.86 points on average and by up to 36 points on the best subtask, demonstrating the effectiveness of step-level dynamic rule integration.

\begin{figure*}[tbp]
    \centerline{\includegraphics[scale=0.425]{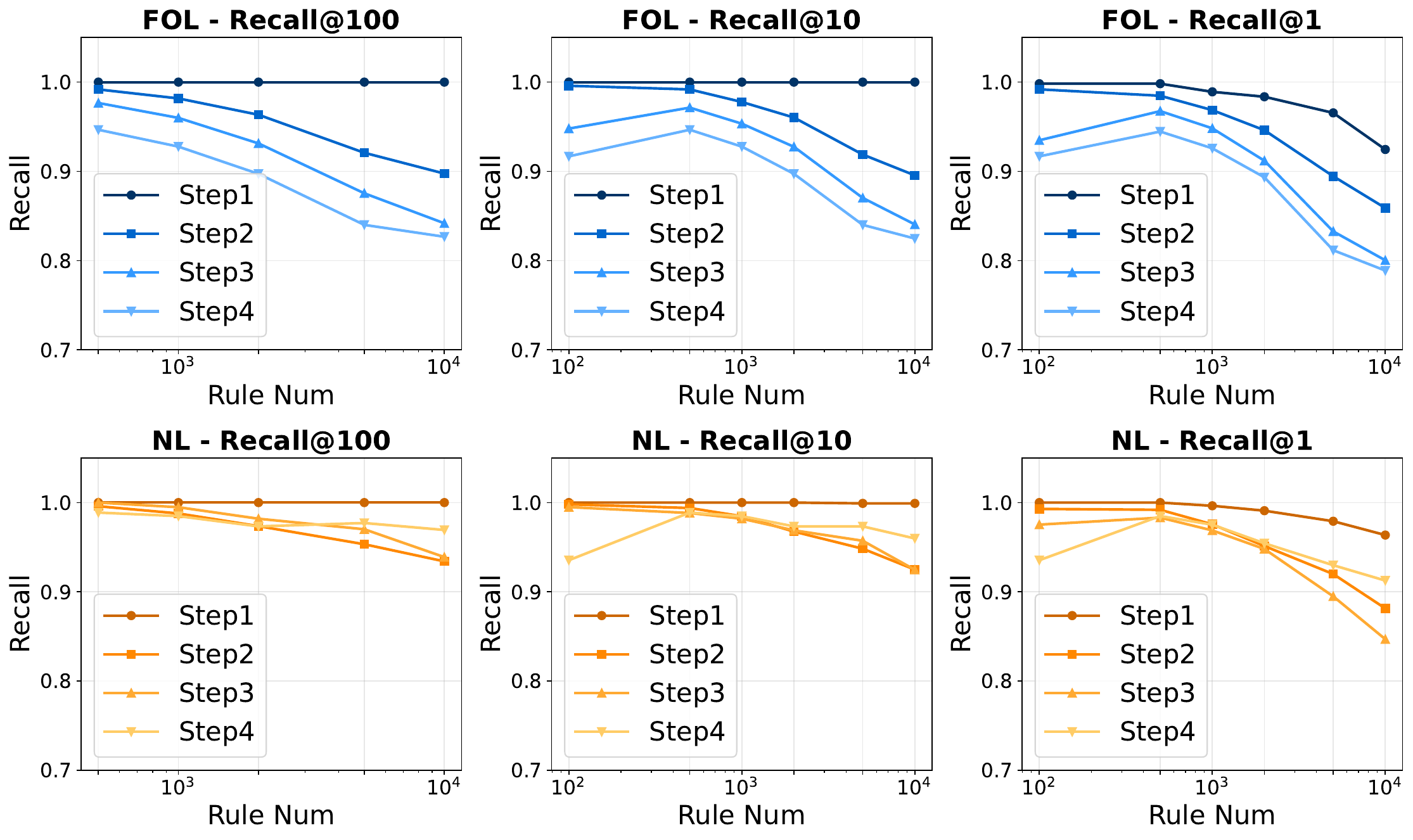}}%
    \caption{Retrieval performance in terms of Recall@100, Recall@10, and Recall@1 across different reasoning steps and numbers of injected rules under both FOL and NL representations.}
    \label{fig:step_retrieval_performance}
\end{figure*}

\begin{table*}
    \centering 
    \setlength{\tabcolsep}{10.3pt} 
    \scalebox{0.48}{%
      \begin{tabular}{%
        c
        *{1}{c}
        *{7}{c}
        *{3}{c}
        c
      }
      \toprule
      \multirow{2}{*}{Method}
        & \multicolumn{1}{c}{\textbf{Single-Rule QA}}
        & \multicolumn{7}{c}{\textbf{Parallel Multi-Rule QA}}
        & \multicolumn{3}{c}{\textbf{Multi-Hop Rule QA}}
        & \multirow{2}{*}{\textbf{Avg.}} \\
      \cmidrule(lr){2-2}
      \cmidrule(lr){3-9}
      \cmidrule(lr){10-12}
        & \multicolumn{1}{c}{single-rule}
        & \multicolumn{1}{c}{multi-rule-2}
        & \multicolumn{1}{c}{multi-rule-3}
        & \multicolumn{1}{c}{multi-rule-4}
        & \multicolumn{1}{c}{multi-rule-5}
        & \multicolumn{1}{c}{multi-rule-6}
        & \multicolumn{1}{c}{multi-rule-7}
        & \multicolumn{1}{c}{multi-rule-8}
        & \multicolumn{1}{c}{multi-hop-2}
        & \multicolumn{1}{c}{multi-hop-3}
        & \multicolumn{1}{c}{multi-hop-4}
        & \\  
      \midrule
      \multicolumn{13}{c}{\textit{Rule Num = 100}} \\ \midrule
      Prompting & 0.8200 & \underline{0.7000} & 0.7367 & 0.7800 & 0.6983 & 0.5183 & 0.4650 & 0.2850 & 0.5000 & 0.0600 & 0.0400 & 0.5094 \\
          KBLaM & \underline{0.9800} & \textbf{1.0000} & \underline{0.9817} & \textbf{1.0000} & \textbf{1.0000} & \underline{0.9833} & \textbf{0.9750} & \underline{0.9900} & \textbf{1.0000} & 0.9000 & \underline{0.9400} & \underline{0.9773} \\
          SR-KI & \textbf{1.0000} & \textbf{1.0000} & \textbf{0.9950} & 0.9700 & \underline{0.9950} & 0.9550 & \textbf{0.9750} & 0.9750 & \underline{0.9800} & \underline{0.9200} & 0.9000 & 0.9695 \\
                    \rowcolor{blue!7} DynaRule & \textbf{1.0000} & \textbf{1.0000} & \textbf{0.9950} & \underline{0.9800} & \underline{0.9950} & \textbf{0.9850} & \underline{0.9700} & \textbf{0.9950} & 0.9600 & \textbf{0.9300} & \textbf{0.9500} & \textbf{0.9782} \\
          \midrule
      \multicolumn{13}{c}{\textit{Rule Num = 500}} \\ \midrule
      Prompting & 0.7400 & 0.6900 & 0.6867 & 0.7583 & 0.5333 & 0.4383 & 0.2700 & 0.1500 & 0.3400 & 0.0400 & 0.0000 & 0.4224 \\
          $\text{RAG}_{\text{dense}}$ & 0.7800 & 0.5100 & 0.4350 & 0.3467 & 0.3317 & 0.2450 & 0.2200 & 0.1750 & 0.0200 & 0.0200 & 0.0400 & 0.2839 \\
          $\text{RAG}_{\text{bm25}}$ & 0.6400 & 0.5300 & 0.6267 & 0.5400 & 0.5717 & 0.3283 & 0.3300 & 0.2950 & 0.0200 & 0.0200 & 0.0200 & 0.3565 \\
          $\text{RAG}_{\text{hybrid}}$ & 0.6800  & 0.7100  & 0.6250  & 0.5217  & 0.5583  & 0.4517  & 0.3400  & 0.2500  & 0.0400  & 0.0200  & 0.0200  & 0.3833  \\
          KBLaM & 0.9600 & \underline{0.9900} & 0.9333 & \underline{0.9550} & 0.9683 & \underline{0.9733} & \underline{0.9500} & \underline{0.9650} & \underline{0.9200} & \underline{0.6200} & \underline{0.6400} & \underline{0.8977} \\
          SR-KI & \underline{0.9800} & 0.9700 & \underline{0.9817} & \underline{0.9550} & \underline{0.9733} & 0.9300 & 0.9250 & 0.9400 & 0.8000 & 0.3400 & 0.2600 & 0.8232 \\
                    \rowcolor{blue!7} DynaRule & \textbf{1.0000} & \textbf{1.0000} & \textbf{0.9867} & \textbf{0.9950} & \textbf{0.9833} & \textbf{1.0000} & \textbf{0.9650} & \textbf{0.9850} & \textbf{0.9800} & \textbf{0.7200} & \textbf{0.8000} & \textbf{0.9468} \\
          \midrule
      \multicolumn{13}{c}{\textit{Rule Num = 1000}} \\ \midrule
      Prompting & 0.8000 & 0.5900 & 0.5800 & 0.6933 & 0.4533 & 0.3133 & 0.2150 & 0.0800 & 0.2800 & 0.0400 & 0.0000 & 0.3677 \\
          $\text{RAG}_{\text{dense}}$ & 0.6600 & 0.4400 & 0.3433 & 0.2933 & 0.2467 & 0.2100 & 0.2050 & 0.1500 & 0.0400 & 0.0200 & 0.0200 & 0.2389 \\
          $\text{RAG}_{\text{bm25}}$ & 0.5200 & 0.6100 & 0.5600 & 0.4750 & 0.4633 & 0.2717 & 0.2250 & 0.1700 & 0.0000 & 0.0000 & 0.0000 & 0.2995 \\
          $\text{RAG}_{\text{hybrid}}$ & 0.6200  & 0.6300  & 0.6600  & 0.5233  & 0.4467  & 0.3600  & 0.3250  & 0.2400  & 0.0000  & 0.0000  & 0.0200  & 0.3477 \\
          KBLaM & 0.9000 & 0.9500 & 0.8133 & 0.8284 & 0.8867 & \underline{0.9083} & \underline{0.8650} & \underline{0.8950} & \underline{0.8400} & \underline{0.4200} & \underline{0.4400} & \underline{0.7952} \\
           SR-KI & \underline{0.9800} & \underline{0.9600} & \underline{0.9750} & \underline{0.9100} & \underline{0.9433} & 0.8733 & 0.8450 & 0.8500 & 0.7800 & 0.2200 & 0.2800 & 0.7833 \\
                    \rowcolor{blue!7} DynaRule & \textbf{1.0000} & \textbf{1.0000} & \textbf{0.9917} & \textbf{0.9875} & \textbf{0.9792} & \textbf{0.9625} & \textbf{0.9812} & \textbf{0.9688} & \textbf{0.8500} & \textbf{0.6250} & \textbf{0.6500} & \textbf{0.9087} \\
          \midrule
      \multicolumn{13}{c}{\textit{Rule Num = 2000}} \\ \midrule
      Prompting & \underline{0.6800} & 0.4700 & 0.5267 & 0.5900 & 0.4717 & 0.2500 & 0.1950 & 0.0750 & 0.1000 & 0.0200 & 0.0400 & 0.3108 \\
          $\text{RAG}_{\text{dense}}$ & 0.6200 & 0.2700 & 0.3167 & 0.2200 & 0.2383 & 0.1783 & 0.1400 & 0.0900 & 0.0000 & 0.0000 & 0.0400 & 0.1921 \\
          $\text{RAG}_{\text{bm25}}$ & 0.5600 & 0.4700 & 0.4050 & 0.2850 & 0.2500 & 0.1783 & 0.1400 & 0.1400 & 0.0000 & 0.0000 & 0.0000 & 0.2208 \\
          $\text{RAG}_{\text{hybrid}}$ & 0.6000  & 0.6300  & 0.5633  & 0.4483  & 0.3500  & 0.3083  & 0.2850  & 0.1750  & 0.0200  & 0.0000  & 0.0000  & 0.3073  \\
          KBLaM & 0.0400 & 0.0700 & 0.0133 & 0.1300 & 0.0800 & 0.1100 & 0.0750 & 0.0450 & 0.0200 & 0.0200 & 0.0000 & 0.0548 \\
          SR-KI & \textbf{1.0000} & \underline{0.9300} & \underline{0.9050} & \underline{0.8650} & \underline{0.8583} & \underline{0.8217} & \underline{0.7750} & \underline{0.7750} & \underline{0.7000} & \underline{0.1600} & \underline{0.2000} & \underline{0.7264} \\
                    \rowcolor{blue!7} DynaRule & \textbf{1.0000} & \textbf{0.9800} & \textbf{0.9734} & \textbf{0.9833} & \textbf{0.9400} & \textbf{0.9400} & \textbf{0.9350} & \textbf{0.9350} & \textbf{0.8600} & \textbf{0.6000} & \textbf{0.4600} & \textbf{0.8733} \\
          \midrule
      \multicolumn{13}{c}{\textit{Rule Num = 5000}} \\ \midrule
      $\text{RAG}_{\text{dense}}$ & 0.5800 & 0.2700 & 0.2400 & 0.2167 & 0.1950 & 0.1067 & 0.0800 & 0.1000 & 0.0000 & 0.0000 & 0.0000 & 0.1626 \\
          $\text{RAG}_{\text{bm25}}$ & 0.5000 & 0.3800 & 0.2383 & 0.1250 & 0.1167 & 0.1117 & 0.1150 & 0.0750 & 0.0000 & 0.0000 & 0.0000 & 0.1511 \\
          $\text{RAG}_{\text{hybrid}}$ & 0.5800  & 0.4200  & 0.4466  & 0.3533  & 0.2950  & 0.2583  & 0.2600  & 0.1400  & 0.0400  & 0.0200  & 0.0400  & 0.2594  \\
          KBLaM & 0.0000 & 0.0000 & 0.0000 & 0.0000 & 0.0000 & 0.0000 & 0.0000 & 0.0000 & 0.0000 & 0.0000 & 0.0000 & 0.0000 \\
          SR-KI & \underline{0.9000} & \underline{0.8000} & \underline{0.7817} & \underline{0.7250} & \underline{0.6717} & \underline{0.6550} & \underline{0.6050} & \underline{0.5900} & \underline{0.5800} & \underline{0.1800} & \textbf{0.2200} & \underline{0.6099} \\
                    \rowcolor{blue!7} DynaRule & \textbf{0.9200} & \textbf{0.9400} & \textbf{0.9567} & \textbf{0.9567} & \textbf{0.9167} & \textbf{0.9150} & \textbf{0.8550} & \textbf{0.8650} & \textbf{0.6600} & \textbf{0.3200} & \underline{0.1800} & \textbf{0.7714} \\
          \midrule
      \multicolumn{13}{c}{\textit{Rule Num = 10000}} \\ \midrule
      $\text{RAG}_{\text{dense}}$ & 0.4000 & 0.3000 & 0.2367 & 0.1867 & 0.1617 & 0.1083 & 0.0800 & 0.0850 & 0.0000 & 0.0400 & 0.0000 & 0.1453 \\
          $\text{RAG}_{\text{bm25}}$ & 0.3400 & 0.2800 & 0.1667 & 0.1483 & 0.1267 & 0.1183 & 0.0850 & 0.0500 & 0.0200 & 0.0000 & 0.0000 & 0.1214 \\
          $\text{RAG}_{\text{hybrid}}$ & 0.4400  & 0.5000  & 0.4000  & 0.3067  & 0.2650  & 0.1500  & 0.1650  & 0.1400  & 0.0000  & 0.0200  & 0.0000 & 0.2170 \\
          KBLaM & 0.0000 & 0.0000 & 0.0000 & 0.0000 & 0.0000 & 0.0000 & 0.0000 & 0.0000 & 0.0000 & 0.0000 & 0.0000 & 0.0000 \\
          SR-KI & \underline{0.7400} & \underline{0.6800} & \underline{0.6300} & \underline{0.5483} & \underline{0.5550} & \underline{0.5333} & \underline{0.4600} & \underline{0.4300} & \underline{0.3800} & \textbf{0.2000} & \underline{0.1200} & \underline{0.4797} \\
                    \rowcolor{blue!7} DynaRule & \textbf{0.9200} & \textbf{0.9100} & \textbf{0.9300} & \textbf{0.9233} & \textbf{0.8417} & \textbf{0.8250} & \textbf{0.7550} & \textbf{0.7400} & \textbf{0.5200} & \underline{0.1600} & \textbf{0.1400} & \textbf{0.6968} \\
      \bottomrule
      \end{tabular}%
    }
    \caption{Exact match accuracy on RuleWorld (\textbf{FOL}), using Qwen2.5-7B-Instruct as the base model. $\text{RAG}_{\text{dense}}$ and $\text{RAG}_{\text{bm25}}$ use Qwen3-Embedding-8B and BM25 retrievers, respectively; $\text{RAG}_{\text{hybrid}}$ uses RRF with the fusion
constant set to 60 to fuse BM25 and dense retrieval results; suffixes denote the gold parallel rule count or hop number.}
    \label{tab:FOl_main_qa_results}
    \vspace{-0.6em}
  \end{table*}

\begin{table*}[t]
    \centering 
    \setlength{\tabcolsep}{10.3pt} 
    \scalebox{0.48 }{%
      \begin{tabular}{%
        c
        *{1}{c}
        *{7}{c}
        *{3}{c}
        c
      }
      \toprule
      \multirow{2}{*}{Method}
        & \multicolumn{1}{c}{\textbf{Single-Rule QA}}
        & \multicolumn{7}{c}{\textbf{Parallel Multi-Rule QA}}
        & \multicolumn{3}{c}{\textbf{Multi-Hop Rule QA}}
        & \multirow{2}{*}{\textbf{Avg.}} \\
      \cmidrule(lr){2-2}
      \cmidrule(lr){3-9}
      \cmidrule(lr){10-12}
        & \multicolumn{1}{c}{single-rule}
        & \multicolumn{1}{c}{multi-rule-2}
        & \multicolumn{1}{c}{multi-rule-3}
        & \multicolumn{1}{c}{multi-rule-4}
        & \multicolumn{1}{c}{multi-rule-5}
        & \multicolumn{1}{c}{multi-rule-6}
        & \multicolumn{1}{c}{multi-rule-7}
        & \multicolumn{1}{c}{multi-rule-8}
        & \multicolumn{1}{c}{multi-hop-2}
        & \multicolumn{1}{c}{multi-hop-3}
        & \multicolumn{1}{c}{multi-hop-4}
        & \\ 
      \midrule
      \multicolumn{13}{c}{\textit{Rule Num = 100}} \\ \midrule
          Prompting & \underline{0.8200} & 0.8600 & 0.7833 & 0.8150 & 0.7617 & 0.5217 & 0.5350 & 0.4550 & 0.3400 & 0.1000 & 0.1200 & 0.5556 \\
          KBLaM & \textbf{1.0000} & \textbf{1.0000} & \underline{0.9950} & \underline{0.9800} & \textbf{0.9950} & \textbf{0.9850} & \underline{0.9700} & \textbf{0.9950} & \textbf{0.9600} & 0.4800 & \textbf{0.7200} & \underline{0.9164} \\
          SR-KI & \textbf{1.0000} & \underline{0.9800} & \textbf{1.0000} & 0.9450 & \textbf{0.9950} & 0.9583 & 0.9200 & 0.9600 & \textbf{0.9600} & \underline{0.5400} & 0.5600 & 0.8926 \\
                    \rowcolor{blue!7} DynaRule & \textbf{1.0000} & \textbf{1.0000} & \underline{0.9950} & \textbf{0.9950} & \underline{0.9900} & \underline{0.9800} & \textbf{0.9750} & \underline{0.9850} & \underline{0.9200} & \textbf{0.9600} & \underline{0.7000} & \textbf{0.9545} \\
       \midrule
      \multicolumn{13}{c}{\textit{Rule Num = 500}} \\ \midrule
          Prompting & 0.7400 & 0.7500 & 0.7367 & 0.7217 & 0.5550 & 0.3733 & 0.2700 & 0.1950 & 0.1600 & 0.0400 & 0.0200 & 0.4147 \\
          $\text{RAG}_{\text{dense}}$ & 0.8600 & 0.6300 & 0.6150 & 0.4716 & 0.4983 & 0.2483 & 0.3450 & 0.3000 & 0.0600 & 0.0400 & 0.0000 & 0.3698 \\
          $\text{RAG}_{\text{bm25}}$ & 0.7600 & 0.4500 & 0.3000 & 0.3183 & 0.4217 & 0.2433 & 0.2100 & 0.1750 & 0.0400 & 0.0400 & 0.0200 & 0.2708 \\
           $\text{RAG}_{\text{hybrid}}$ & 0.8400  & 0.7200  & 0.5850  & 0.4967  & 0.4983  & 0.3717  & 0.3900  & 0.3300  & 0.1400  & 0.0000  & 0.0200  & 0.3992  \\
          KBLaM & 0.9600 & \underline{0.9700} & 0.9183 & \underline{0.9133} & 0.9483 & \underline{0.9600} & \textbf{0.9650} & \textbf{0.9750} & \underline{0.8200} & \underline{0.5600} & \underline{0.7600} & \underline{0.8864} \\
          SR-KI & \textbf{1.0000} & 0.9500 & \underline{0.9300} & 0.9067 & \underline{0.9600} & 0.8983 & 0.8700 & \underline{0.8800} & 0.6600 & 0.2600 & 0.1200 & 0.7668 \\
                    \rowcolor{blue!7} DynaRule & \underline{0.9800} & \textbf{0.9800} & \textbf{0.9950} & \textbf{1.0000} & \textbf{0.9783} & \textbf{0.9800} & \underline{0.9600} & \textbf{0.9750} & \textbf{0.8800} & \textbf{0.7000} & \textbf{0.8200} & \textbf{0.9317} \\
      \midrule
      \multicolumn{13}{c}{\textit{Rule Num = 1000}} \\ \midrule
          Prompting & 0.6200 & 0.6300 & 0.6500 & 0.6617 & 0.5333 & 0.2833 & 0.2200 & 0.0800 & 0.0600 & 0.0200 & 0.0000 & 0.3417 \\
          $\text{RAG}_{\text{dense}}$ & \underline{0.8800} & 0.6900 & 0.6100 & 0.4250 & 0.3333 & 0.1800 & 0.2000 & 0.2400 & 0.0200 & 0.0000 & 0.0000 & 0.3253 \\
          $\text{RAG}_{\text{bm25}}$ & 0.5800 & 0.4100 & 0.3400 & 0.3817 & 0.3950 & 0.2617 & 0.1900 & 0.1050 & 0.0000 & 0.0000 & 0.0000 & 0.2421 \\
          $\text{RAG}_{\text{hybrid}}$ & 0.8200  & 0.5800  & 0.5133  & 0.4967  & 0.3667  & 0.2650  & 0.2800  & 0.3500  & 0.1600  & 0.0000  & 0.0200  & 0.3502  \\
          KBLaM & 0.7800 & 0.8100 & 0.8383 & 0.8333 & 0.8683 & \underline{0.8833} & \underline{0.8700} & \underline{0.8700} & \underline{0.6800} & \underline{0.3400} & \underline{0.4200} & \underline{0.7448} \\
          SR-KI & \textbf{0.9800} & \underline{0.8700} & \underline{0.9033} & \underline{0.8567} & \underline{0.9133} & 0.8417 & 0.7700 & 0.7800 & 0.5800 & 0.1400 & 0.0400 & 0.6977 \\
                    \rowcolor{blue!7} DynaRule & \textbf{0.9800} & \textbf{0.9800} & \textbf{0.9817} & \textbf{0.9883} & \textbf{0.9783} & \textbf{0.9783} & \textbf{0.9550} & \textbf{0.9450} & \textbf{0.8400} & \textbf{0.6400} & \textbf{0.6800} & \textbf{0.9042} \\
      \midrule
      \multicolumn{13}{c}{\textit{Rule Num = 2000}} \\ \midrule
          Prompting & 0.5600 & 0.5500 & 0.6833 & 0.5950 & 0.3600 & 0.3050 & 0.1700 & 0.0700 & 0.0400 & 0.0400 & 0.0200 & 0.3085 \\
          $\text{RAG}_{\text{dense}}$ & 0.7600 & 0.5700 & 0.4500 & 0.4117 & 0.2517 & 0.1233 & 0.1700 & 0.1500 & 0.0400 & 0.0000 & 0.0400 & 0.2697 \\
          $\text{RAG}_{\text{bm25}}$ & 0.7000 & 0.3700 & 0.2700 & 0.3433 & 0.3667 & 0.1867 & 0.1450 & 0.0800 & 0.0400 & 0.0200 & 0.0200 & 0.2311 \\
          $\text{RAG}_{\text{hybrid}}$ & 0.8200  & 0.5100  & 0.5200  & 0.3967  & 0.4350  & 0.3067  & 0.2300  & 0.1900  & 0.1200  & 0.0600  & 0.0400  & 0.3299  \\
          KBLaM & 0.4400 & 0.5800 & 0.5200 & 0.5217 & 0.4783 & 0.4900 & 0.4950 & 0.4300 & 0.3400 & \underline{0.2000} & 0.1200 & 0.4195 \\
          SR-KI & \underline{0.9000} & \underline{0.8700} & \underline{0.7983} & \underline{0.7583} & \underline{0.7900} & \underline{0.7567} & \underline{0.6350} & \underline{0.6900} & \underline{0.5000} & 0.1200 & \underline{0.1400} & \underline{0.6326} \\
                    \rowcolor{blue!7} DynaRule & \textbf{0.9800} & \textbf{0.9000} & \textbf{0.9583} & \textbf{0.9667} & \textbf{0.9317} & \textbf{0.9517} & \textbf{0.9250} & \textbf{0.8600} & \textbf{0.7000} & \textbf{0.3600} & \textbf{0.5200} & \textbf{0.8230} \\
      \midrule
      \multicolumn{13}{c}{\textit{Rule Num = 5000}} \\ \midrule
          $\text{RAG}_{\text{dense}}$ & 0.7200 & 0.3800 & 0.3617 & 0.2600 & 0.2200 & 0.1150 & 0.1150 & 0.0650 & 0.0200 & 0.0200 & 0.0600 & 0.2124 \\
          $\text{RAG}_{\text{bm25}}$ & 0.5800 & 0.2900 & 0.2900 & 0.3183 & 0.2950 & 0.1767 & 0.1200 & 0.0650 & 0.0400 & 0.0000 & 0.0200 & 0.1995 \\
          $\text{RAG}_{\text{hybrid}}$ & 0.7000  & 0.4500  & 0.4133  & 0.3950  & 0.4633  & 0.2400  & 0.2250  & 0.1350  & 0.0400  & 0.0000  & 0.0000  & 0.2783  \\
          KBLaM & 0.0000 & 0.0300 & 0.0200 & 0.0067 & 0.0050 & 0.0100 & 0.0000 & 0.0000 & 0.0000 & 0.0200 & 0.0000 & 0.0083 \\
          SR-KI & \underline{0.8600} & \underline{0.7300} & \underline{0.6367} & \underline{0.5450} & \underline{0.5667} & \underline{0.5267} & \underline{0.5000} & \underline{0.5150} & \underline{0.2200} & \underline{0.0600} & \underline{0.0800} & \underline{0.4764} \\
                    \rowcolor{blue!7} DynaRule & \textbf{0.9400} & \textbf{0.8900} & \textbf{0.9050} & \textbf{0.9167} & \textbf{0.8733} & \textbf{0.8867} & \textbf{0.8300} & \textbf{0.7700} & \textbf{0.5800} & \textbf{0.1800} & \textbf{0.3800} & \textbf{0.7411} \\
       \midrule
      \multicolumn{13}{c}{\textit{Rule Num = 10000}} \\ \midrule
          $\text{RAG}_{\text{dense}}$ & 0.6600 & 0.3200 & 0.3500 & 0.1533 & 0.1667 & 0.0500 & 0.0800 & 0.0400 & 0.0000 & 0.0000 & \underline{0.1000} & 0.1745 \\
          $\text{RAG}_{\text{bm25}}$ & 0.6400 & 0.2600 & 0.2600 & 0.3250 & 0.3100 & 0.1550 & 0.1200 & 0.0700 & 0.0000 & 0.0000 & 0.0400 & 0.1982 \\
          $\text{RAG}_{\text{hybrid}}$ & 0.6000  & 0.4200  & 0.4033  & 0.3983  & 0.3600  & 0.1933  & 0.1550  & 0.1550  & 0.0200  & 0.0000  & 0.0400  & 0.2495 \\
          KBLaM & 0.0000 & 0.0000 & 0.0000 & 0.0000 & 0.0000 & 0.0000 & 0.0000 & 0.0000 & 0.0000 & 0.0000 & 0.0000 & 0.0000 \\
          SR-KI & \underline{0.7400} & \underline{0.6100} & \underline{0.5083} & \underline{0.4217} & \underline{0.4733} & \underline{0.4150} & \underline{0.3600} & \underline{0.3800} & \underline{0.1200} & \underline{0.0800} & 0.0400 & \underline{0.3771} \\
                    \rowcolor{blue!7} DynaRule & \textbf{0.8000} & \textbf{0.8600} & \textbf{0.8550} & \textbf{0.8650} & \textbf{0.7983} & \textbf{0.7767} & \textbf{0.6750} & \textbf{0.6650} & \textbf{0.6000} & \textbf{0.1600} & \textbf{0.2400} & \textbf{0.6632} \\
      \bottomrule
      \end{tabular}%
    }
    \caption{Exact match accuracy on RuleWorld (\textbf{NL}), using Qwen2.5-7B-Instruct as the base model. $\text{RAG}_{\text{dense}}$ and $\text{RAG}_{\text{bm25}}$ use Qwen3-Embedding-8B and BM25 retrievers, respectively; $\text{RAG}_{\text{hybrid}}$ uses RRF with the fusion
constant set to 60 to fuse BM25 and dense retrieval results; suffixes denote the gold parallel rule count or hop number.}
    \label{tab:NL_main_qa_results}
  \end{table*}

\begin{table}[t]
    \centering 
    \setlength{\tabcolsep}{8pt} 
    \scalebox{0.48}{%
      \begin{tabular}{%
        c
        *{3}{c}
        *{3}{c}
      }
      \toprule
      \multirow{2}{*}{Method}
        & \multicolumn{3}{c}{\textbf{First-order Logic}}
        & \multicolumn{3}{c}{\textbf{Natural Language}} \\
      \cmidrule(lr){2-4}
      \cmidrule(lr){5-7}
        & \multicolumn{1}{c}{Recall@100}
        & \multicolumn{1}{c}{Recall@10}
        & \multicolumn{1}{c}{Recall@1}
        & \multicolumn{1}{c}{Recall@100}
        & \multicolumn{1}{c}{Recall@10}
        & \multicolumn{1}{c}{Recall@1} \\ 
      \midrule
      \multicolumn{7}{c}{\textit{Rule Num = 100}} \\ \midrule
          Dense & - & 0.5325 & 0.4448 & - & 0.6646 & 0.5414 \\
          BM25 & - & 0.6901 & 0.4375 & - & 0.5869 & 0.5292 \\
          Hybrid & - & 0.6364  & 0.4793 & - & 0.5972  & 0.5128  \\
          KBLaM & - & 0.5151 & 0.4122 & - & 0.5278 & 0.4025 \\
          SR-KI & - & \underline{0.9100} & \underline{0.8000} & - & \underline{0.8963} & \underline{0.7768} \\
          \rowcolor{blue!7} DynaRule & - & \textbf{0.9773} & \textbf{0.9733} & - & \textbf{0.9909} & \textbf{0.9857} \\
      \midrule
      \multicolumn{7}{c}{\textit{Rule Num = 500}} \\ \midrule
          Dense & 0.6208 & 0.4079 & 0.3371 & 0.7700 & 0.5114 & 0.4039 \\
          BM25 & 0.8186 & 0.3934 & 0.2395 & 0.6247 & 0.5136 & 0.4680 \\
          Hybrid & 0.8399  & 0.4376  & 0.2974 & 0.8026  & 0.4965  & 0.4111 \\
          KBLaM & 0.6148 & 0.4137 & 0.2647 & 0.6215 & 0.4044 & 0.2452 \\
          SR-KI & \underline{0.9520} & \underline{0.8047} & \underline{0.6372} & \underline{0.9493} & \underline{0.7886} & \underline{0.6163} \\
          \rowcolor{blue!7} DynaRule & \textbf{0.9862} & \textbf{0.9852} & \textbf{0.9814} & \textbf{0.9973} & \textbf{0.9941} & \textbf{0.9920} \\
      \midrule
      \multicolumn{7}{c}{\textit{Rule Num = 1000}} \\ \midrule
          Dense & 0.5529 & 0.3635 & 0.3028 & 0.6961 & 0.4490 & 0.3587 \\
          BM25 & 0.7559 & 0.2962 & 0.1887 & 0.5946 & 0.4892 & 0.4424 \\
          Hybrid & 0.7935  & 0.3561  & 0.2456 & 0.7618  & 0.4464  & 0.3839  \\
          KBLaM & 0.5237 & 0.3532 & 0.1739 & 0.5389 & 0.3351 & 0.1686 \\
          SR-KI & \underline{0.9331} & \underline{0.7369} & \underline{0.5318} & \underline{0.9288} & \underline{0.7213} & \underline{0.5429} \\
          \rowcolor{blue!7} DynaRule & \textbf{0.9762} & \textbf{0.9737} & \textbf{0.9656} & \textbf{0.9932} & \textbf{0.9892} & \textbf{0.9810} \\
      \midrule
      \multicolumn{7}{c}{\textit{Rule Num = 2000}} \\ \midrule
          Dense & 0.4929 & 0.3192 & 0.2717 & 0.6293 & 0.3969 & 0.3150 \\
          BM25 & 0.5916 & 0.2273 & 0.1496 & 0.5792 & 0.4614 & 0.4182 \\
          Hybrid & 0.7167  & 0.2943  & 0.1929 & 0.7118  & 0.4086  & 0.3620  \\
          KBLaM & 0.4728 & 0.2532 & 0.1106 & 0.4734 & 0.2352 & 0.1068 \\
          SR-KI & \underline{0.9070} & \underline{0.6503} & \underline{0.4343} & \underline{0.8989} & \underline{0.6465} & \underline{0.4661} \\
          \rowcolor{blue!7} DynaRule & \textbf{0.9618} & \textbf{0.9595} & \textbf{0.9464} & \textbf{0.9848} & \textbf{0.9798} & \textbf{0.9640} \\
      \midrule
      \multicolumn{7}{c}{\textit{Rule Num = 5000}} \\ \midrule
          Dense & 0.4232 & 0.2789 & 0.2403 & 0.5282 & 0.3343 & 0.2575 \\
          BM25 & 0.4209 & 0.1548 & 0.0800 & 0.5258 & 0.4299 & \underline{0.3869} \\
          Hybrid & 0.5754  & 0.2570  & 0.1490 & 0.6370  & 0.3806  & 0.3226  \\
          KBLaM & 0.4231 & 0.1297 & 0.0559 & 0.4189 & 0.1289 & 0.0508 \\
          SR-KI & \underline{0.8491} & \underline{0.5168} & \underline{0.3145} & \underline{0.8261} & \underline{0.5402} & 0.3662 \\
          \rowcolor{blue!7} DynaRule & \textbf{0.9307} & \textbf{0.9286} & \textbf{0.8964} & \textbf{0.9745} & \textbf{0.9682} & \textbf{0.9346} \\
      \midrule
      \multicolumn{7}{c}{\textit{Rule Num = 10000}} \\ \midrule
          Dense & 0.3710 & 0.2573 & 0.2227 & 0.4619 & 0.2940 & 0.2241 \\
          BM25 & 0.3017 & 0.1102 & 0.0418 & 0.4970 & 0.4058 & \underline{0.3703} \\
          Hybrid & 0.5023  & 0.2344  & 0.1338 & 0.5876  & 0.3559  & 0.3023 \\
          KBLaM & 0.3884 & 0.0642 & 0.0239 & 0.3554 & 0.0776 & 0.0282 \\
          SR-KI & \underline{0.7758} & \underline{0.4180} & \underline{0.2415} & \underline{0.7539} & \underline{0.4465} & 0.2898 \\
          \rowcolor{blue!7} DynaRule & \textbf{0.9145} & \textbf{0.9130} & \textbf{0.8613} & \textbf{0.9614} & \textbf{0.9526} & \textbf{0.9045} \\
      \bottomrule
      \end{tabular}
    }
    \caption{Retrieval performance on FOL and NL rules. Dense uses Qwen3-Embedding-8B; Hybrid uses RRF (fusion constant 60) to combine BM25 and dense retrieval; KBLaM, SR-KI, and DynaRule use confidence-layer retrieval. For Recall@1, $K$ is set to the number of gold rules per instance or step.}
    \label{tab:retrieval_results}
  \end{table}

\begin{figure}[h]
  \centering
  \begin{subfigure}[t]{0.23\textwidth}
    \centering
    \includegraphics[width=\linewidth]{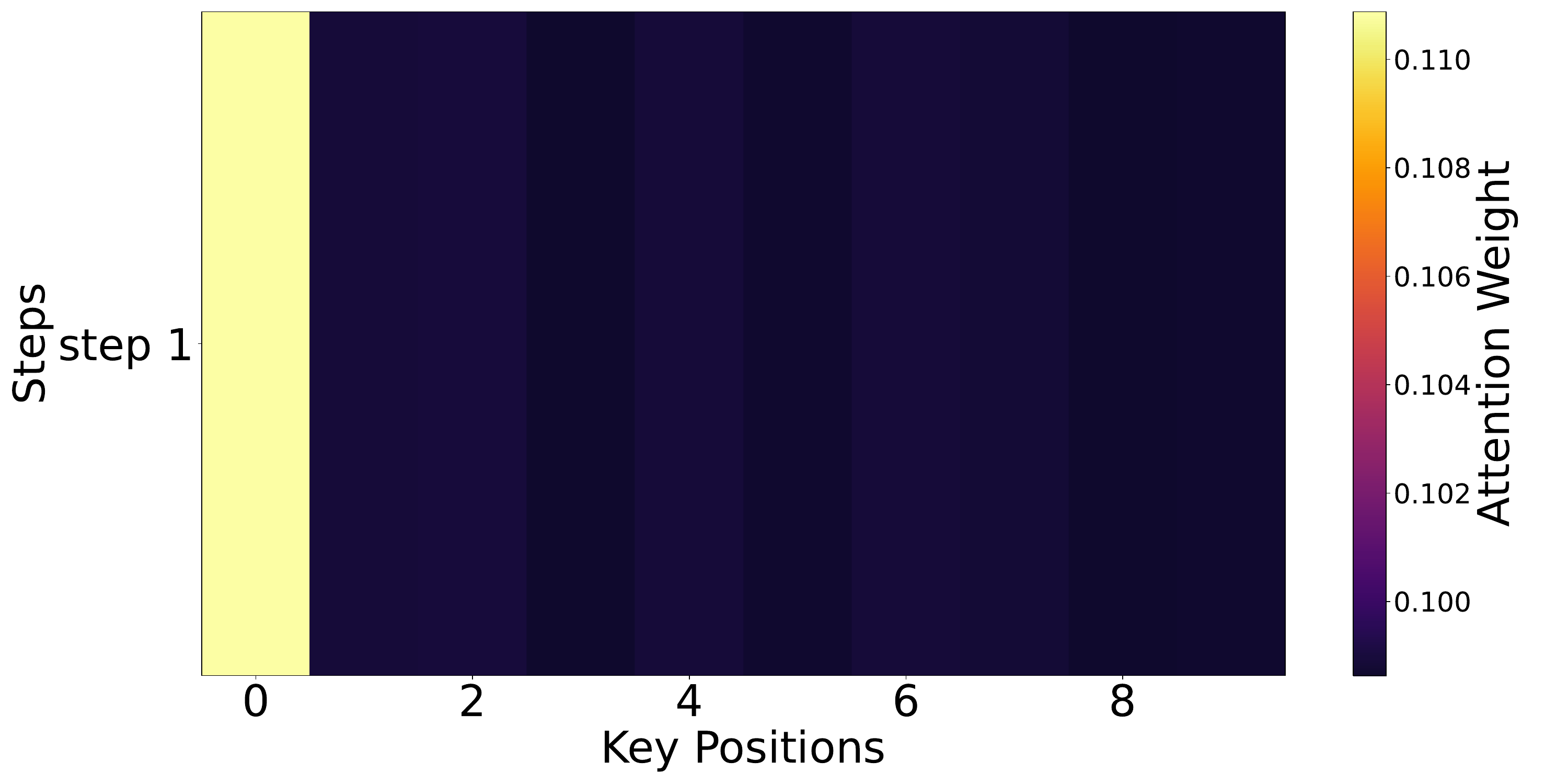}
    \caption{1-Step attention heatmap.}
    \label{fig:step_1}
  \end{subfigure}
  \begin{subfigure}[t]{0.23\textwidth}
    \centering
    \includegraphics[width=\linewidth]{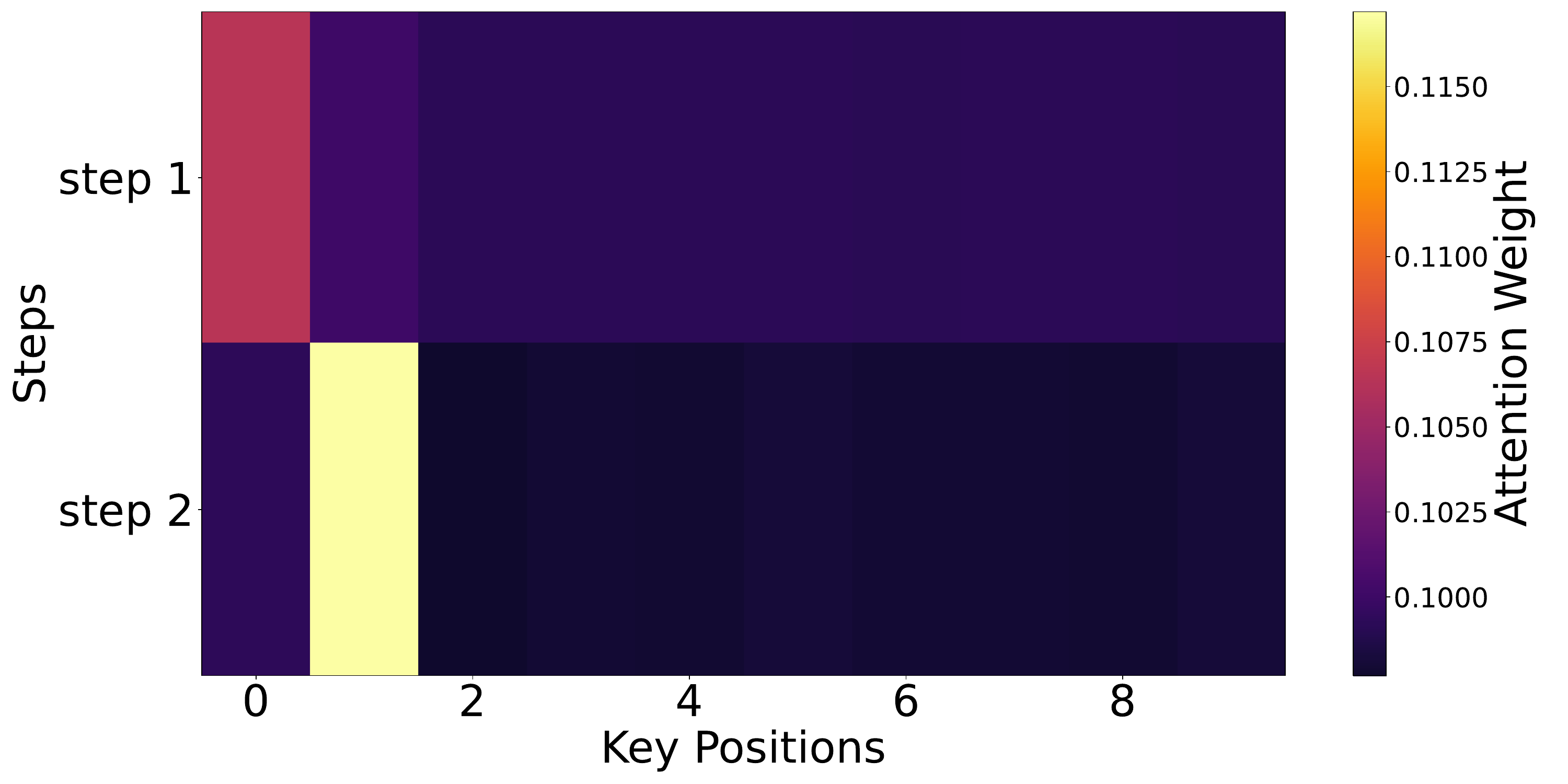}
    \caption{2-Step attention heatmap.}
    \label{fig:step_2}
  \end{subfigure}
  \begin{subfigure}[t]{0.23\textwidth}
    \centering
    \includegraphics[width=\linewidth]{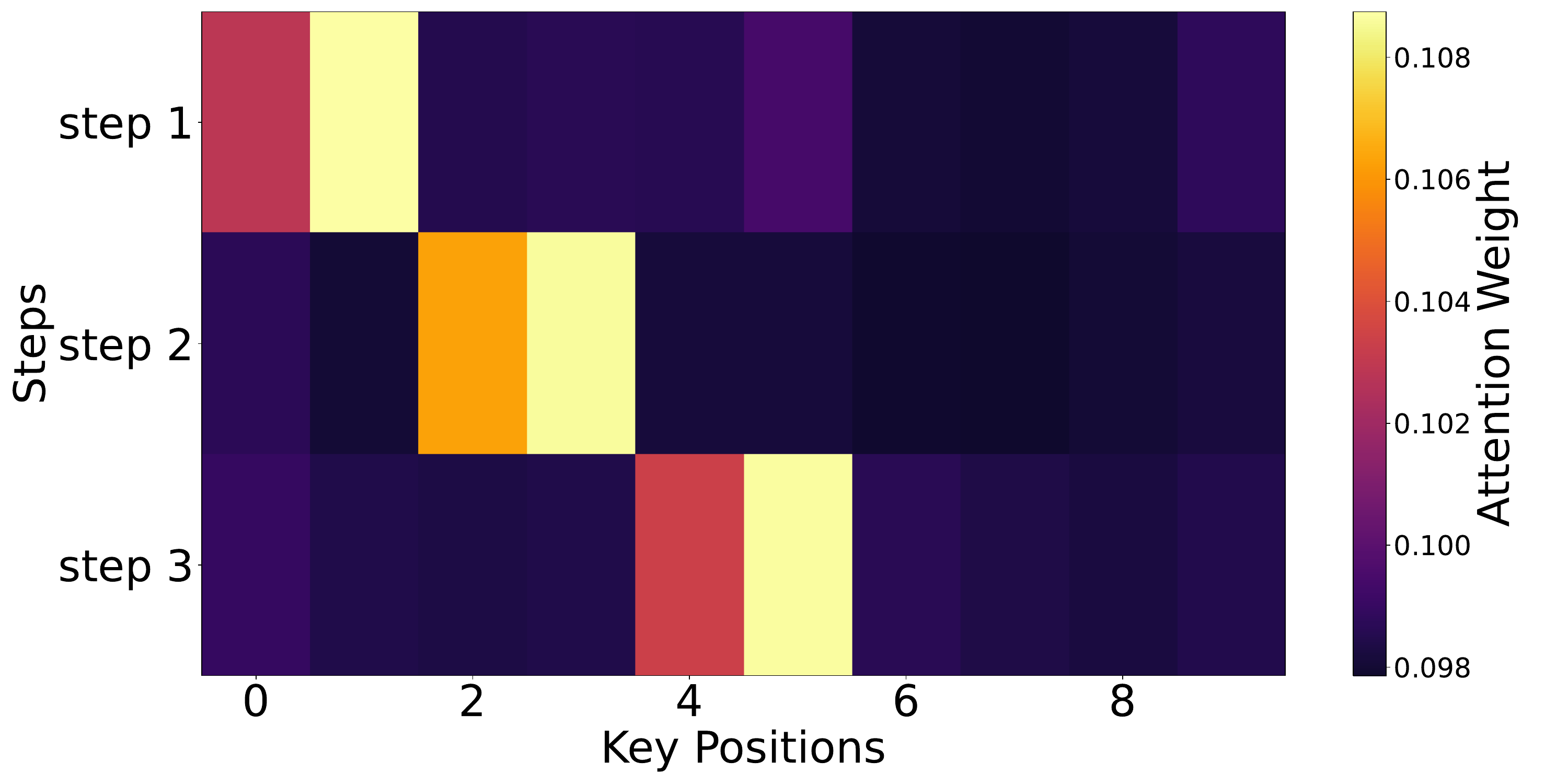}
    \caption{3-Step attention heatmap.}
    \label{fig:step_3}
  \end{subfigure}
  \begin{subfigure}[t]{0.23\textwidth}
    \centering
    \includegraphics[width=\linewidth]{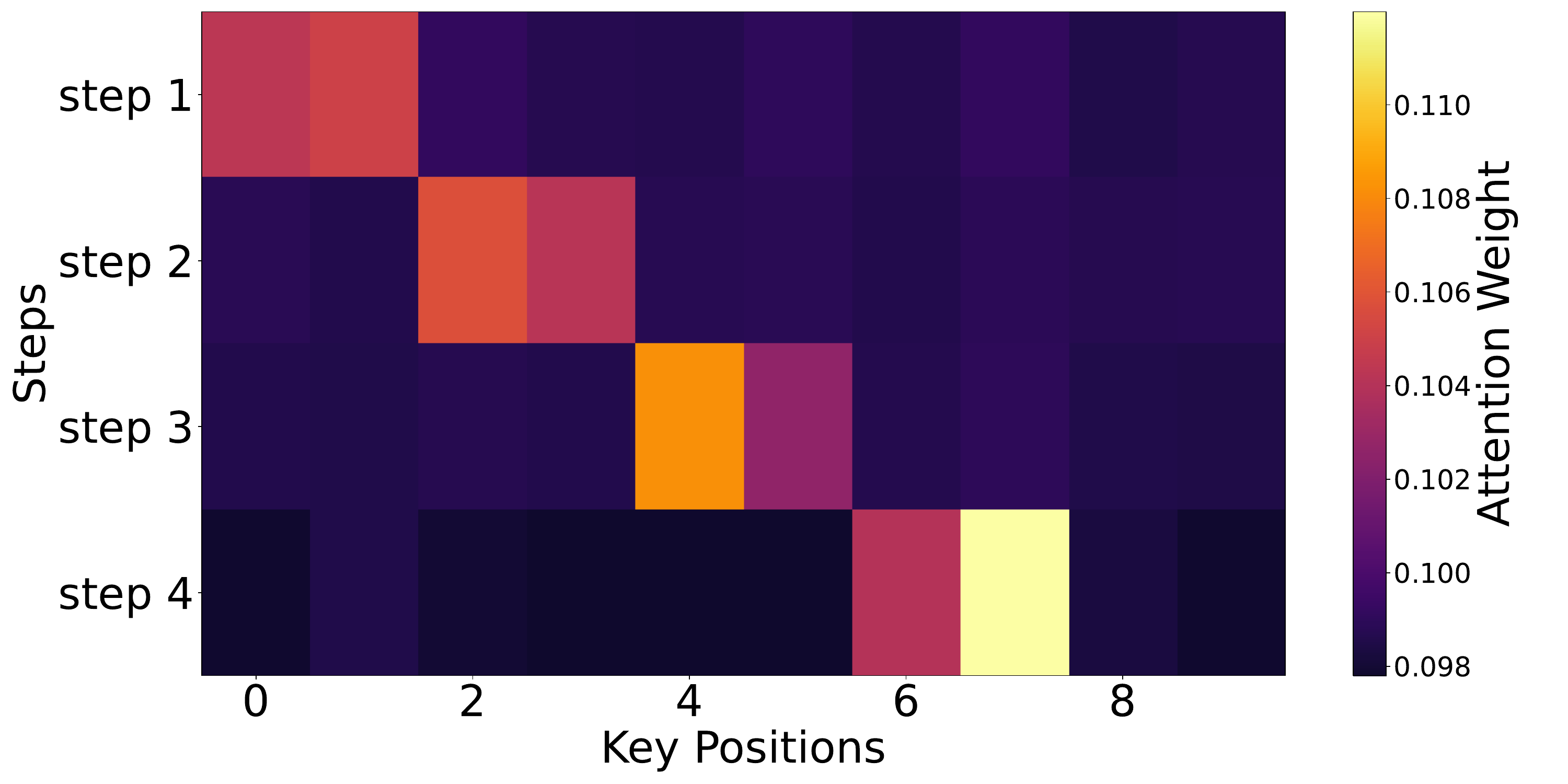}
    \caption{4-Step attention heatmap.}
    \label{fig:step_4}
  \end{subfigure}
  \caption{Heatmap of attention weights across reasoning steps. Required rules are ordered for clarity.
}
  \label{fig:attention_heatmap}
  \vspace{-1em}
\end{figure}

\subsection{Case Study}
\label{appendix:case-study}
To illustrate how LLMs fail under complex rule interactions, we present three case studies highlighting typical errors: rule omission, incorrect application, and action-type misalignment.

\begin{figure}[tbp]
    \centerline{\includegraphics[scale=0.22]{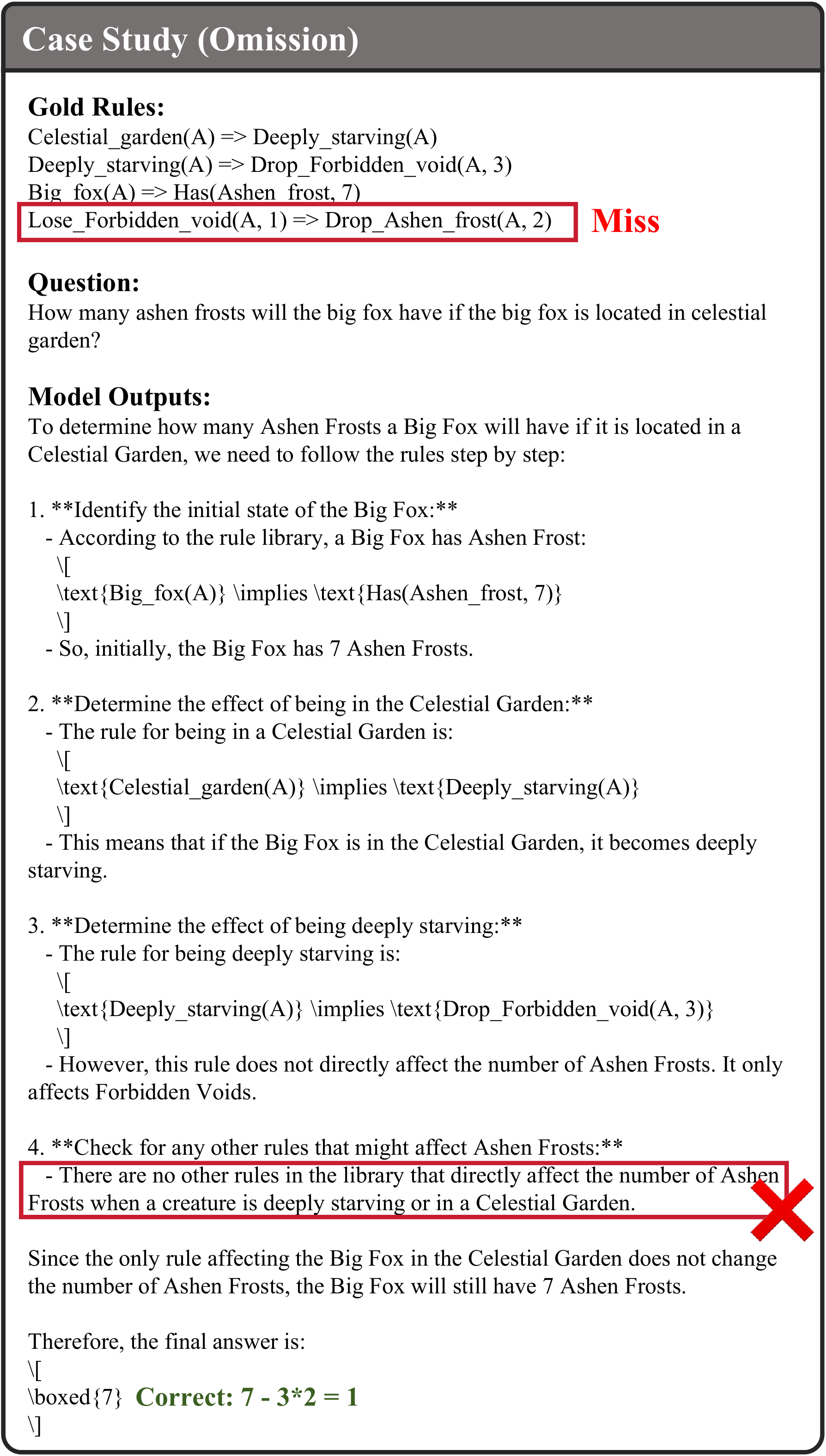}}%
    \caption{Case example of rule omission (missing rule application).}
    \label{fig:case_study_miss}
\end{figure}

\paragraph{Rule Omission}In the first case, as shown in Figure~\ref{fig:case_study_miss}, the model is required to determine how many ashen frosts a big fox has when placed in a celestial garden. Although the rule chain clearly specifies a cascading effect: ``\textit{Celestial\_garden(A)}'' induces ``\textit{Deeply\_starving(A)}'', which further triggers ``\textit{Drop\_Forbidden\_void(A, 3)}'', and the loss of Forbidden Voids in turn causes ``\textit{Drop\_Ashen\_frost(A, 2)}'', the model fails to propagate through this full multi-step sequence. Instead, it only identifies the initial attribute rule ``\textit{Big\_fox(A) $\Rightarrow$ Has(Ashen\_frost, 7)}'' and entirely omits the subsequent dependency leading to ashen-frost reduction. This omission produces the incorrect final answer of 7, while the correct reasoning yields $1 = (7 - 3\times2)$. This demonstrates that even when all rules are available, the model may prematurely terminate the reasoning chain and overlook critical transitions.

\begin{figure}[htbp]
    \centerline{\includegraphics[scale=0.22]{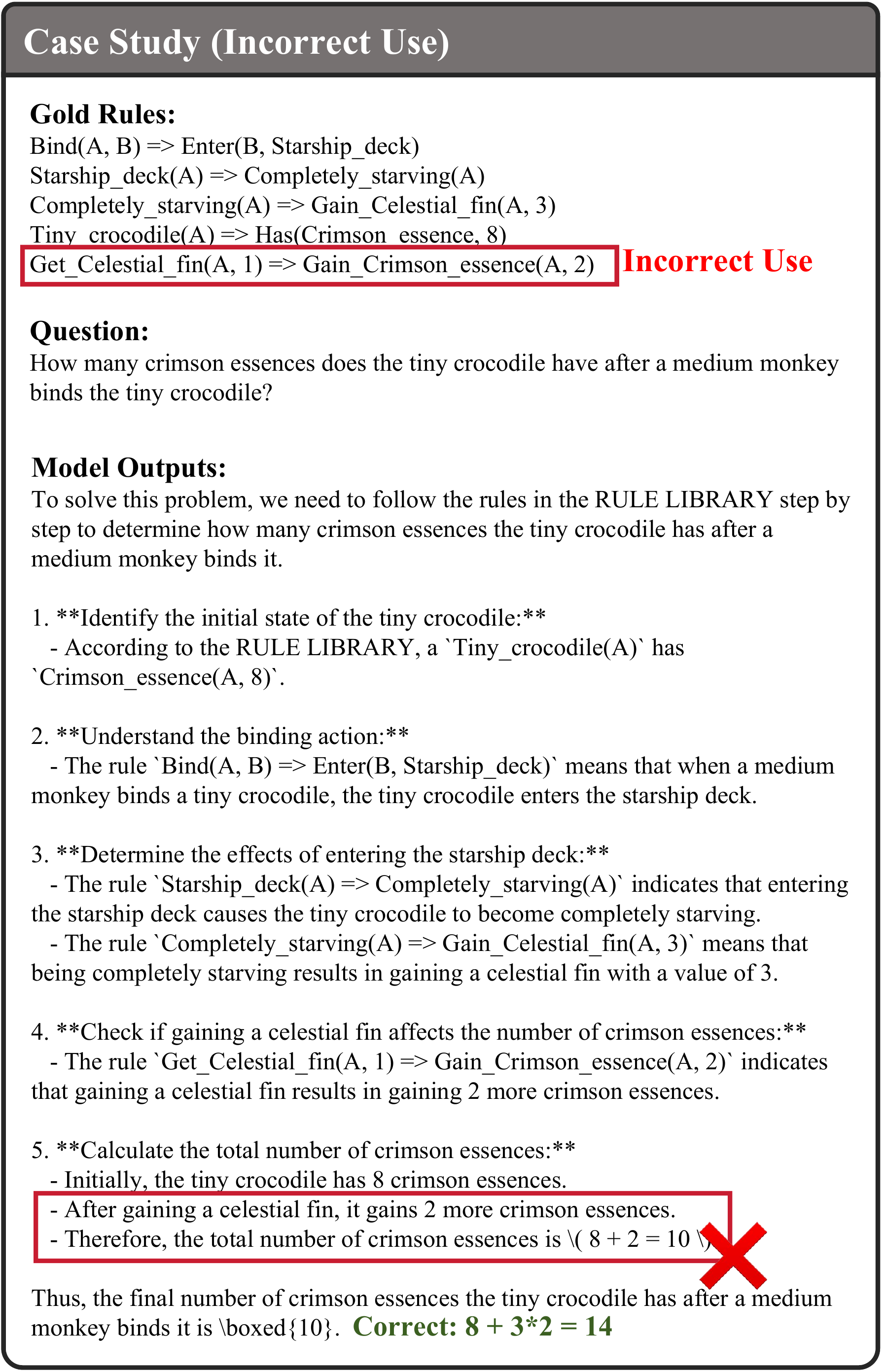}}%
    \caption{Case example of incorrect rule application.}
    \label{fig:case_study_wrong}
\end{figure}

\paragraph{Incorrect Use}The second case involves computing how many crimson essences a tiny crocodile possesses after being bound by another entity, as illustrated in Figure~\ref{fig:case_study_wrong}. Here, the model successfully identifies several relevant rules, including the induced state transitions from ``\textit{Bind(A, B)}'' to ``\textit{Enter(B, Starship\_deck)}'' and subsequently to ``\textit{Completely\_starving(B)}''. However, it misinterprets the rule that governs attribute changes: the model incorrectly applies the rule ``\textit{Get\_Celestial\_fin(A, 1) $\Rightarrow$ Gain\_Crimson\_essence(A, 2)}'' assuming that ``gaining a celestial fin'' automatically counts as ``getting'' one. This confusion leads the model to treat the growth of 3 celestial fins as the gain of only one fin, resulting in a total of 10 crimson essences, while the correct answer should be $14 = (8 + 3\times2)$. This reflects a deeper failure in aligning rule semantics, misidentifying rule triggers and misapplying effect operators.

\begin{figure}[htbp]
    \centerline{\includegraphics[scale=0.22]{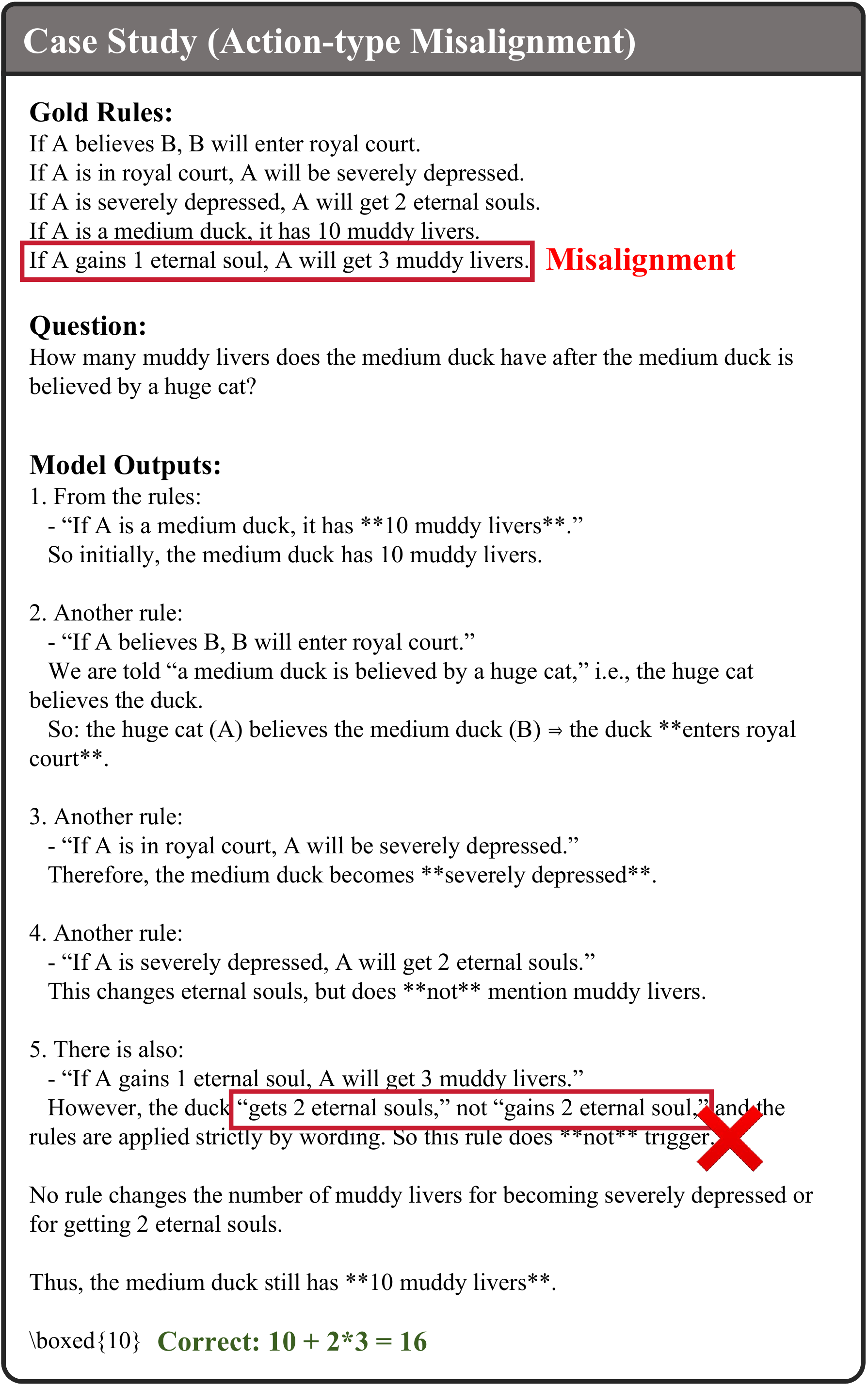}}%
    \caption{Case example of action-type misalignment.}
    \label{fig:case_study_act_mis}
\end{figure}

\paragraph{Action-type Misalignment} As shown in Figure~\ref{fig:case_study_act_mis}, the model makes an action-type misalignment error in which a neutral action operator such as ``gain'' fails to trigger when the system encounters a related but distinct operator, ``get''. Although the rules specify that gaining 1 eternal soul leads to receiving 3 muddy livers, the model incorrectly treats ``get 2 eternal souls'' as unrelated rather than decomposing it into two unit gain events or recognizing its semantic compatibility. As a result, the inference chain is prematurely terminated: the model concludes that no rule affects muddy livers after the eternal soul change, leading to the incorrect retention of 10 muddy livers. The correct reasoning requires triggering the gain rule twice, producing the correct result of 16 muddy livers. This example demonstrates that when a model rigidly separates action operators without recognizing their functional equivalence, even neutral and intuitively compatible operators fail to activate, causing a silent chain break and ultimately incorrect reasoning outcomes. 

Overall, the model frequently misuses injected rules by stopping early, misreading semantics, or missing operator equivalences, resulting in broken reasoning and wrong answers.

\subsection{Statistical Robustness}
\label{appendix:statistical_robustness}

For each method and rule scale, we evaluate five independently sampled, stratified test sets, each containing 110 questions for each rule representation. Thus, each scale involves 1,100 QA evaluations per method across FOL and NL. For QA, we first compute one seed-level overall score by averaging the displayed Single, PM(2--4), PM(5--8), MH-2, MH-3, and MH-4 scores under both FOL and NL. For retrieval, we average Recall@100, Recall@10, and Recall@1 under both representations. We then compute the sample standard deviation over the five seed-level scores. Tables~\ref{tab:qa_statistical_robustness} and~\ref{tab:retrieval_statistical_robustness} report the standard deviation and the $t$-based 95\% confidence-interval half-width in percentage points (\textit{std.} / $\pm$CI), where the half-width is $t_{0.975,4}\cdot \mathrm{std}/\sqrt{5}$.

\begin{table*}[t]
  \centering
  \small
  \setlength{\tabcolsep}{6pt}
  \begin{tabular}{lrrrrrr}
    \toprule
    Method & 100 & 500 & 1K & 2K & 5K & 10K \\
    \midrule
    Prompting & 1.6 / $\pm$1.9 & 2.4 / $\pm$3.0 & 2.0 / $\pm$2.5 & 4.1 / $\pm$5.1 & -- & -- \\
    $\text{RAG}_{\text{dense}}$ & 3.1 / $\pm$3.8 & 3.6 / $\pm$4.5 & 1.5 / $\pm$1.9 & 1.7 / $\pm$2.1 & 2.3 / $\pm$2.9 & 0.8 / $\pm$1.0 \\
    $\text{RAG}_{\text{bm25}}$ & 3.1 / $\pm$3.8 & 1.1 / $\pm$1.4 & 1.0 / $\pm$1.2 & 1.5 / $\pm$1.9 & 2.7 / $\pm$3.4 & 3.2 / $\pm$4.0 \\
    $\text{RAG}_{\text{hybrid}}$ & 2.0 / $\pm$2.5 & 1.1 / $\pm$1.4 & 2.3 / $\pm$2.9 & 1.4 / $\pm$1.7 & 3.5 / $\pm$4.3 & 2.3 / $\pm$2.9 \\
    KBLaM & 2.1 / $\pm$2.6 & 1.8 / $\pm$2.2 & 2.9 / $\pm$3.6 & 3.6 / $\pm$4.4 & 0.7 / $\pm$0.9 & 0.0 / $\pm$0.0 \\
    SR-KI & 2.0 / $\pm$2.5 & 1.7 / $\pm$2.2 & 0.6 / $\pm$0.8 & 1.2 / $\pm$1.4 & 1.8 / $\pm$2.3 & 3.0 / $\pm$3.7 \\
    \rowcolor{blue!7} DynaRule & 1.1 / $\pm$1.4 & 1.7 / $\pm$2.2 & 3.0 / $\pm$3.7 & 2.4 / $\pm$3.0 & 2.6 / $\pm$3.2 & 3.4 / $\pm$4.3 \\
    \bottomrule
  \end{tabular}
  \caption{Seed-level variation for exact match accuracy, reported as standard deviation / $t$-based 95\% confidence-interval half-width (percentage points).}
  \label{tab:qa_statistical_robustness}
  \vspace{-0.6em}
\end{table*}

\begin{table*}[t]
  \centering
  \small
  \setlength{\tabcolsep}{6pt}
  \begin{tabular}{lrrrrrr}
    \toprule
    Method & 100 & 500 & 1K & 2K & 5K & 10K \\
    \midrule
    Dense & 1.0 / $\pm$1.2 & 1.7 / $\pm$2.1 & 1.6 / $\pm$2.0 & 1.6 / $\pm$1.9 & 1.5 / $\pm$1.8 & 1.3 / $\pm$1.6 \\
    BM25 & 0.8 / $\pm$1.1 & 1.1 / $\pm$1.4 & 1.2 / $\pm$1.5 & 1.8 / $\pm$2.2 & 1.5 / $\pm$1.9 & 2.0 / $\pm$2.5 \\
    Hybrid & 0.7 / $\pm$0.8 & 1.0 / $\pm$1.3 & 0.6 / $\pm$0.8 & 0.6 / $\pm$0.7 & 1.0 / $\pm$1.2 & 1.4 / $\pm$1.7 \\
    KBLaM & 1.1 / $\pm$1.3 & 1.3 / $\pm$1.6 & 1.3 / $\pm$1.6 & 1.4 / $\pm$1.7 & 1.3 / $\pm$1.6 & 4.1 / $\pm$5.1 \\
    SR-KI & 0.5 / $\pm$0.6 & 0.7 / $\pm$0.8 & 0.6 / $\pm$0.8 & 1.0 / $\pm$1.3 & 1.4 / $\pm$1.8 & 1.3 / $\pm$1.6 \\
    \rowcolor{blue!7} DynaRule & 0.4 / $\pm$0.4 & 0.3 / $\pm$0.4 & 0.6 / $\pm$0.7 & 0.8 / $\pm$1.0 & 1.3 / $\pm$1.6 & 1.1 / $\pm$1.3 \\
    \bottomrule
  \end{tabular}
  \caption{Seed-level variation for retrieval, reported as standard deviation / $t$-based 95\% confidence-interval half-width (percentage points).}
  \label{tab:retrieval_statistical_robustness}
  \vspace{-0.6em}
\end{table*}

Overall, the seed-level variation is modest and the confidence intervals are much smaller than the main performance gaps at large rule scales. At 10K rules, DynaRule's QA confidence interval is $\pm$4.3 points, while its gains over SR-KI and Hybrid RAG are +19.9 and +37.6 points, respectively. Its retrieval confidence interval is only $\pm$1.3 points.

\subsection{Detailed Retrieval Results across Reasoning Steps and QA Types}
\label{appendix:Retrieval-Performance-across-different-Reasoning-Steps}
We summarize the retrieval results into three main findings.

\begin{figure*}[tbp]
    \centerline{\includegraphics[scale=0.355]{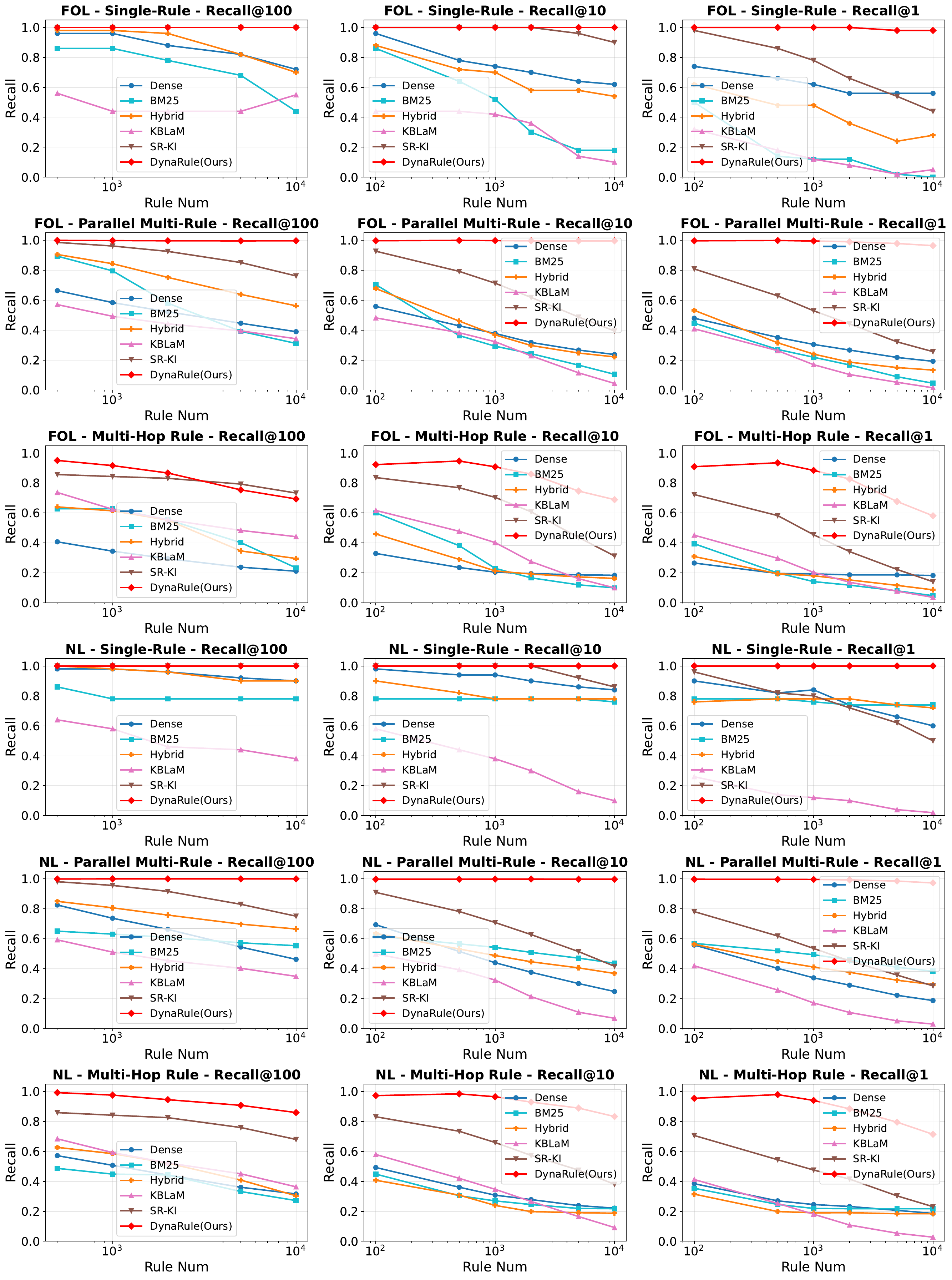}}%
    \caption{Comparison of retrieval performance across QA types under FOL (top) and NL (bottom) rule representations. Dense uses Qwen3-Embedding-8B; Hybrid uses RRF with a fusion constant of 60; KBLaM, SR-KI, and DynaRule use confidence-layer retrieval. Recall@1 sets $K$ to the number of gold rules per instance or step.}
    \label{fig:retrieval_qa_performance}
\end{figure*}

\textbf{(1)} Figure~\ref{fig:step_retrieval_performance} shows that step-level retrieval remains strong across both FOL and NL rule representations, with recall generally staying above 0.8 as the rule set size increases from 100 to 10000. In particular, Step 1 remains nearly perfect under all metrics, indicating that early-stage retrieval is highly reliable.
\textbf{(2)} As reasoning depth increases from Step 2 to Step 4 in Figure~\ref{fig:step_retrieval_performance}, retrieval performance gradually declines, especially under stricter metrics such as Recall@1. This trend suggests that deeper multi-step reasoning introduces cumulative retrieval difficulty, with FOL showing a slightly sharper drop than NL at larger rule scales.
\textbf{(3)} Figure~\ref{fig:retrieval_qa_performance} further shows that DynaRule consistently outperforms all baselines across QA types and rule injection settings. Its performance is near perfect on Single-Rule and Parallel Multi-Rule QA, while Multi-Hop Rule QA shows more noticeable degradation, highlighting the greater sensitivity of sequential reasoning to intermediate retrieval errors.

\subsection{Attention Weights across Reasoning Steps}
\label{appendix:attention_weight_across_steps}
We visualize the attention distribution over rule keys under different reasoning depths, with step-specific rules arranged sequentially for clarity, as shown in Figure~\ref{fig:attention_heatmap}. For single-step tasks, the model concentrates almost entirely on the first required rule. As the reasoning depth increases, attention shifts progressively toward the rules needed at each step. When a step requires multiple rules, the model jointly attends to them, forming a distinct cluster clearly separated from those used in earlier steps. Overall, this produces a stepwise, staircase-like attention pattern, highlighting the model’s ability to dynamically select and apply the correct rules throughout multi-step reasoning.

\begin{figure}[htbp]
    \centerline{\includegraphics[scale=0.25]{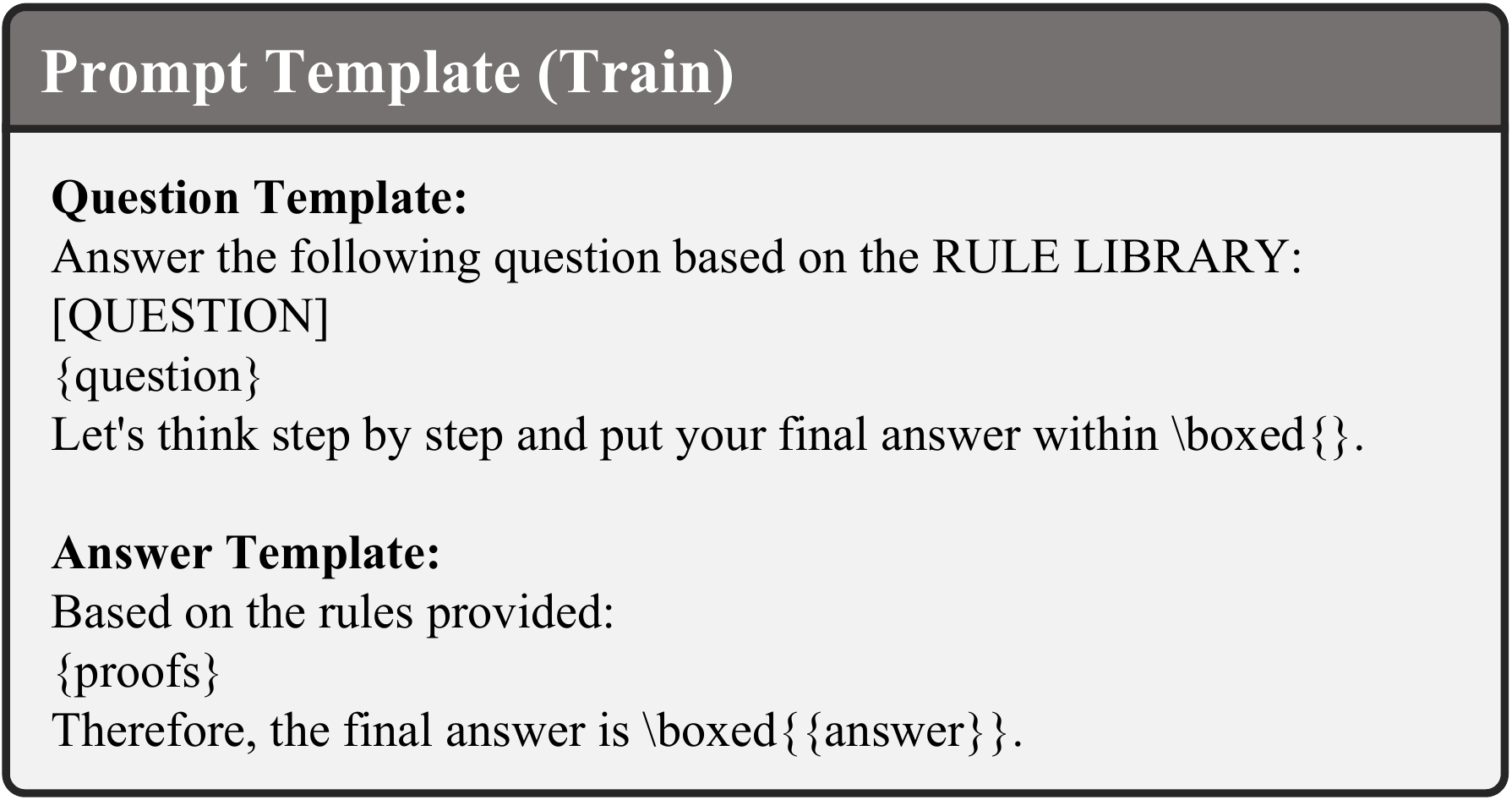}}%
    \caption{Question-answering template for DynaRule training.}
    \label{fig:QA_template_train}
\end{figure}

\begin{table*}[h]
  \centering 
  \setlength{\tabcolsep}{6pt} 
  \scalebox{0.5}{%
    \begin{tabular}{%
      c
      *{1}{c}
      *{7}{c}
      *{3}{c}
      c
    }
    \toprule
    \multirow{2}{*}{Method}
      & \multicolumn{1}{c}{\textbf{Single-Rule QA}}
      & \multicolumn{7}{c}{\textbf{Parallel Multi-Rule QA}}
      & \multicolumn{3}{c}{\textbf{Multi-Hop Rule QA}}
      & \multirow{2}{*}{\textbf{Avg.}} \\
    \cmidrule(lr){2-2}
    \cmidrule(lr){3-9}
    \cmidrule(lr){10-12}
      & \multicolumn{1}{c}{single-rule}
      & \multicolumn{1}{c}{multi-rule-2}
      & \multicolumn{1}{c}{multi-rule-3}
      & \multicolumn{1}{c}{multi-rule-4}
      & \multicolumn{1}{c}{multi-rule-5}
      & \multicolumn{1}{c}{multi-rule-6}
      & \multicolumn{1}{c}{multi-rule-7}
      & \multicolumn{1}{c}{multi-rule-8}
      & \multicolumn{1}{c}{multi-hop-2}
      & \multicolumn{1}{c}{multi-hop-3}
      & \multicolumn{1}{c}{multi-hop-4}
      & \\ 
    \midrule[0.8pt]\multicolumn{13}{c}{\textbf{FOL Representation}} \\ \midrule[0.8pt]    
    \multicolumn{13}{c}{\textit{Rule Num = 100}} \\ \midrule
        Qwen2.5-14B-Instruct & 0.9200 & 0.8300 & 0.8400 & 0.7633 & 0.7483 & 0.6450 & 0.6950 & 0.6250 & 0.6400 & 0.1600 & 0.1000 & 0.6333 \\
        Qwen2.5-32B-Instruct & 0.9600 & \underline{0.9900} & 0.9000 & 0.8433 & 0.8267 & 0.7433 & 0.6950 & 0.6700 & 0.6000 & 0.2600 & 0.1400 & 0.6935 \\
        Qwen2.5-72B-Instruct & 0.9200 & 0.9700 & 0.9433 & 0.9250 & 0.8783 & 0.8117 & 0.8950 & 0.8850 & 0.7400 & 0.6000 & 0.4800 & 0.8226 \\
        Llama-3.1-70B-Instruct & \underline{0.9800} & 0.9600 & \underline{0.9567} & 0.8183 & 0.8683 & 0.8033 & 0.8250 & 0.8300 & 0.8000 & 0.4600 & 0.4400 & 0.7947 \\
        Llama-3.3-70B-Instruct & 0.9600 & 0.9200 & 0.8467 & 0.7983 & 0.8333 & 0.7900 & 0.7900 & 0.8250 & 0.8600 & 0.5200 & 0.2400 & 0.7621 \\
        Qwen3-32B & 0.8000 & 0.9500 & 0.9133 & 0.8950 & 0.9267 & 0.8917 & 0.8050 & 0.7850 & 0.8800 & 0.7800 & 0.4400 & 0.8242 \\
        deepseek-chat & 0.8600 & 0.9400 & 0.9433 & 0.8733 & \underline{0.9517} & 0.9017 & 0.9000 & 0.9300 & \underline{0.9200} & 0.4400 & 0.2600 & 0.8109 \\
        gpt-5.1-2025-11-13 & 0.9600 & 0.9100 & 0.9233 & 0.8750 & 0.9017 & 0.8583 & 0.9000 & 0.9450 & 0.7400 & 0.3000 & 0.1800 & 0.7721 \\
        gpt-5.5-2026-04-23 & \underline{0.9800} & 0.9200 & 0.9500 & 0.9750 & 0.8100 & 0.9700 & \textbf{0.9900} & \underline{0.9500} & 0.7300 & 0.3300 & 0.3100 & 0.8105 \\
        claude-sonnet-4-6 & 0.9200 & 0.6200 & 0.9500 & \textbf{0.9850} & 0.8200 & \textbf{0.9950} & \underline{0.9800} & \underline{0.9500} & 0.5200 & \textbf{0.9750} & \underline{0.6250} & \underline{0.8491} \\
                                \rowcolor{blue!7} DynaRule & \textbf{1.0000} & \textbf{1.0000} & \textbf{0.9950} & \underline{0.9800} & \textbf{0.9950} & \underline{0.9850} & 0.9700 & \textbf{0.9950} & \textbf{0.9600} & \underline{0.9300} & \textbf{0.9500} & \textbf{0.9782} \\
    \midrule
    \multicolumn{13}{c}{\textit{Rule Num = 500}} \\ \midrule
        Qwen2.5-14B-Instruct & 0.8000 & 0.8500 & 0.8300 & 0.7617 & 0.5633 & 0.4283 & 0.3000 & 0.1450 & 0.4200 & 0.0400 & 0.0000 & 0.4671 \\
        Qwen2.5-32B-Instruct & 0.8200 & 0.8800 & 0.8333 & 0.7600 & 0.6700 & 0.5050 & 0.3350 & 0.2000 & 0.5400 & 0.0800 & 0.0400 & 0.5148 \\
        Qwen2.5-72B-Instruct & 0.7600 & \underline{0.9500} & 0.9167 & 0.8566 & 0.7933 & 0.6350 & 0.5050 & 0.3250 & 0.6800 & 0.2600 & 0.1600 & 0.6220 \\
        Llama-3.1-70B-Instruct & 0.9000 & 0.8500 & 0.8300 & 0.7283 & 0.7617 & 0.6017 & 0.5050 & 0.3050 & 0.5200 & 0.2400 & 0.1200 & 0.5783 \\
        Llama-3.3-70B-Instruct & 0.9000 & 0.8600 & 0.8500 & 0.6833 & 0.6167 & 0.4000 & 0.3250 & 0.1800 & 0.7000 & 0.0200 & 0.0800 & 0.5105 \\
        Qwen3-32B & 0.7000 & 0.8100 & 0.7900 & 0.6950 & 0.6150 & 0.4617 & 0.3450 & 0.3350 & 0.7800 & 0.3800 & 0.3000 & 0.5647 \\
        deepseek-chat & 0.7800 & 0.9400 & 0.9500 & 0.8333 & \underline{0.8617} & 0.8067 & 0.8300 & \underline{0.7750} & \underline{0.8200} & 0.2800 & 0.1600 & 0.7306 \\
        gpt-5.1-2025-11-13 & 0.9000 & 0.9300 & 0.8933 & 0.8600 & 0.8183 & 0.7800 & 0.7800 & 0.7450 & 0.6800 & 0.2200 & 0.1000 & 0.7006 \\
        gpt-5.5-2026-04-23 & \underline{0.9800} & 0.9250 & \underline{0.9550} & 0.9750 & 0.8100 & 0.8000 & 0.9000 & \underline{0.7750} & 0.6200 & 0.3200 & 0.2100 & 0.7518 \\
        claude-sonnet-4-6 & 0.9100 & 0.7250 & 0.9500 & \underline{0.9850} & 0.8250 & \underline{0.9250} & \textbf{0.9750} & \textbf{0.9850} & 0.4300 & \textbf{0.7400} & \underline{0.4400} & \underline{0.8082} \\
                                \rowcolor{blue!7} DynaRule & \textbf{1.0000} & \textbf{1.0000} & \textbf{0.9867} & \textbf{0.9950} & \textbf{0.9833} & \textbf{1.0000} & \underline{0.9650} & \textbf{0.9850} & \textbf{0.9800} & \underline{0.7200} & \textbf{0.8000} & \textbf{0.9468} \\
    \midrule
    \multicolumn{13}{c}{\textit{Rule Num = 1000}} \\ \midrule
        Qwen2.5-14B-Instruct & 0.7200 & 0.6900 & 0.6333 & 0.7150 & 0.5266 & 0.2950 & 0.1900 & 0.0900 & 0.5400 & 0.0200 & 0.0200 & 0.4036 \\
        Qwen2.5-32B-Instruct & 0.8400 & 0.7900 & 0.7167 & 0.7100 & 0.6200 & 0.3600 & 0.2200 & 0.1100 & 0.5400 & 0.0800 & 0.0000 & 0.4533 \\
        Qwen2.5-72B-Instruct & 0.7200 & 0.7700 & 0.8067 & 0.8483 & 0.7067 & 0.5000 & 0.3700 & 0.1600 & 0.6000 & 0.1000 & 0.0600 & 0.5129 \\
        Llama-3.1-70B-Instruct & 0.8600 & 0.8000 & 0.8133 & 0.6983 & 0.6400 & 0.3983 & 0.3600 & 0.1900 & 0.4600 & 0.1000 & 0.0600 & 0.4891 \\
        Llama-3.3-70B-Instruct & 0.8600 & 0.7000 & 0.6833 & 0.6517 & 0.6083 & 0.2750 & 0.2200 & 0.0500 & 0.4600 & 0.0400 & 0.0000 & 0.4135 \\
        Qwen3-32B & 0.5800 & 0.7100 & 0.7834 & 0.7117 & 0.6483 & 0.4100 & 0.3300 & 0.1350 & 0.5600 & 0.1200 & 0.1000 & 0.4626 \\
        deepseek-chat & 0.8800 & \underline{0.9200} & 0.9033 & 0.7817 & \underline{0.8250} & 0.7050 & 0.6950 & 0.6450 & \underline{0.7800} & 0.1400 & 0.1200 & 0.6723 \\
        gpt-5.1-2025-11-13 & 0.9600 & 0.8800 & 0.8933 & 0.7867 & 0.7733 & 0.6950 & 0.6850 & 0.5900 & 0.7400 & 0.1400 & 0.0800 & 0.6567 \\
        gpt-5.5-2026-04-23 & \underline{0.9800} & 0.9100 & \underline{0.9550} & 0.9250 & 0.7000 & 0.8500 & 0.7750 & 0.9000 & 0.5300 & 0.3150 & 0.2150 & 0.7323 \\
        claude-sonnet-4-6 & 0.9000 & 0.8300 & 0.9500 & \underline{0.9650} & 0.7750 & \underline{0.9250} & \underline{0.9750} & \underline{0.9620} & 0.4100 & \textbf{0.7200} & \underline{0.4100} & \underline{0.8020} \\
                                \rowcolor{blue!7} DynaRule & \textbf{1.0000} & \textbf{1.0000} & \textbf{0.9917} & \textbf{0.9875} & \textbf{0.9792} & \textbf{0.9625} & \textbf{0.9812} & \textbf{0.9688} & \textbf{0.8500} & \underline{0.6250} & \textbf{0.6500} & \textbf{0.9087} \\
    \midrule[0.8pt]\multicolumn{13}{c}{\textbf{NL Representation}} \\ \midrule[0.8pt]   
    \multicolumn{13}{c}{\textit{Rule Num = 100}} \\ \midrule
        Qwen2.5-14B-Instruct & 0.9000 & 0.8600 & 0.9067 & 0.8217 & 0.7900 & 0.7333 & 0.6900 & 0.7350 & 0.7400 & 0.2200 & 0.3800 & 0.7070 \\
        Qwen2.5-32B-Instruct & 0.9400 & 0.9300 & 0.9333 & 0.8633 & 0.9084 & 0.7767 & 0.8400 & 0.7600 & 0.7600 & 0.2400 & 0.2400 & 0.7447 \\
        Qwen2.5-72B-Instruct & 0.8800 & 0.9100 & \underline{0.9600} & 0.9133 & \underline{0.9717} & 0.9133 & 0.9250 & 0.9200 & 0.7600 & 0.5400 & 0.5400 & 0.8394 \\
        Llama-3.1-70B-Instruct & 0.9600 & 0.8700 & 0.8567 & 0.8500 & 0.8983 & 0.7317 & 0.8300 & 0.9050 & 0.5200 & 0.4200 & 0.5600 & 0.7638 \\
        Llama-3.3-70B-Instruct & \underline{0.9800} & 0.9200 & 0.9433 & 0.7800 & 0.8950 & 0.8150 & 0.7950 & 0.8100 & 0.7000 & 0.4400 & 0.4800 & 0.7780 \\
        Qwen3-32B & 0.8400 & 0.9500 & 0.9500 & 0.9200 & 0.9250 & 0.8017 & 0.8250 & 0.7950 & 0.7800 & 0.7800 & \underline{0.7600} & 0.8479 \\
        deepseek-chat & 0.9000 & \underline{0.9600} & 0.9333 & 0.8950 & 0.9500 & 0.8783 & 0.8700 & 0.9300 & \textbf{0.9600} & 0.7200 & 0.6600 & 0.8779 \\
        gpt-5.1-2025-11-13 & \underline{0.9800} & 0.9300 & 0.9400 & 0.9000 & 0.9083 & 0.8517 & 0.9100 & 0.8900 & 0.8600 & 0.3600 & 0.3600 & 0.8082 \\
        gpt-5.5-2026-04-23 & \textbf{1.0000} & 0.9000 & 0.9500 & \textbf{1.0000} & 0.7500 & \underline{0.9750} & \underline{0.9500} & 0.8750 & 0.7000 & \underline{0.9000} & 0.6000 & 0.8727 \\
        claude-sonnet-4-6 & \textbf{1.0000} & 0.8200 & 0.9500 & 0.9667 & 0.8000 & 0.9500 & \textbf{0.9750} & \textbf{1.0000} & 0.8000 & \underline{0.9000} & \textbf{0.9000} & \underline{0.9147} \\
                                \rowcolor{blue!7} DynaRule & \textbf{1.0000} & \textbf{1.0000} & \textbf{0.9950} & \underline{0.9950} & \textbf{0.9900} & \textbf{0.9800} & \textbf{0.9750} & \underline{0.9850} & \underline{0.9200} & \textbf{0.9600} & 0.7000 & \textbf{0.9545} \\
    \midrule
    \multicolumn{13}{c}{\textit{Rule Num = 500}} \\ \midrule
        Qwen2.5-14B-Instruct & 0.8000 & 0.6500 & 0.7000 & 0.6233 & 0.4850 & 0.3650 & 0.2100 & 0.1150 & 0.6600 & 0.0200 & 0.0200 & 0.4226 \\
        Qwen2.5-32B-Instruct & 0.8800 & 0.8600 & 0.8133 & 0.7900 & 0.7067 & 0.5133 & 0.4100 & 0.3700 & 0.6000 & 0.0600 & 0.0800 & 0.5530 \\
        Qwen2.5-72B-Instruct & 0.7200 & 0.9000 & 0.8700 & 0.8067 & 0.8350 & 0.7050 & 0.6850 & 0.6700 & 0.5000 & 0.2200 & 0.1400 & 0.6411 \\
        Llama-3.1-70B-Instruct & \textbf{1.0000} & 0.9100 & 0.8733 & 0.7967 & 0.7267 & 0.5917 & 0.5000 & 0.5850 & 0.5400 & 0.1400 & 0.1200 & 0.6167 \\
        Llama-3.3-70B-Instruct & \textbf{1.0000} & 0.8500 & 0.8433 & 0.7483 & 0.5667 & 0.4817 & 0.3950 & 0.3250 & 0.6000 & 0.0600 & 0.0600 & 0.5391 \\
        Qwen3-32B & 0.8000 & 0.7400 & 0.7633 & 0.7117 & 0.5850 & 0.4550 & 0.3500 & 0.3250 & 0.7200 & 0.3400 & 0.5000 & 0.5718 \\
        deepseek-chat & 0.8600 & \underline{0.9200} & 0.8933 & 0.8450 & 0.8400 & 0.7617 & 0.7250 & 0.6900 & \underline{0.8200} & 0.4200 & 0.3800 & 0.7414 \\
        gpt-5.1-2025-11-13 & 0.9600 & 0.9100 & 0.9200 & \underline{0.8866} & \underline{0.8467} & 0.7433 & 0.7400 & 0.7600 & \underline{0.8200} & 0.1800 & 0.2000 & 0.7242 \\
        gpt-5.5-2026-04-23 & \textbf{1.0000} & 0.9000 & \underline{0.9500} & \textbf{1.0000} & 0.8000 & 0.9000 & 0.8500 & \underline{0.9750} & 0.5000 & \textbf{0.9000} & 0.5000 & 0.8432 \\
        claude-sonnet-4-6 & 0.8900 & 0.8000 & \underline{0.9500} & \textbf{1.0000} & 0.7667 & \underline{0.9500} & \textbf{1.0000} & \textbf{1.0000} & 0.6000 & \textbf{0.9000} & \underline{0.8000} & \underline{0.8779} \\
                                \rowcolor{blue!7} DynaRule & \underline{0.9800} & \textbf{0.9800} & \textbf{0.9950} & \textbf{1.0000} & \textbf{0.9783} & \textbf{0.9800} & \underline{0.9600} & \underline{0.9750} & \textbf{0.8800} & \underline{0.7000} & \textbf{0.8200} & \textbf{0.9317} \\
     \midrule
    \multicolumn{13}{c}{\textit{Rule Num = 1000}} \\ \midrule
        Qwen2.5-14B-Instruct & 0.6200 & 0.6000 & 0.6067 & 0.6533 & 0.4600 & 0.3117 & 0.1450 & 0.0450 & 0.5600 & 0.0200 & 0.0000 & 0.3656 \\
        Qwen2.5-32B-Instruct & 0.7400 & 0.8200 & 0.7700 & 0.7767 & 0.5683 & 0.4050 & 0.3050 & 0.1050 & 0.6000 & 0.0400 & 0.0200 & 0.4682 \\
        Qwen2.5-72B-Instruct & 0.7600 & 0.8000 & 0.7300 & 0.8533 & 0.7100 & 0.6100 & 0.5400 & 0.4450 & 0.4400 & 0.0800 & 0.0400 & 0.5462 \\
        Llama-3.1-70B-Instruct & 0.9600 & 0.7800 & 0.7800 & 0.7200 & 0.6600 & 0.4933 & 0.2950 & 0.2150 & 0.4000 & 0.0000 & 0.0400 & 0.4858 \\
        Llama-3.3-70B-Instruct & 0.9600 & 0.6900 & 0.7467 & 0.7050 & 0.5083 & 0.3400 & 0.2250 & 0.0650 & 0.5600 & 0.0000 & 0.0000 & 0.4364 \\
        Qwen3-32B & 0.8800 & 0.7000 & 0.7433 & 0.6433 & 0.5133 & 0.3733 & 0.3600 & 0.1200 & 0.6400 & 0.1800 & 0.1800 & 0.4848 \\
        deepseek-chat & 0.9000 & 0.8700 & 0.8567 & 0.7883 & \underline{0.7600} & 0.6450 & 0.6000 & 0.5450 & \underline{0.8000} & 0.1800 & 0.1600 & 0.6459 \\
        gpt-5.1-2025-11-13 & \textbf{1.0000} & 0.8900 & 0.8833 & 0.8717 & 0.7583 & 0.6583 & 0.6250 & 0.6350 & 0.7000 & 0.1000 & 0.0400 & 0.6511 \\
        gpt-5.5-2026-04-23 & 0.8000 & \underline{0.9000} & \underline{0.9500} & \textbf{1.0000} & 0.7000 & \textbf{1.0000} & \underline{0.8750} & \textbf{0.9500} & \underline{0.8000} & 0.4000 & 0.2000 & 0.7795 \\
        claude-sonnet-4-6 & 0.8300 & 0.7000 & \underline{0.9500} & \textbf{1.0000} & 0.7417 & 0.8250 & \underline{0.8750} & 0.9250 & 0.3300 & \textbf{0.8100} & \textbf{0.7900} & \underline{0.7979} \\
                                \rowcolor{blue!7} DynaRule & \underline{0.9800} & \textbf{0.9800} & \textbf{0.9817} & \underline{0.9883} & \textbf{0.9783} & \underline{0.9783} & \textbf{0.9550} & \underline{0.9450} & \textbf{0.8400} & \underline{0.6400} & \underline{0.6800} & \textbf{0.9042} \\
    \bottomrule
    \end{tabular}%
  }
  \caption{Prompting comparison results on the RuleWorld benchmark under the \textbf{FOL} and \textbf{NL} rule representations, compared with DynaRule.}
  \label{tab:extended_icl_results}
\end{table*}

\end{document}